\documentclass[letterpaper, 10 pt, conference]{ieeeconf}  

\IEEEoverridecommandlockouts                              

\usepackage{float}
\usepackage{soul}
\usepackage{mathtools}
\usepackage{amssymb}
\usepackage{bbm}
\usepackage{graphicx}
\usepackage{multicol}
\usepackage{multirow}
\usepackage[export]{adjustbox}
\usepackage[bookmarks=true]{hyperref}
\usepackage[table]{xcolor}    
\hypersetup{
    colorlinks=true,
    linkcolor=blue,
    filecolor=magenta,      
    urlcolor=cyan,
}
\usepackage[raggedright]{sidecap}
\usepackage[font=small,skip=5pt]{caption}
\usepackage{subcaption}
\usepackage[ruled]{algorithm2e}

\let\labelindent\relax
\usepackage{enumitem,kantlipsum}
\usepackage[tight-spacing=true]{siunitx}
\usepackage{booktabs}
\usepackage{dsfont}
\usepackage{algcompatible}
\usepackage{gensymb}
\usepackage[hang,flushmargin]{footmisc}
\usepackage{lipsum}
\renewcommand{\baselinestretch}{0.96}

\newcommand{\rulesep}{\unskip\ \vrule\ }

\newcommand{\edit}[1]{{{#1}}}

\let\OLDthebibliography\thebibliography
\renewcommand\thebibliography[1]{
  \OLDthebibliography{#1}
  \setlength{\parskip}{0pt}
  \setlength{\itemsep}{7pt plus 0.3ex}
}

\makeatletter
\let\NAT@parse\undefined
\makeatother
\usepackage[numbers]{natbib}

\title{\LARGE \bf ReLMoGen: Integrating Motion Generation in Reinforcement Learning for Mobile Manipulation}

\author{
  Fei Xia$^{*1}$, 
  Chengshu Li$^{*1}$,
  Roberto Mart\'in-Mart\'in$^1$,
  Or Litany$^2$,
  Alexander Toshev$^3$,
  Silvio Savarese$^1$
  \thanks{$^*$ Equal contribution. $^{1}$ Stanford University. $^{2}$ Nvidia. $^{3}$ Robotics at Google.}
}
\begin{document}
\maketitle
\thispagestyle{empty}
\pagestyle{empty}

\begin{abstract}
Many Reinforcement Learning (RL) approaches use joint control signals (positions, velocities, torques) as action space for continuous control tasks. We propose to lift the action space to a higher level in the form of subgoals for a motion generator (a combination of motion planner and trajectory executor). We argue that, by lifting the action space and by leveraging sampling-based motion planners, we can efficiently use RL to solve complex, long-horizon tasks that could not be solved with existing RL methods in the original action space. We propose ReLMoGen -- a framework that combines a learned policy to predict subgoals and a motion generator to plan and execute the motion needed to reach these subgoals. To validate our method, we apply ReLMoGen to two types of tasks: 1) Interactive Navigation tasks, navigation problems where interactions with the environment are required to reach the destination, and 2) Mobile Manipulation tasks, manipulation tasks that require moving the robot base. These problems are challenging because they are usually long-horizon, hard to explore during training, and comprise alternating phases of navigation and interaction. Our method is benchmarked on a diverse set of seven robotics tasks in photo-realistic simulation environments. In all settings, ReLMoGen outperforms state-of-the-art RL and Hierarchical RL baselines. ReLMoGen also shows outstanding transferability between different motion generators at test time, indicating a great potential to transfer to real robots. For more information, please visit project website: \url{http://svl.stanford.edu/projects/relmogen}.  
\end{abstract}

%


\section{Introduction}
\label{s_intro}

Many tasks in mobile manipulation are defined by a sequence of navigation and manipulation subgoals. Navigation moves the robot's base to a configuration where arm interaction can succeed. For example, when trying to access a closed room, the robot needs to navigate to the front of the door to push it with the arm or, alternatively, to press a button next to the door that activates its automatic opening mechanism. Such a sequence of subgoals is well parameterized as spatial points of interest in the environment to reach with the robot's base or end-effector. The path towards these points is mostly irrelevant as long as it is feasible for the robot's kinematics and does not incur collisions. 

Collision-free feasible trajectories to points of interest can be efficiently computed and executed by a motion generator (MG) composed of a motion planner (MP) and a trajectory controller~\cite{lavalle2006planning,Siciliano:2007:SHR:1209344}. MGs specialize in moving the robot's base or end-effector to a given short-range point, usually within the field of view so that they can use an accurate model of the environment. However, due to the sample complexity of large Euclidean space and the lack of accurate models of the entire environment, MGs cannot solve the problem of long-range planning to a point beyond sight. Moreover, MGs excel at answering ``how'' to move to a point, but not ``where'' to move to achieve the task based on current observations, where deep RL~\cite{sutton2018reinforcement,arulkumaran2017deep} has shown strong results.

\begin{figure}[!t]
\centering
\begin{subfigure}{0.36\textwidth}
\includegraphics[width=\linewidth]{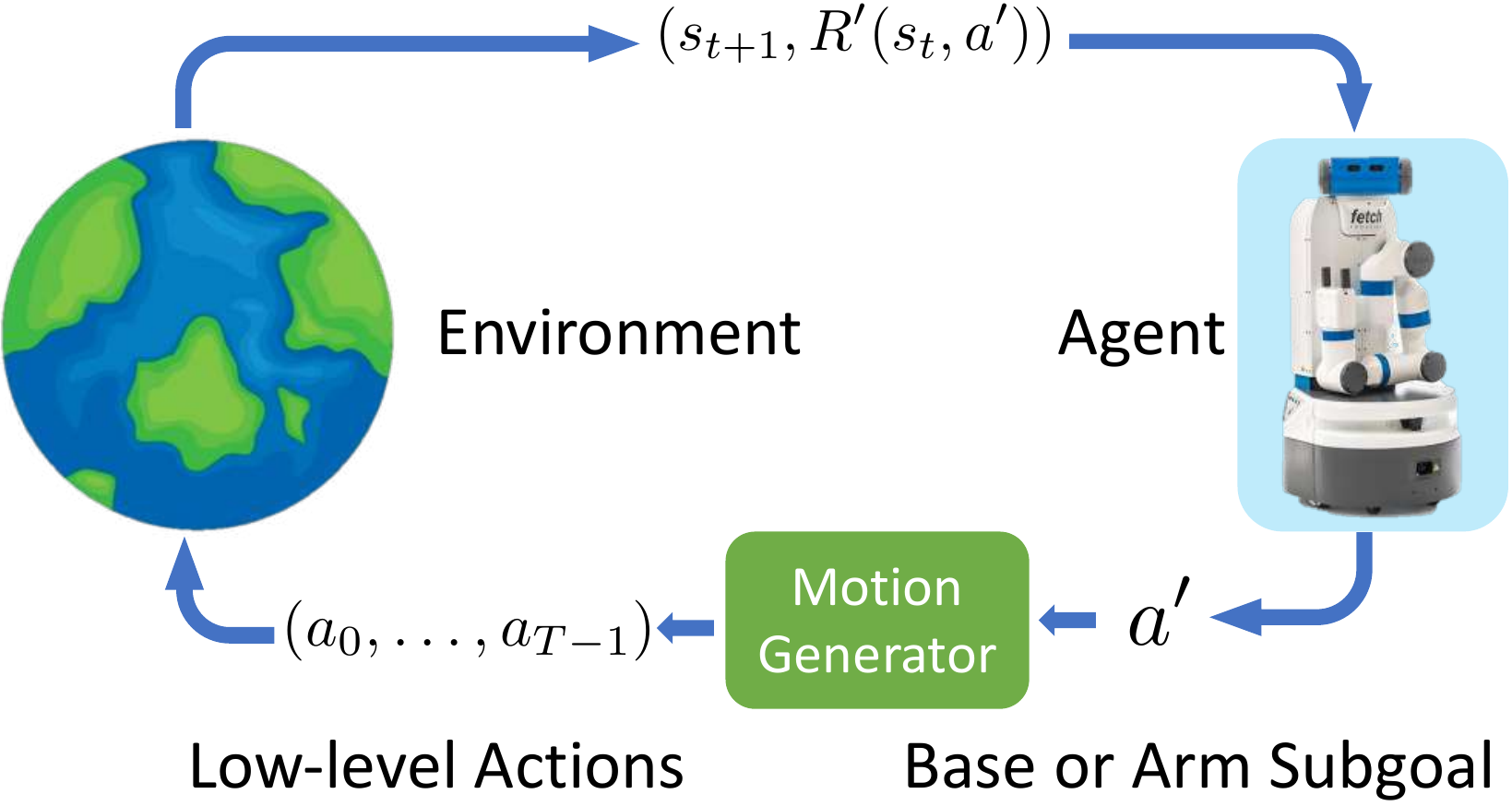}
\vspace{-3mm}
\label{fig:subgoal1}
\end{subfigure}

\begin{subfigure}{0.36\textwidth}
\includegraphics[width=\linewidth]{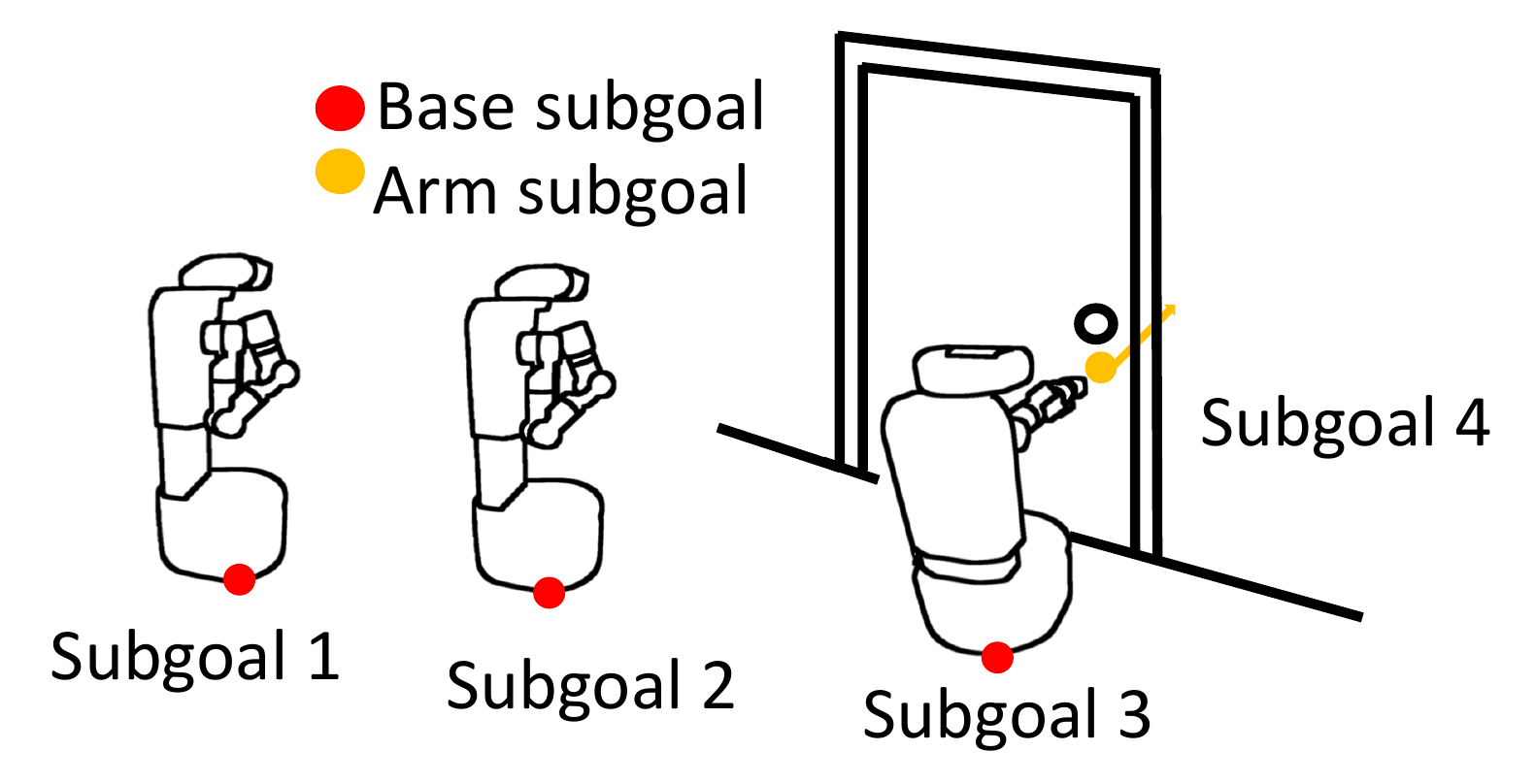}
\label{fig:subgoal2}
\vspace{-4mm}
\end{subfigure}

\caption{\small{{(top) We propose to integrate motion generation into a reinforcement learning loop to lift the action space from low-level robot actions $a$ to subgoals for the motion generator $a'$ (bottom) The mobile manipulation tasks we can solve with ReLMoGen are composed by a sequence of base and arm subgoals (e.g. pushing open a door for Interactive Navigation).}}}
\label{fig:method}
\vspace{-5mm}


\end{figure}


RL has been successfully applied to solve visuo-motor tasks dealing with continuous control based on high dimensional observations~\cite{zhu2017target, quan2020novel, zeng2020tossingbot, kalashnikov2018qt, mahler2017dex, levine2016end, zeng2018learning}. However, this methodology falls short for mobile manipulation tasks, which involve long sequences of precise low-level actions to reach the aforementioned spatial points of interest. Often, the steps in free space do not return any reward, rendering mobile manipulation hard exploration problems~\cite{osband2016deep,osband2019deep}. While the exploration challenge may be mitigated by a hierarchical structure~\cite{levy2018learning, nachum2018data, nachum2019does, li2020hrl4in}, the RL agent still dedicates a large portion of its training steps to learning to move towards spatial points of interest without collisions from scratch. 


In this work, we present ReLMoGen (Reinforcement Learning + Motion Generation), a novel approach that combines the strengths of RL and MG to overcome their individual limitations in mobile manipulation domains. Specifically, we propose to employ RL to obtain policies that map observations to subgoals that indicate desired base or arm motion. These subgoals are then passed to a MG that solves for precise, collision-free robot control between consecutive subgoals. We refer to the resulting policy as Subgoal Generation Policy (SGP). 

Considering the mobile manipulation task as a Markov Decision Processes (MDPs), ReLMoGen can be thought of as creating a \textit{lifted} MDP, where the action space is re-defined to the space of MG subgoals. This presents a temporal abstraction where the policy learns to produce a shorter sequence of ``subgoal actions'', which makes exploration easier for the policy, as demonstrated in the experimental analysis. From a control perspective, ReLMoGen is a hierarchical controller, whose high-level module is a learned controller while the low-level module is a classical one.

The contributions of this paper are as follows. First, we demonstrate how to marry learning-based methods with classical methods to leverage their advantages. We thoroughly study the interplay between two RL algorithms, Deep Q Learning~\cite{mnih2013playing} and Soft-Actor Critic~\cite{haarnoja2018soft}, and two established motion planners, Rapidly expanding Random Trees~\cite{kuffner2000rrt} and Probabilistic Random Maps~\cite{bohlin2000path}. 
Further, we demonstrate that ReLMoGen consistently achieves higher performance across a wide variety of long-horizon robotics tasks. We study the approach in the context of navigation, stationary manipulation and mobile manipulation. ReLMoGen is shown to explore more efficiently, converge faster and output highly interpretable subgoal actions. Finally, we show that the learned policies can be applied with different motion planners, even with those not used during training. This demonstrates the robustness and practicality of our approach and great potential in real-world deployment.

\section{Related Work}

\label{s_rw}

ReLMoGen relates to previous efforts to combine robot learning and motion generation. At a conceptual level, it can also be thought of as a hierarchical RL approach with a stationary low-level policy. Therefore, we will relate to previous work in these areas.

\textbf{Combining Learning and Motion Generation:}
Recently, researchers have attempted to overcome limitations of \textit{classical} sampling- or optimization-based motion generators by combining them with machine learning techniques. There are two well-known limitations of classical MGs: 1) they depend on an accurate environment model, and 2) their computational complexity grows exponentially with the search space dimension. Researchers have proposed learning-based solutions that map partial observations to waypoints~\cite{muller2018driving,kaufmann2019beauty,Tamar-RSS-19,qureshi2019motion} or trajectories~\cite{bansal2019combining}, thus bypassing the trajectory searching problem. They rely on expert MG supervision to learn via imitation. In contrast, we do not attempt to improve MG but rather integrate it into a RL loop as is. The opposite has also been attempted: improving the exploration of RL agents using experiences from a MG~\cite{levine2013guided, jetchev2010trajectory,rana2018towards, ota2020efficient}. 
We only use MGs to map from our lifted action space to low-level motor control signals during training.

Closely related to our approach are works that integrate a planner or a motion generator as-it-is into RL procedure. For example, \citet{jiang2018integrating} integrates a task and motion planner (TAMP) with RL: the TAMP planner provides solutions for room-to-room navigation that RL refines. \citet{dragan2011learning} learns to set goals for an optimization-based motion generator based on predefined features that describe the task. In a concurrent work~\cite{yamadamotion}, the authors propose to augment an RL agent with the option of using a motion planner and formulate the learning problem as a semi-MDP. Unlike their work, we propose to lift the action space completely instead of using a semi-MDP setup. We also tackle much more complex domain than the domain of 2D block pushing with stationary arm in \citet{yamadamotion}. \citet{wu2020spatial} is the most similar method to ours. They propose an approach for mobile manipulation that learns to set goals for a 2D navigation motion planner by selecting pixels on the local occupancy map. Their ``spatial action maps'' serve as a new action space for policy learning with DQN~\cite{mnih2013playing}. As we will see later, this approach is similar to our variant of ReLMoGen with Q-learning based RL (see ReLMoGen-D Sec.~\ref{subsection:parameterization_architecture}). However, our solution enables both navigation and manipulation with a robotic arm. Moreover, we demonstrate with ReLMoGen-R (Sec.~\ref{subsection:parameterization_architecture}) that our proposed method can be also applied to policy-gradient methods.

\textbf{Hierarchical Reinforcement Learning:} 
Often in HRL solutions, the main benefit comes from a better exploration thanks to a longer temporal commitment of a low-level policy towards the goal commanded by a high-level policy~\cite{nachum2019does}. Therefore, in many HRL methods the high level learns to set subgoals for the low level~\cite{heess2016learning,kulkarni2016hierarchical, vezhnevets2017feudal, nachum2018data, nachum2018near, levy2018learning}. Notably, \citet{konidaris2011autonomous} applies HRL to Interactive Navigation (IN) tasks, problems that require the agent to interact with the environment to reach its goal. Their algorithms generate actions to solve subcomponents of the original task and reuses them to solve new task instances. \citet{li2020hrl4in} propose an end-to-end HRL solution for IN that also decides on the different parts of the embodiment to use for each subgoal. HRL solutions often suffer from training instability because the high level and low level are learned simultaneously. Previous attempts to alleviate this include off-policy corrections~\cite{nachum2018data}, hindsight subgoal sampling~\cite{levy2018learning} and low-level policy pre-training~\cite{heess2016learning}. While ReLMoGen is not a full HRL solution, it is structurally similar: a high level sets subgoals for a low level. Therefore, ReLMoGen benefits from better exploration due to temporal abstraction while avoiding the aforementioned cold-start problem because our low level is not a learned policy but a predefined MG solution. 

An orthogonal but related area of RL research is deep exploration. These methods typically rely on uncertainty modeling, random priors, or noisy data, and have proven to be effective in simple tasks such as Cartpole, DeepSea and Atari games~\cite{osband2016deep, osband2018randomized, osband2019deep}. Closest to our task setup, \citet{ciosek2019better} proposes an Optimistic Actor Critic that approximates a lower and upper confidence bound on the Q-functions, and shows favorable results in MuJoCo environments. ReLMoGen can be thought of as improving exploration, not by relying on optimism of Q-functions, but by lifting the action space, circumventing the hard exploration problem with commitment towards an interaction point.

\section{RL with Motion Generation}
\label{s_method}


We formulate a visuo-motor mobile manipulation control task as a time-discrete Partially Observable Markov Decision Process (POMDP) defined by the tuple $\mathcal{M} = (\mathcal{S},\mathcal{A},\mathcal{O}, \mathcal{T},\mathcal{R},\gamma)$. Here, $\mathcal{S}$ is the state space; $\mathcal{A}$ is the action space; $\mathcal{O}$ is the observation space; $\mathcal{T}(s'|s,a), s\in\mathcal{S}, a\in\mathcal{A}$, is the state transition model; $\mathcal{R}(s, a) \in \mathbb{R}$ is the reward function; $\gamma$ is the discount factor. We assume that the state is not directly observable and we learn a policy $\pi(a|o)$ conditioned on observations $o \in \mathcal{O}$. Herein, the agent following the policy $\pi$ obtains an observation $o_t$ at time $t$ and performs an action $a_t$, receiving from the environment an immediate reward $r_t$ and a new observation $o_{t+1}$. The goal of RL is to learn an action selection policy $\pi$ that maximizes the discounted sum of rewards.

We assume that $\mathcal{A}$ is the original action space for continuous control of the mobile manipulator, e.g. positions or velocities of each joint. Our main assumption in ReLMoGen is that, for the considered types of mobile manipulation tasks, a successful strategy can be described as a sequence of subgoals: $\tau_{succ} = \{a'_0, \ldots, a'_{\tau-1}\}$. Each subgoal $a'_i$ corresponds to a goal configuration for a motion generator either to move the base to a desired location or to move the robot's end-effector to a position and perform a parameterized interaction. These subgoals are generated by a subgoal generation policy, $\pi_\mathit{SGP}$. As shown in Sec.~\ref{s_exp}, in this work we focus on mobile manipulation tasks that can be solved by applying a parameterized pushing interaction after positioning the arm at a subgoal location; however, we do not find any aspect of ReLMoGen that fundamentally restricts it from utilizing other parameterized interactions at the desired end-effector position (e.g. pull, pick, place, \ldots). 


To generate collision-free trajectories, we propose to query at each policy step a motion generator, $\mathit{MG}$, a non-preemptable subroutine that attempts to find and execute an embodiment-compliant and collision-free trajectory to reach a given subgoal. The motion generator takes as input a subgoal from the subgoal generator policy, $a'$, and outputs a sequence of variable length $T$ of low-level actions that are executed, $\mathit{MG}(a') = (a_0, \ldots, a_{T-1})$. In case the MG fails to find a path, it returns a no-op. 
The proposed ReLMoGen solution is composed of two elements: the motion generator, $\mathit{MG}$, and the subgoal generation policy, $\pi_\mathit{SGP}$.

Based on the MG, we build with ReLMoGen a new \textit{lifted} PODMP, $\mathcal{M} = (\mathcal{S},\mathcal{A}',\mathcal{O}, \mathcal{T}',\mathcal{R}',\gamma)$, where $a' \in \mathcal{A}'$ is a new action space of subgoals to query the MG. $\mathcal{T}'(s'|s,a'), s,s'\in\mathcal{S}, a'\in\mathcal{A}'$ is the new transition function that corresponds to iteratively querying the original transition function $\mathcal{T}(s'|s,a)$ for $T$ times starting at $s_{t}$, with the sequence of actions returned by the MG, $\mathit{MG}(a') = (a_t, \ldots, a_{t+T-1})$.
Finally, the \textit{lifted} reward is defined as the accumulated reward obtained from executing the sequence of actions from the MG, $\mathcal{R}'(s_t, a'_t) = \sum_{k=t}^{t + T - 1}\mathcal{R}(s_k, a_k)$. 
The subgoal generator policy is trained to solve this lifted POMDP, taking in observations $o$ and outputting actions $a'$, subgoals for the MG.
The composition of the trained subgoal generator policy and the MG is a policy that solves the original POMDP: $\pi = \mathit{MG}(\pi_\mathit{SGP})$.
As a summary, ReLMoGen lifts the original POMDP problem into this new formulation that can be more easily solved using reinforcement learning. 

\begin{figure*}[!t]
\centering
\begin{subfigure}{0.55\textwidth}
\includegraphics[width=\linewidth]{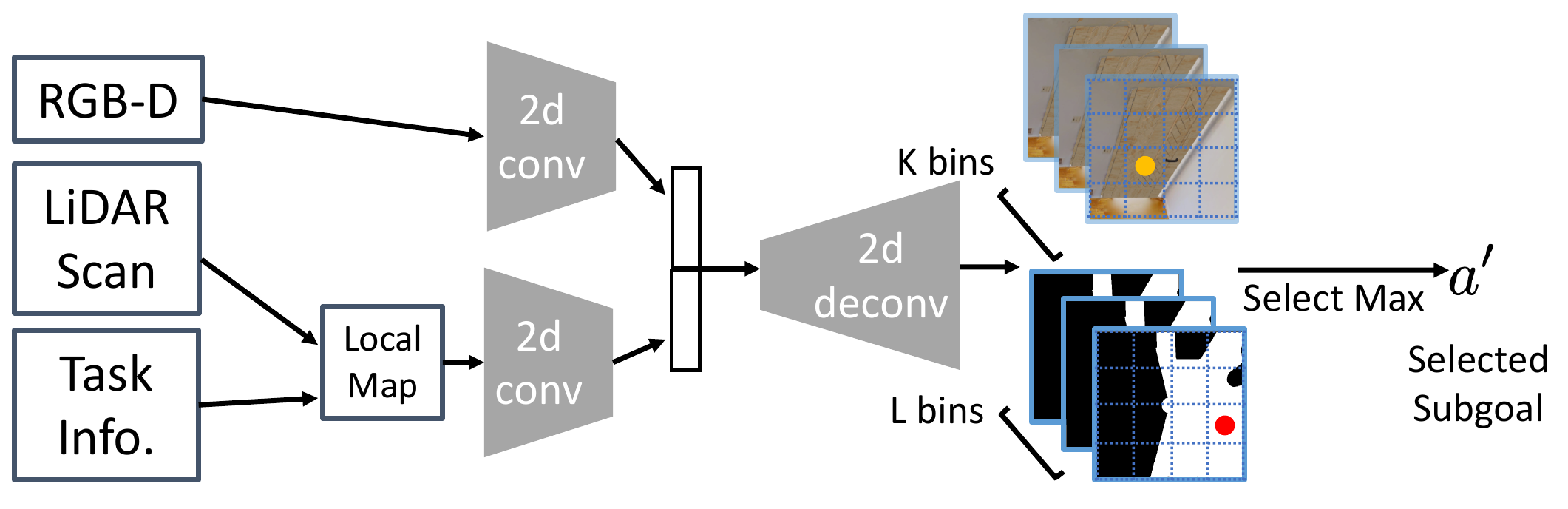}
\caption{\small{Subgoal Generation Policy SGP-D}}
\label{fig:method_architecture_d}
\end{subfigure}
\rulesep
\begin{subfigure}{0.40\textwidth}
\includegraphics[width=\linewidth]{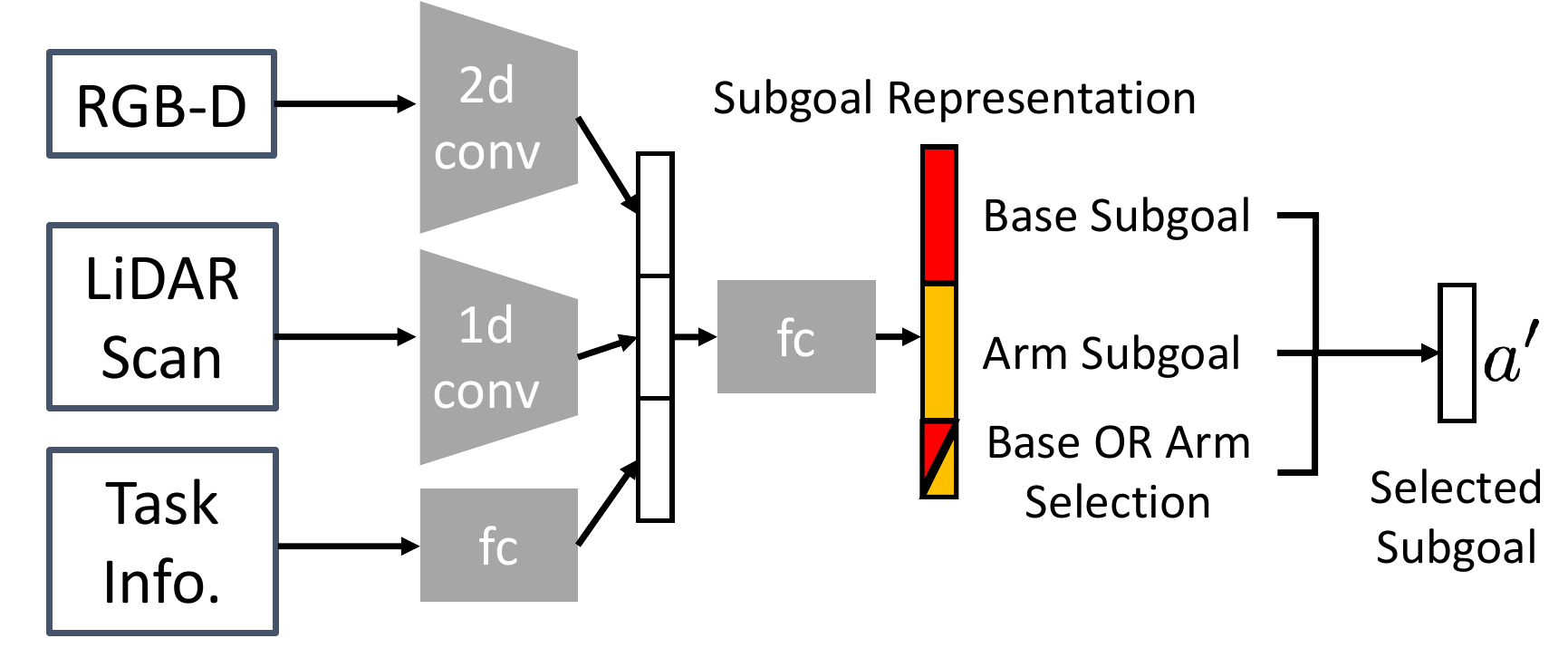}
\caption{\small{Subgoal Generation Policy SGP-R}}
\label{fig:method_architecture_r}
\end{subfigure}
\caption{\small{Two types of action parameterization of ReLMoGen and network architecture of SGP-D and SGP-R.}}
\label{fig:method_architecture}
\vspace{-4mm}
\end{figure*}


\vspace{-5mm}
\subsection{ReLMoGen: RL with Motion Generation Action Space}
\label{subsection:parameterization_architecture}

In this section, we propose our solutions to the lifted POMDP created by ReLMoGen for mobile manipulation tasks. As explained above, ReLMoGen is a general procedure that comprises two elements, a subgoal generation policy (SGP) and a motion generator (MP). We show that ReLMoGen can be instantiated with continuous and discrete action parametrization with two alternative SGPs that we formalize.

\begin{figure}

\begin{center}
\begin{subfigure}{0.15\textwidth}
    \includegraphics[width=\linewidth]{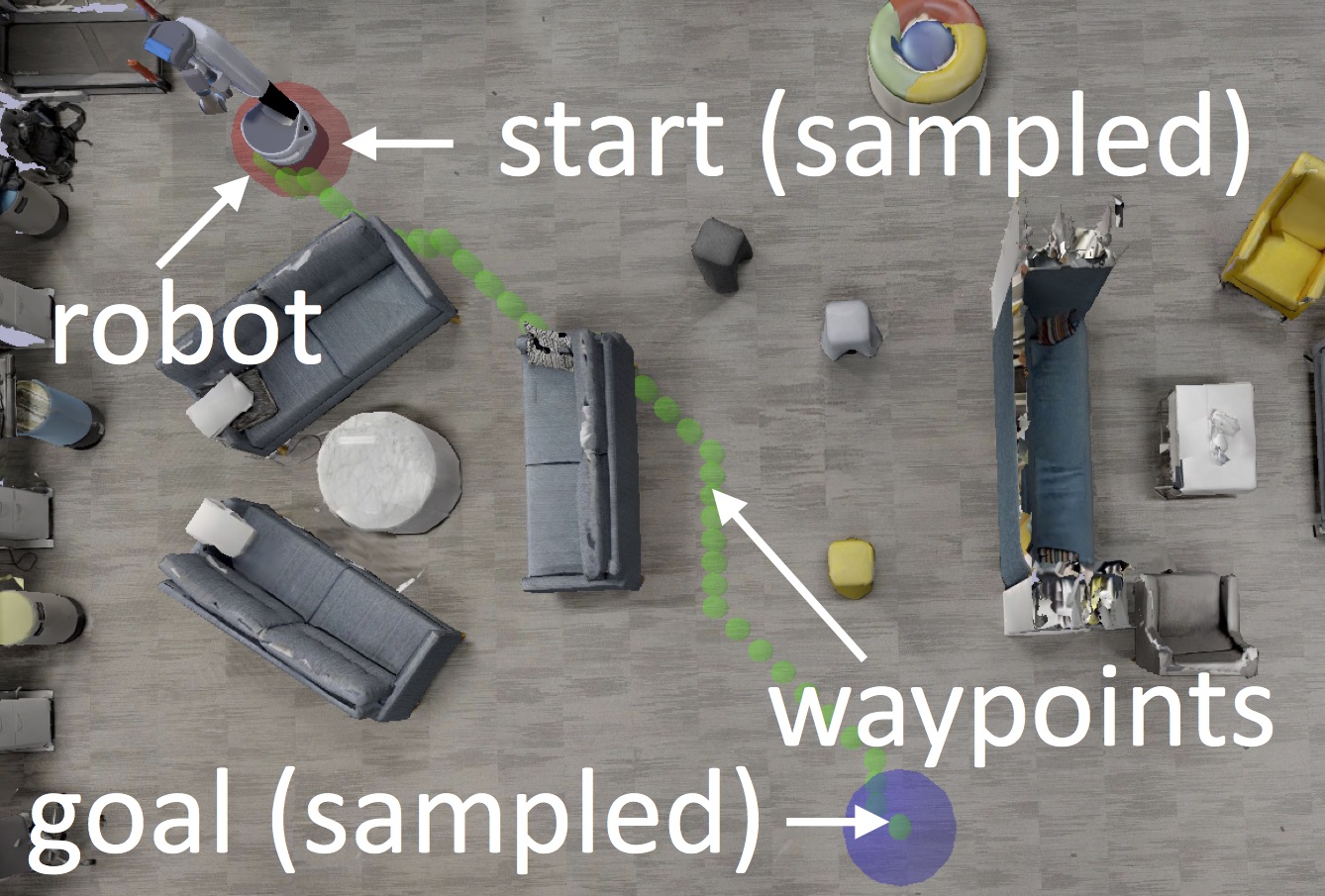}
    \caption{\tiny{\texttt{PointNav}}}
\end{subfigure}
\begin{subfigure}{0.16\textwidth}
    \includegraphics[width=\linewidth]{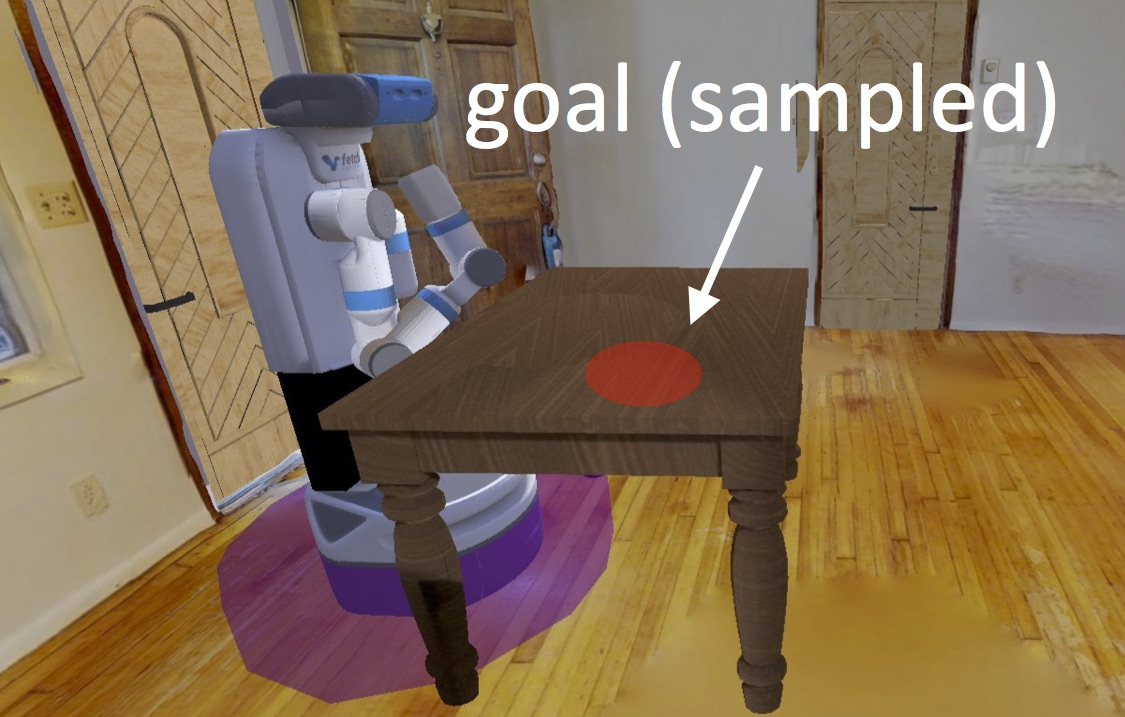}
    \caption{\tiny{\texttt{TabletopReachM}}}
\end{subfigure}
\begin{subfigure}{0.16\textwidth}
    \includegraphics[width=\linewidth]{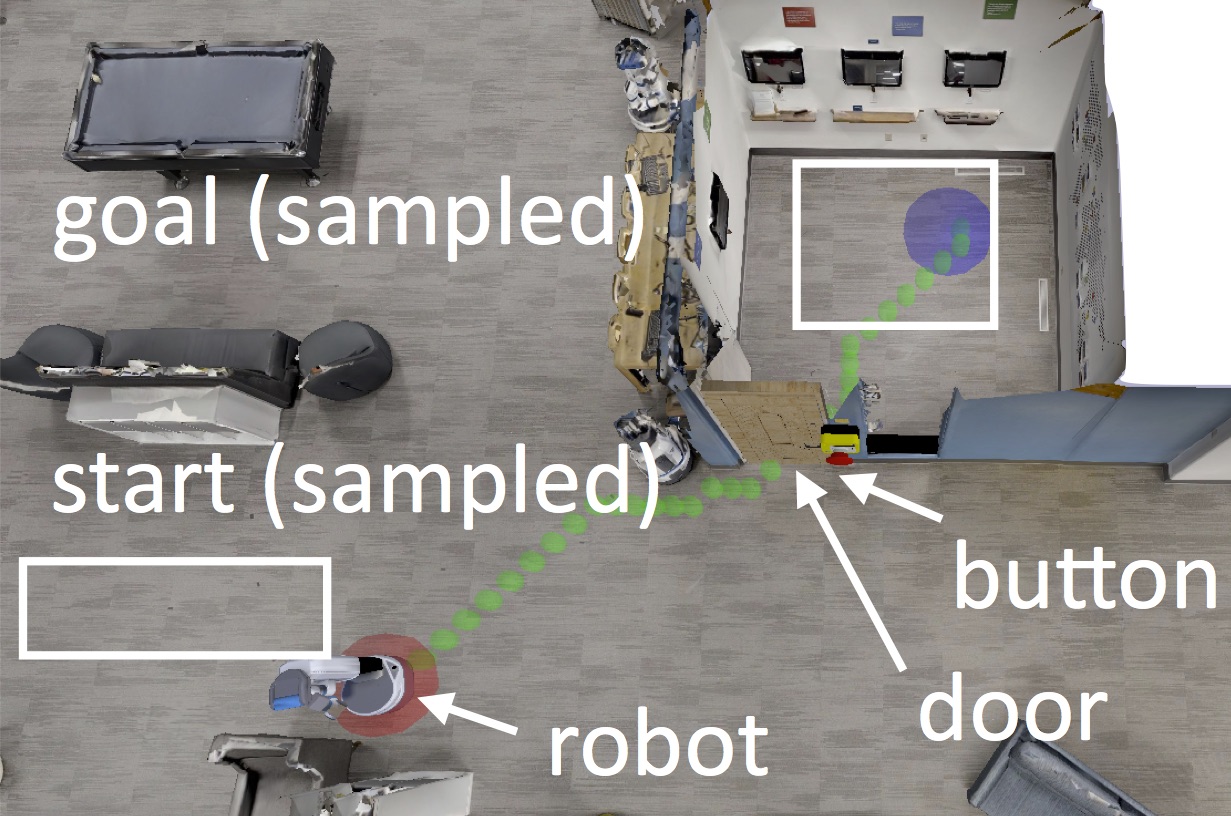}
    \caption{\tiny{\texttt{Push/ButtonDoorNav}}}
\end{subfigure}
\end{center}
\begin{center}
\begin{subfigure}{0.16\textwidth}
    \includegraphics[width=\linewidth]{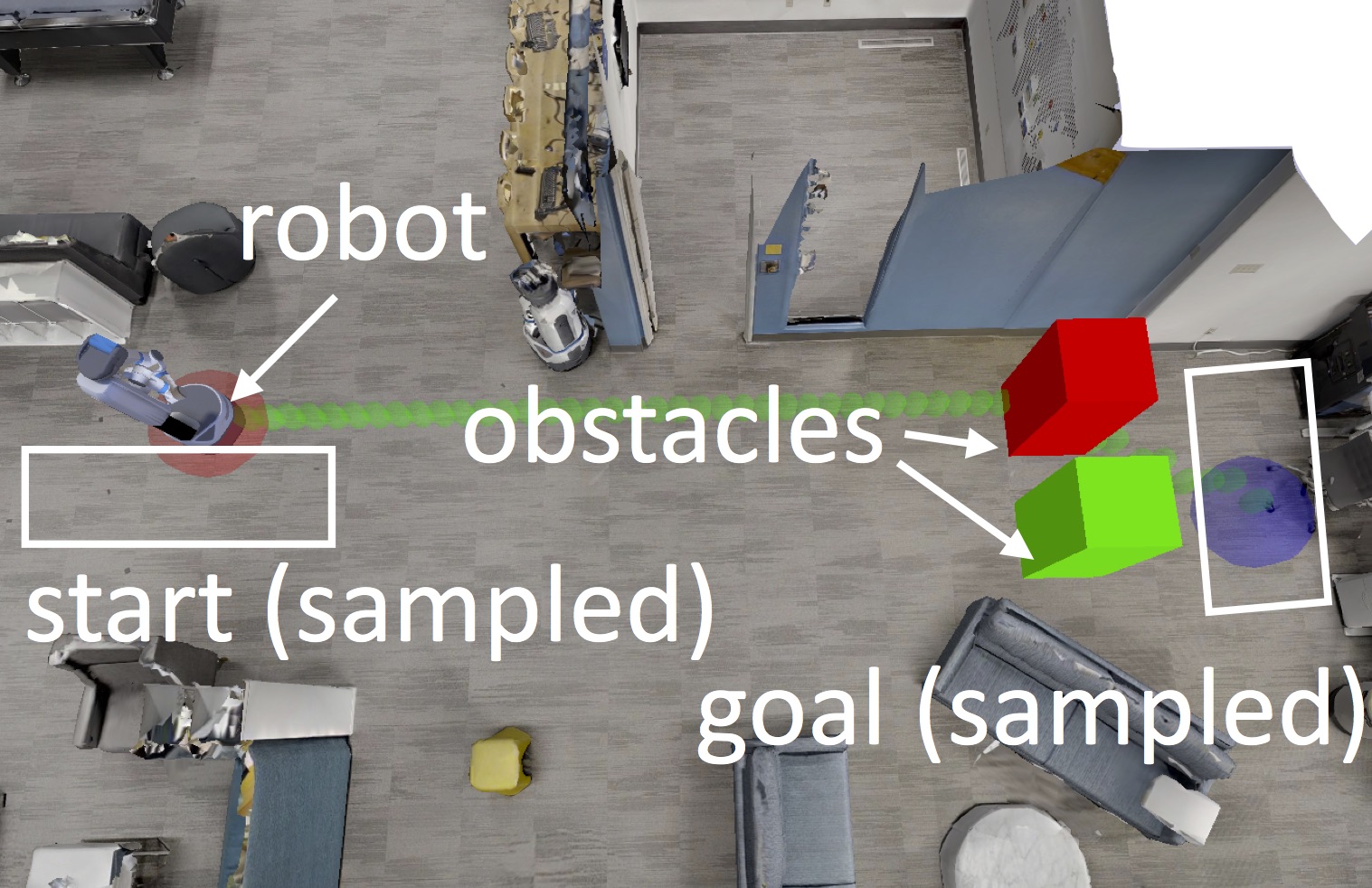}
    \caption{\tiny{\texttt{Int.ObstaclesNav}}}
\end{subfigure}
\begin{subfigure}{0.135\textwidth}
    \includegraphics[width=\linewidth]{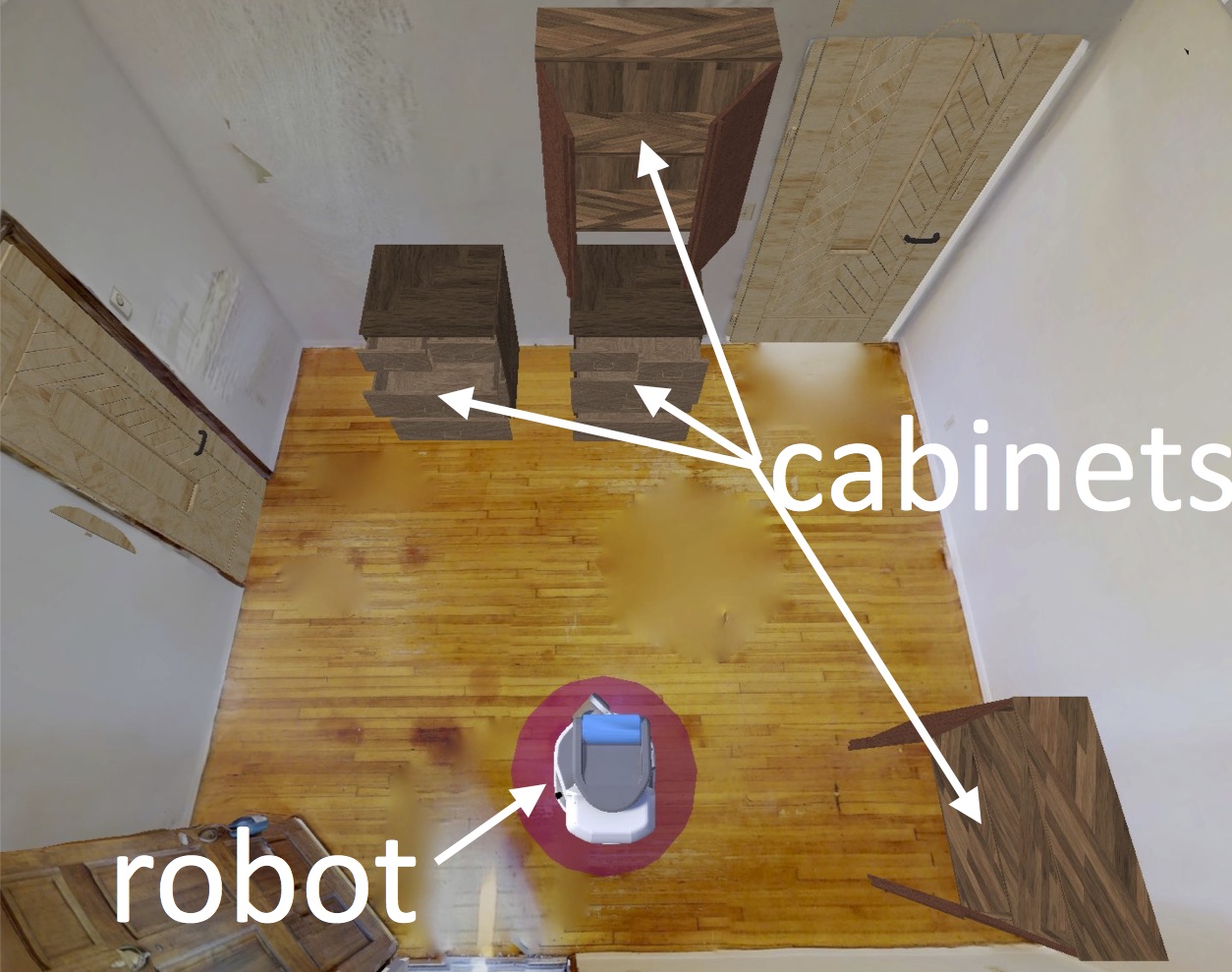}
    \caption{\tiny{\texttt{ArrangeKitchenMM}}}
\end{subfigure}
\begin{subfigure}{0.16\textwidth}
    \includegraphics[width=\linewidth]{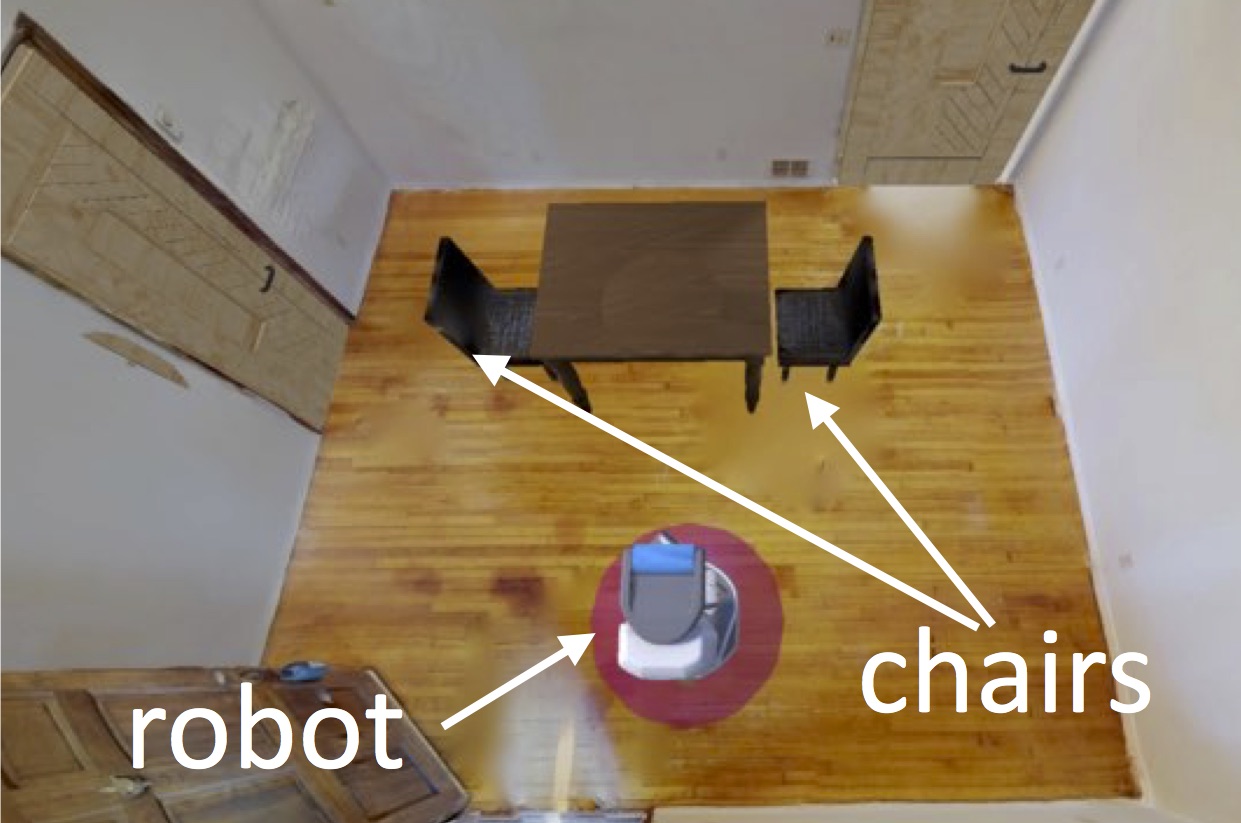}
    \caption{\tiny{\texttt{ArrangeChairMM}}}
\end{subfigure}
\end{center}

\caption{\small{The simulation environments and tasks. (a)(b) navigation-only and manipulation-only tasks, (c)(d) three Interactive Navigation tasks, (e)(f) two Mobile Manipulation tasks.}}
\label{fig:sim_env}
\vspace{-4mm}

\end{figure}

\textbf{Observations:} Our subgoal generation policy, $\pi_\mathit{SGP}$, takes in sensor inputs and outputs MG subgoals (see Fig.~\ref{fig:method}). We assume three common sensor sources (an RGB image and a depth map from a robot's RGB-D camera, and a single-beam LiDAR scan), and, optionally, additional task information. For navigation and Interactive Navigation tasks, the task information is the final goal location together with the next $N$ waypoints separated $d$ meters apart on the shortest path to the final goal, both relative to the current robot's pose ($N=10$ and $d=\SI{0.2}{\meter}$ in our experiments). We assume the goal and the shortest path are provided by the environment and computed based on a floor plan that contains only stationary elements (e.g. walls), regardless of dynamic objects such as doors, and obstacles (see Fig.~\ref{fig:sim_env}). For mobile manipulation (MM) tasks, there is no additional task information.

\textbf{Continuous Action Parameterization Method - SGP-R:}
We call our subgoal generation policy for continuous action parameterization SGP-R, where ``R'' indicates regression. We denote this implementation of ReLMoGen as ReLMoGen-R. The high-level idea is to treat the space of subgoals as a continuous action space, in which the policy network predicts (regresses) one vector. Based on the observation, the policy outputs 1) a base subgoal: the desired base 2D location in polar coordinates and the desired orientation change, 2) an arm subgoal: the desired end-effector 3D location represented by a $(u, v)$ coordinate on the RGB-D image to initiate the interaction, and a 2D interaction vector relative to this position that indicates the final end-effector position after the interaction, and 3) a binary variable that indicates whether the next step is a base-motion or an arm-motion phase (see Fig.~\ref{fig:method_architecture_r}). These subgoals are executed by the motion generator introduced in the Section~\ref{subsection:mp}. We train SGP-R using Soft Actor-Critic~\cite{haarnoja2018soft}. 

\textbf{Discrete Action Parameterization Method - SGP-D:}
We call our subgoal generation policy for discrete action parameterization SGP-D, where ``D'' indicates dense prediction. We denote this implementation of ReLMoGen as ReLMoGen-D. This parameterization aligns the action space with the observation space, and produces dense Q-value maps. The policy action (subgoal) corresponds to the pixel with the maximum Q-value. This parametrization is similar to the ``spatial action maps'' by~\citet{wu2020spatial}. Unlike their policy, our SGP-D predicts two types of action maps: one for base subgoals spatially aligned with the local map and the other for arm subgoals spatially aligned with the RGB-D image from the head camera (see Fig.~\ref{fig:method_architecture_d}). 
To represent the desired orientation of the base subgoal, we discretize the value into $L$ bins per pixel for the base Q-value maps. Similarly, for the desired pushing direction of the arm subgoal, we have $K$ bins per pixel for the arm Q-value maps ($K = L = 12$ in our experiments). We train SGP-D using Deep Q-learning~\cite{mnih2013playing}.

\vspace{-2mm}
\subsection{Motion Generation for Base and Arm}
\label{subsection:mp}
We use a motion generator to lift the action space for robot learning from low-level motor actuation to high-level ``subgoals''. The motion generator consists of two modules: 1) a motion planner that searches for trajectories to a given subgoal using a model of the environment created based on current sensor information, and 2) a set of common low-level controllers (joint space) that execute the planned trajectories. In our solution, we use a bidirectional rapidly-exploring random tree (RRT-Connect)~\cite{kuffner2000rrt} to plan the motion of the base and the arm, although we also experiment with probabilistic road-maps (PRM)~\cite{bohlin2000path} in our evaluation.

The motion planner for the base is a 2D Cartesian space RRT that searches for a collision-free path to the base subgoal location on the local map generated from the most recent LiDAR scan. The base subgoals are represented as the desired base 2D locations and orientations.

The motion planner for the arm comprises a 3D Cartesian space RRT and a simple Cartesian space inverse kinematics (IK) based planner. The arm motion is made of two phases: 1) the motion from the initial configuration to the selected subgoal location, and 2) the pushing interaction starting from the subgoal location. For the first phase, the 3D RRT searches for a collision-free path to reach the subgoal location. 
If the first phase succeeds, as the second phase, the simple IK-based planner is queried to find a sequence of joint configurations to move the end effector in a straight line from the subgoal location along the specified pushing direction. 
Since the intent of the second phase is to interact with the environment, the path is not collision-free. The arm subgoals are thus represented as the desired end-effector 3D locations and parameterized pushing actions. We hypothesize that the pushing actions can be replaced by other types of parameterized actions (e.g. grasping and pulling). More details about algorithm description, network structure, training procedure and hyperparameters can be found on our website.




\begin{figure*}[!t]
\centering
\begin{subfigure}{0.226\textwidth}
    \includegraphics[width=\textwidth]{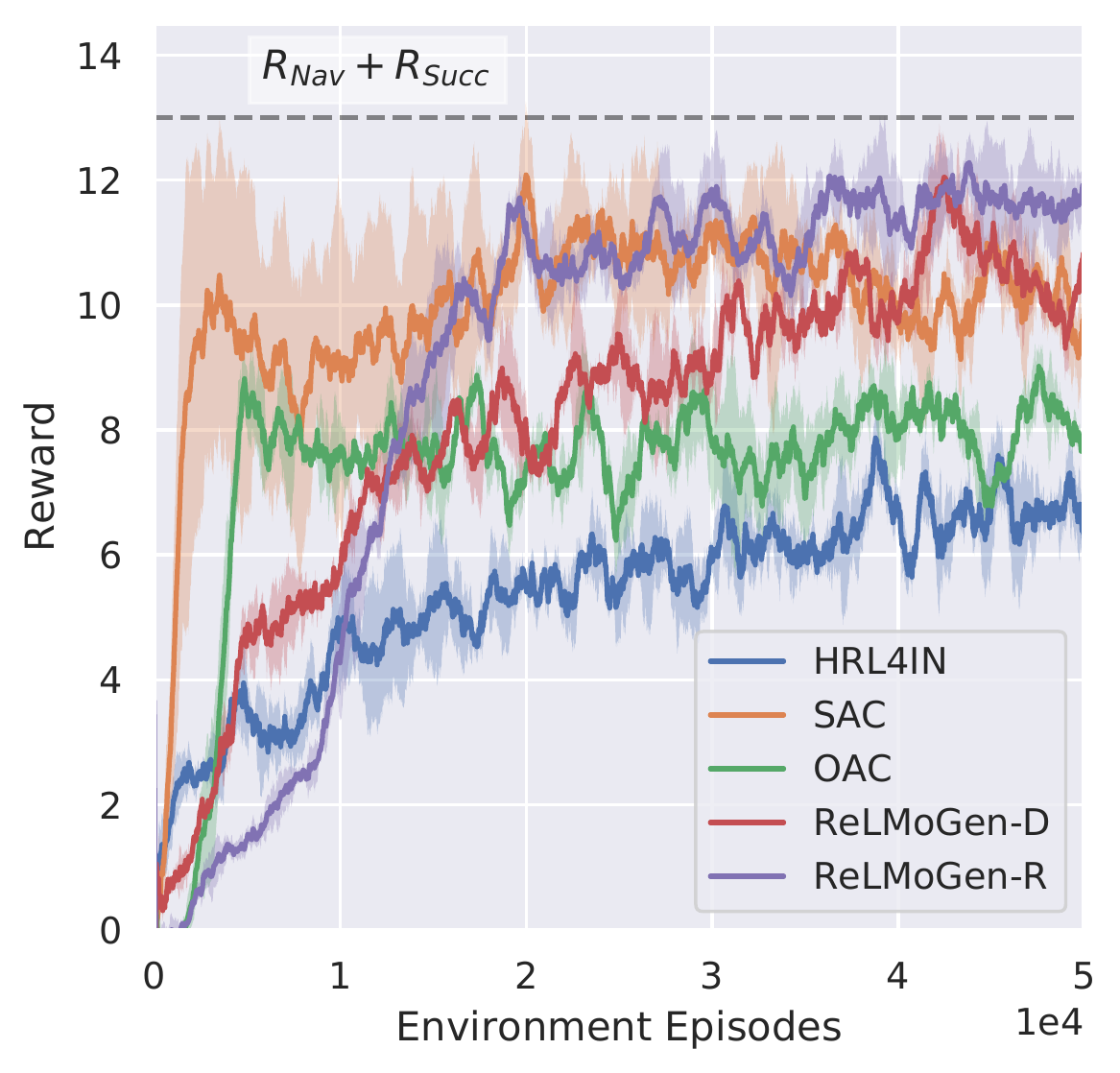}
    \caption{\small{\texttt{PointNav}}}
\end{subfigure}
\begin{subfigure}{0.2305\textwidth}
    \includegraphics[width=\textwidth]{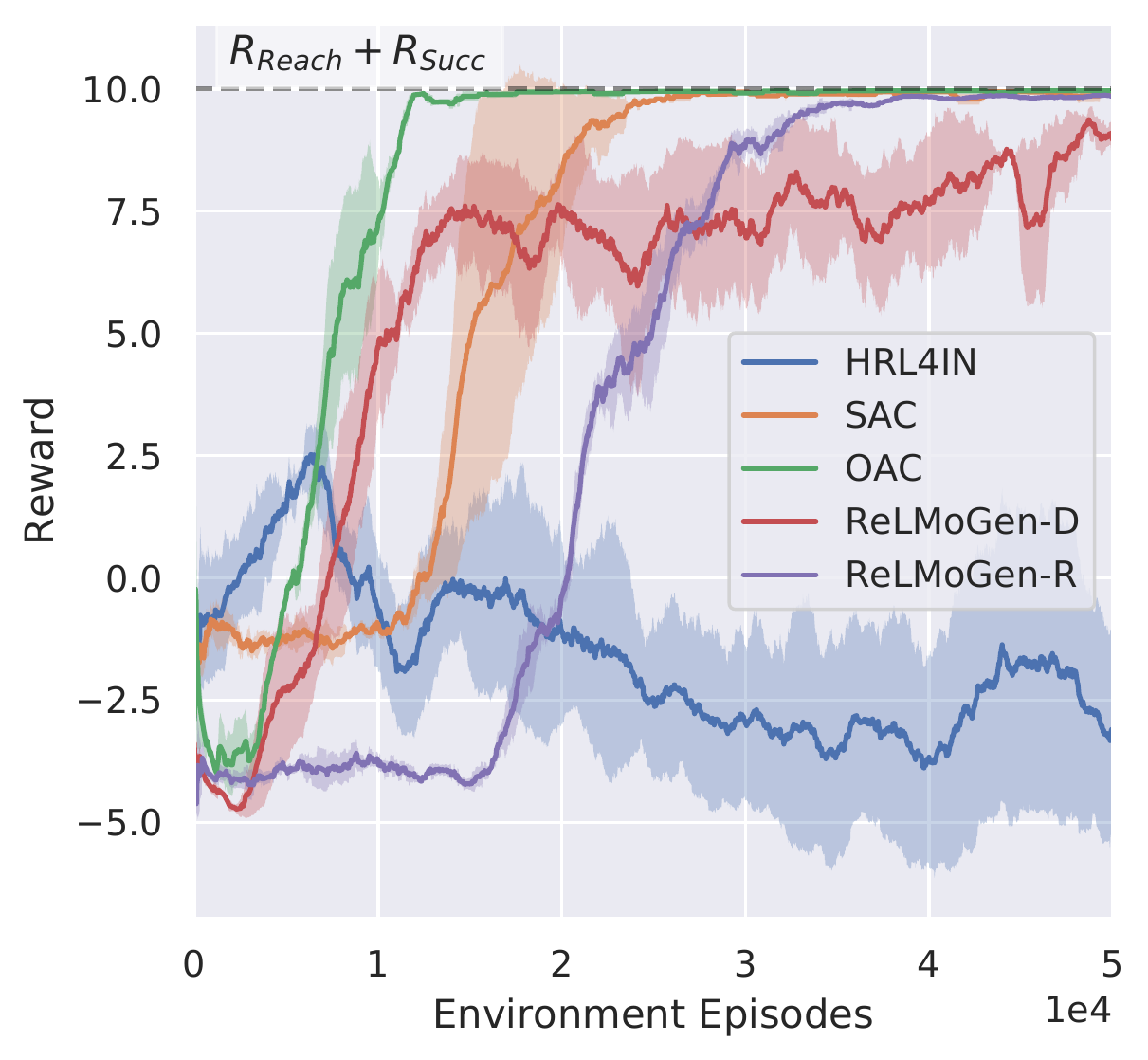}
    \caption{\small{\texttt{TabletopReachM}}}
\end{subfigure}
\begin{subfigure}{0.2278\textwidth}
    \includegraphics[width=\textwidth]{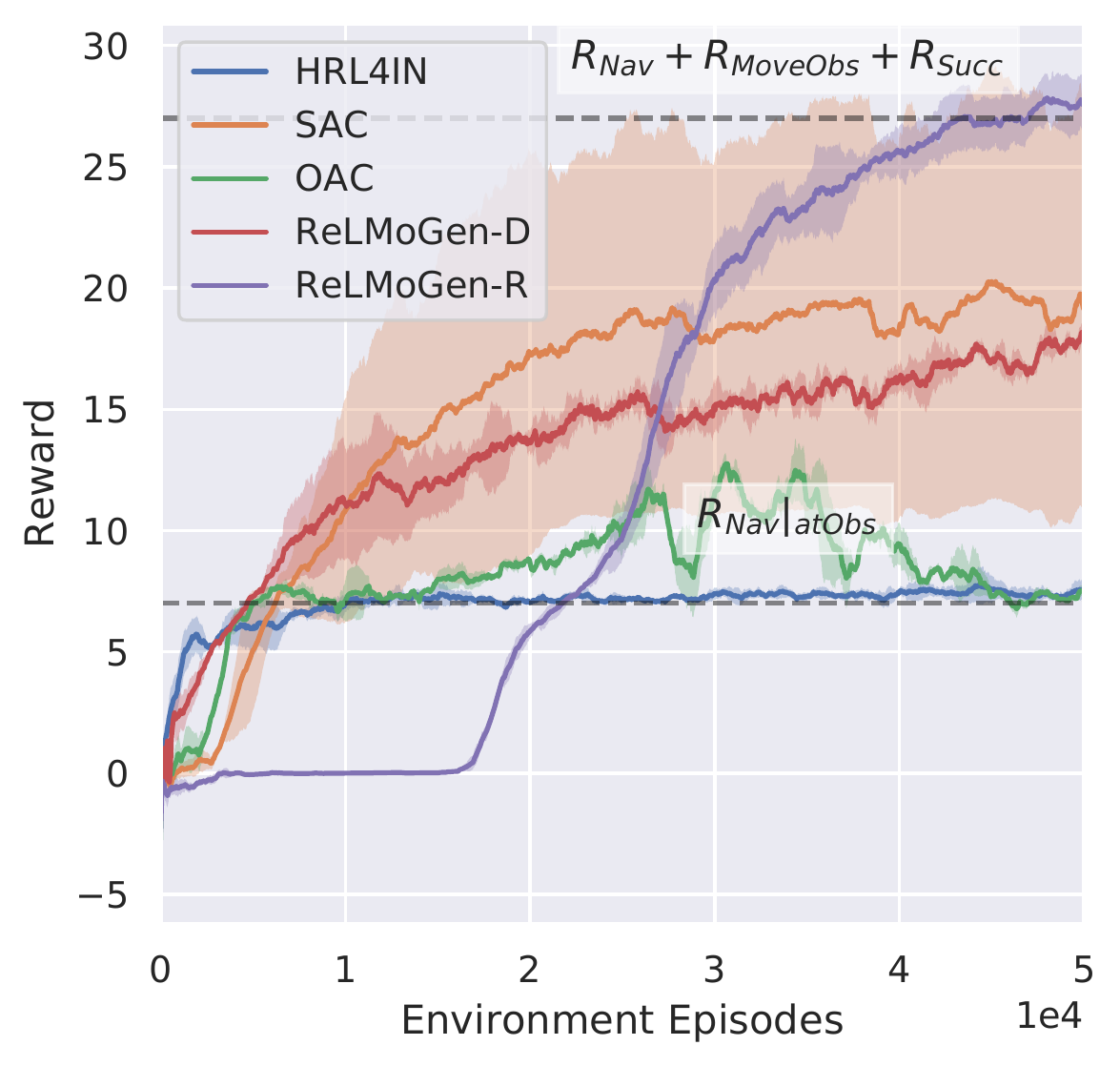}
    \caption{\small{\texttt{Int.ObstaclesNav}}}
\end{subfigure}
\begin{subfigure}{0.2305\textwidth}
    \includegraphics[width=\textwidth]{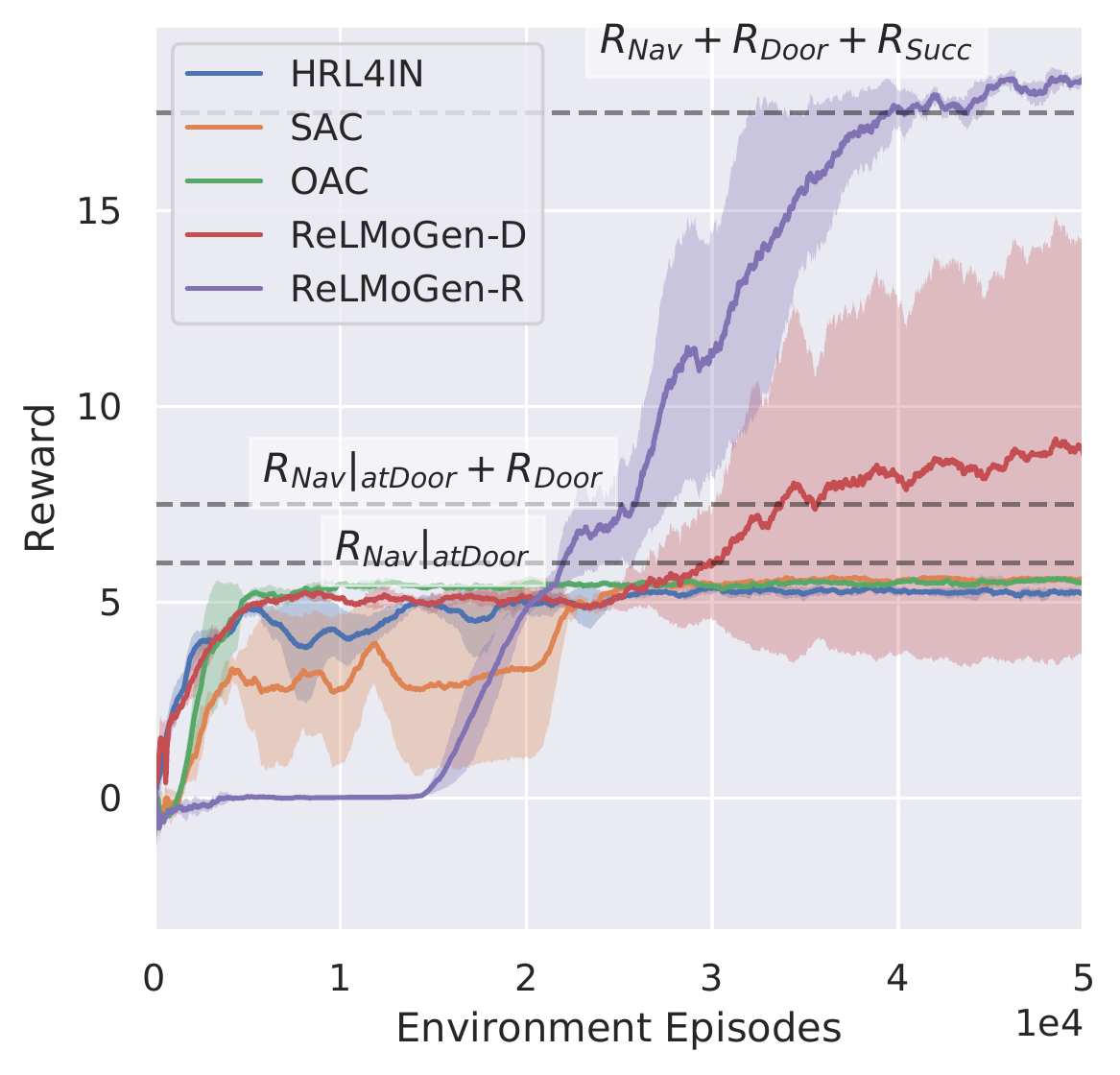}
    \caption{\small{\texttt{PushDoorNav}}}
\end{subfigure}
\begin{subfigure}{0.289\textwidth}
    \includegraphics[width=\textwidth]{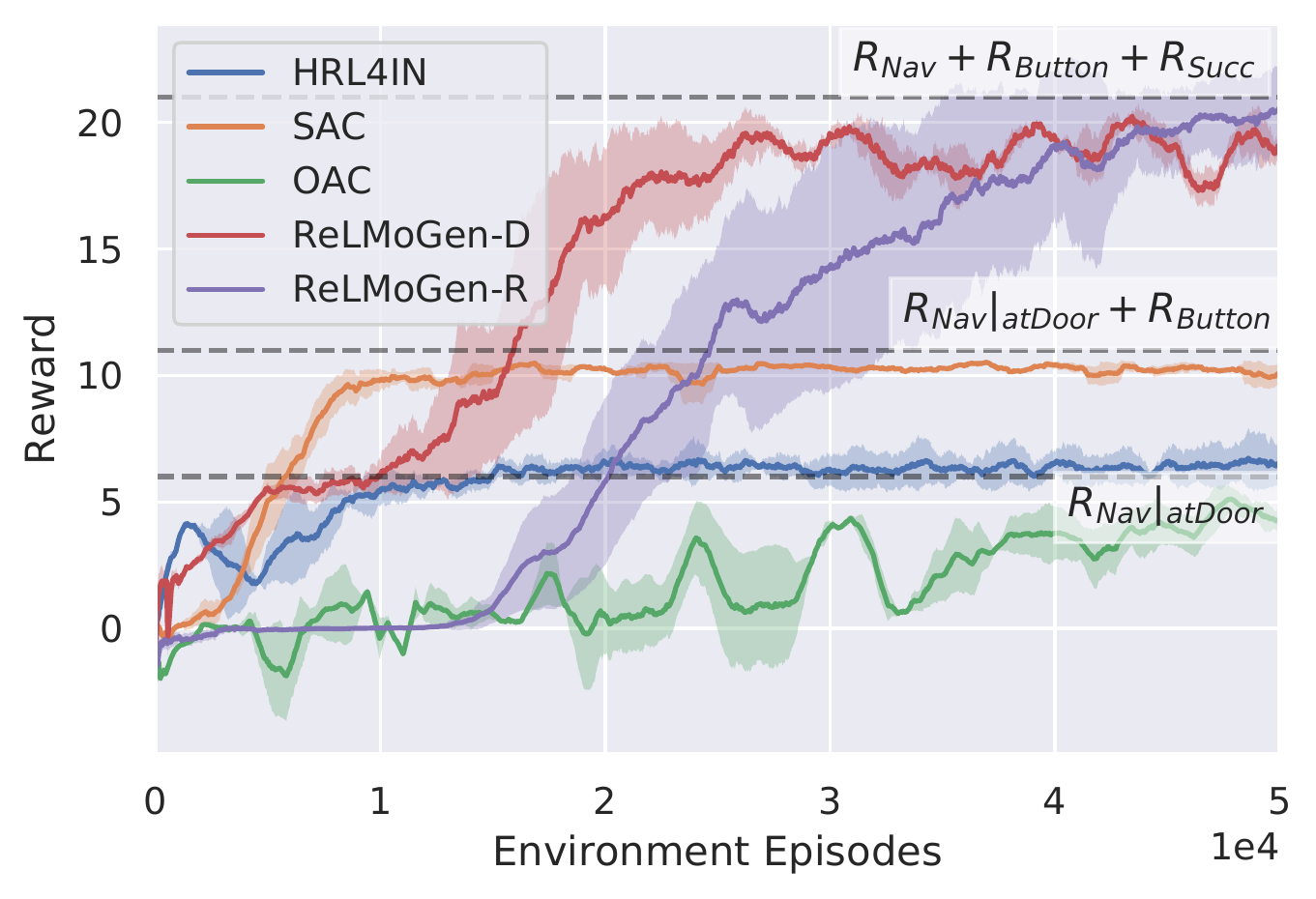}
    \caption{\small{\texttt{ButtonDoorNav}}}
\end{subfigure}
\begin{subfigure}{0.289\textwidth}
    \includegraphics[width=\textwidth]{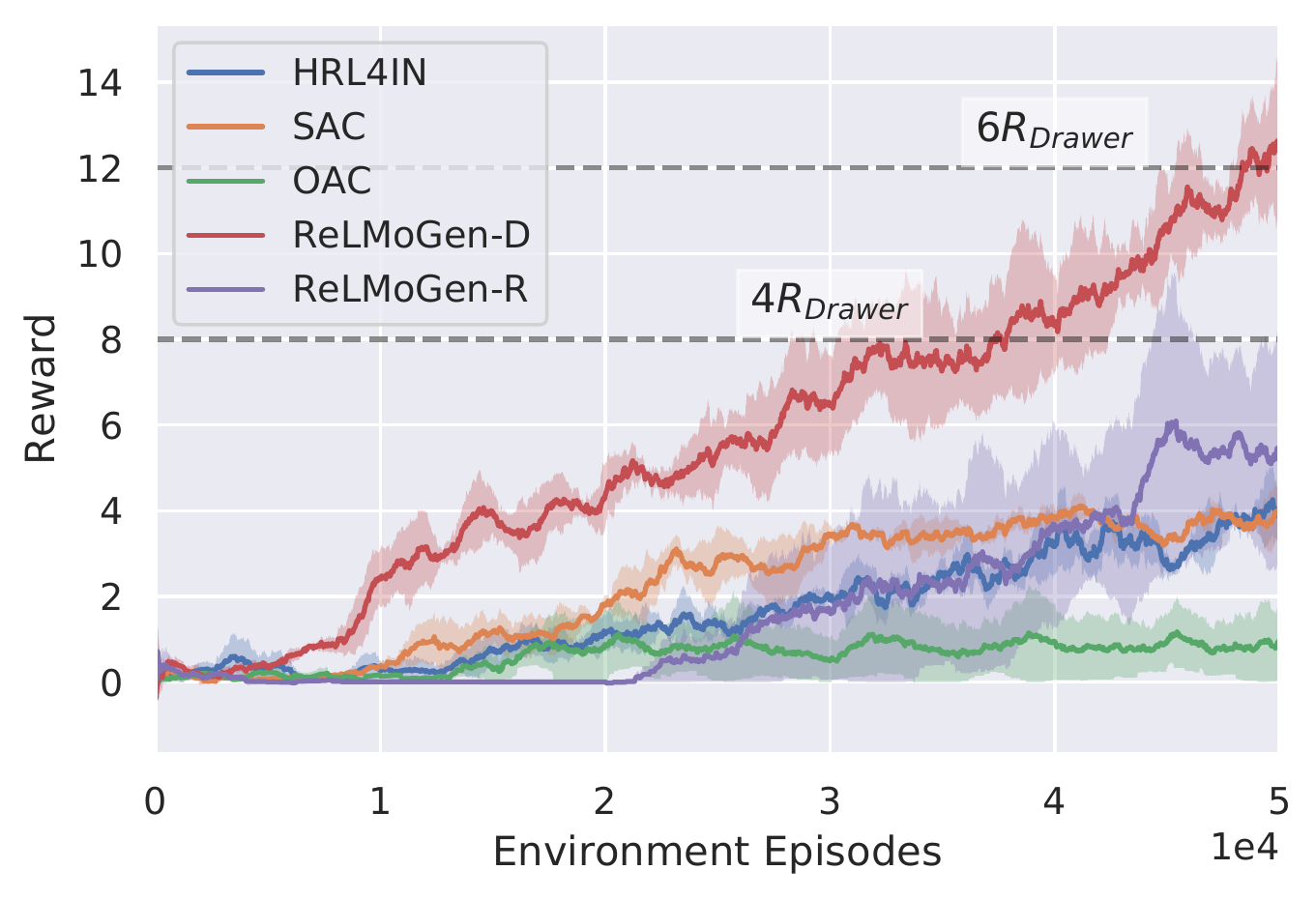}
    \caption{\small{\texttt{ArrangeKitchenMM}}}
\end{subfigure}
\begin{subfigure}{0.289\textwidth}
    \includegraphics[width=\textwidth]{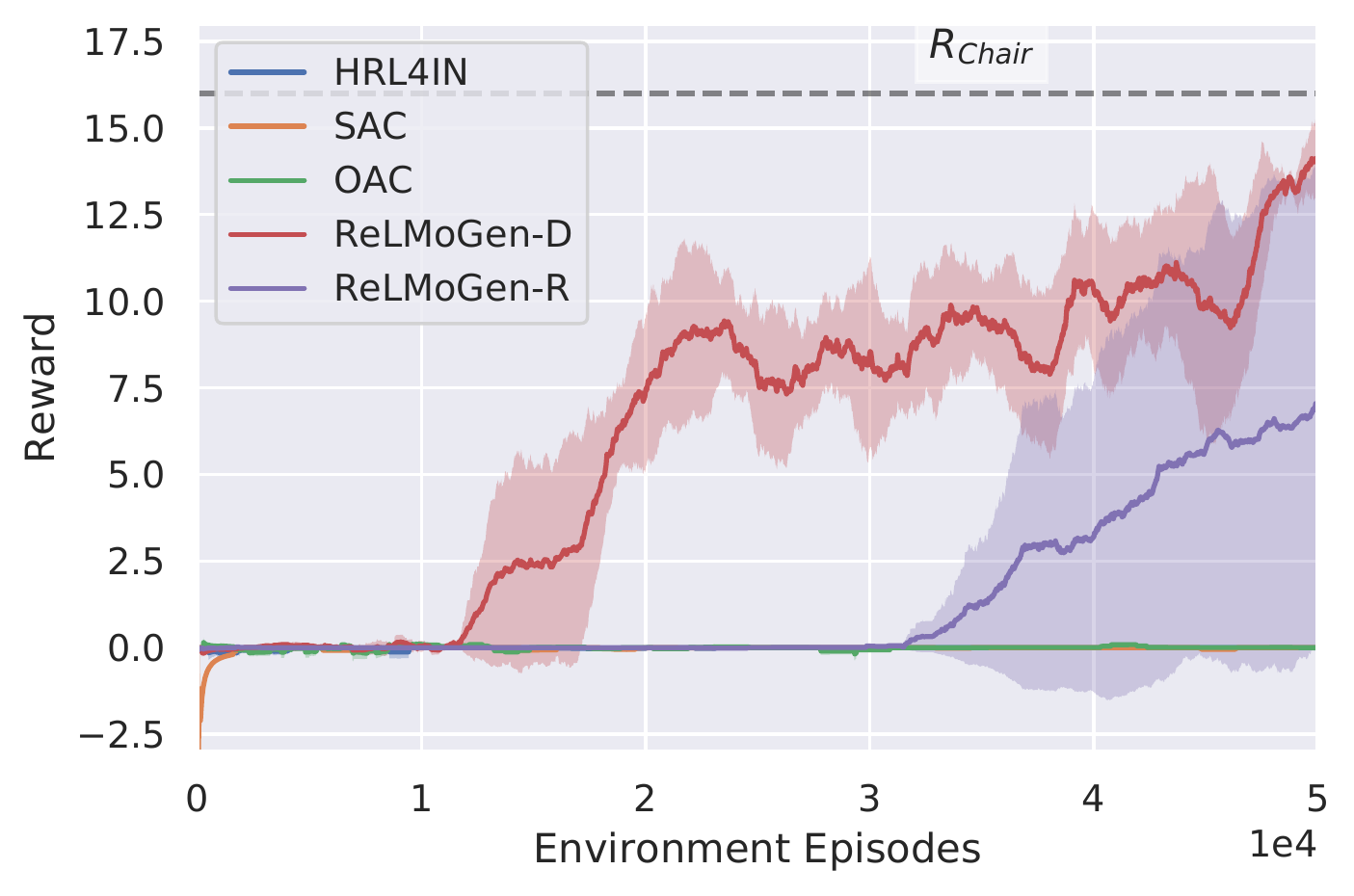}
    \caption{\small{\texttt{ArrangeChairMM}}}
\end{subfigure}

\caption{\small{Training curves for ReLMoGen and the baselines (SAC, OAC, and HRL4IN). ReLMoGen achieves higher reward with the same number of environment episodes and  higher task completion for all seven tasks while the baselines often converge to sub-optimal solutions. The curve indicates the mean and standard deviation of the return across three random seeds. Note that the x-axis indicates environment episodes rather than steps to allow for a fair comparison between solutions that use actions with different time horizons.}}
\label{fig:reward}
 \vspace{-6mm}
\end{figure*}


\section{Experimental Evaluation}\label{s_exp}


We evaluate our method on seven different tasks. These tasks include navigation, manipulation, Interactive Navigation, and Mobile Manipulation (see Fig.~\ref{fig:sim_env}). We believe these tasks represent paradigmatic challenges encountered by robots operating in realistic environments.


\textbf{Navigation-Only and Manipulation-Only Tasks:} PointGoal navigation \cite{anderson2018evaluation, habitat19arxiv} and tabletop tasks \cite{zamora2016extending} are mature robotic benchmarks. In \texttt{PointNav}, the robot needs to move to a goal without collision. In \texttt{TabletopReachM}, the robot needs to touch a point on the table with its end-effector. 

\textbf{Interactive Navigation (IN) Tasks:} In these tasks the robot needs to interact with the environment to change the environment's state in order to facilitate or enable navigation~\cite{xia2020interactive}. In \texttt{PushDoorNav} and \texttt{ButtonDoorNav}, the robot needs to enter a room behind a closed door, by pushing the door or pressing a button, respectively. In \texttt{InteractiveObstaclesNav} task, the robot is blocked by two objects and needs to push them aside to reach the goal. Only one of the objects can be pushed, and the agent needs to judge solely based on visual appearance (color). These tasks require the robot to place its base properly to interact with the objects \cite{berenson2008optimization, klingbeil2010learning}, and to infer where to interact based on a correct interpretation of the RGB-D camera information (e.g. finding the door button).

\textbf{Mobile Manipulation (MM) Tasks:} These are long-horizon tasks known to be difficult for RL~\cite{wang2020learning,li2020hrl4in}, making it a good test for our method. We created two MM tasks, \texttt{ArrangeKitchenMM} and \texttt{ArrangeChairMM}. In \texttt{ArrangeKitchenMM}, the robot needs to close cabinet drawers and doors randomly placed and opened. The challenge is that the robot needs to find the cabinets and drawers using the RGB-D information, and accurately actuate them along their degrees of freedom. In \texttt{ChairArrangeMM}, the robot needs to push chairs under a table. The opening under the table is small so the push needs to be accurate. Object locations are unknown to the robot. Both tasks can be thought of as an ObjectNav~\cite{anderson2018evaluation} task followed by a manipulation task. The reward is only given when the robot makes progress during the manipulation phase.

All experiments are conducted in iGibson Environment~\cite{xia2020interactive}. The Navigation and Interaction Navigation tasks are performed in a 3D reconstruction of an office building. The Mobile Manipulation and Tabletop tasks are performed in a model of a residential house (\texttt{Samuels}) from \cite{xia2020interactive}, populated with furniture from Motion Dataset~\cite{wang2019shape2motion} and ShapeNet Dataset~\cite{chang2015shapenet}. We randomize the initial pose of the robot, objects and goals across training episodes so that the agent cannot simply memorize the solution. 


For (Interactive) Navigation tasks, we have dense reward, $R_{Nav}$, that encourages the robot to minimize the geodesic distance to the goal, and success reward, $R_{Succ}$, for task completion. We have bonus reward for the robot to push obstacles, doors and buttons, denoted as $R_{MoveObs}$, $R_{Door}$ and $R_{Button}$. For Mobile Manipulation tasks, we have dense reward for the robot to close drawers and cabinets, or to tuck chairs, denoted as $R_{Drawer}$ and $R_{Chair}$. We don't provide reward for the robot to approach these objects. Episodes terminate when any part of the robot body other than the gripper collides with the environment. More detailed reward definition and evaluation metrics are on our website. 

\vspace{-1mm}

\subsection{Baselines}\label{sec:baselines}


 \textbf{SAC (on joint velocities):} We run SAC~\cite{haarnoja2018soft} directly on joint velocities for all the joints on our robot (2 wheels, 1 torso joint, 7 arm joints), similar to previous work on visuomotor control~\cite{levine2016end}.
 
 \textbf{OAC (on joint velocities):} We run a variant of SAC called OAC presented by \citet{ciosek2019better}. This work applies the principle of optimism in face of uncertainty to Q-functions and outperform SAC in several continuous control tasks~\cite{ciosek2019better}.
 
 \textbf{HRL4IN:} We run the hierarchical RL algorithm presented by \citet{li2020hrl4in}. This work shows good performance for IN tasks. Similar to ours, a high-level policy produces base and arm subgoals and a variable to decide the part of the embodiment to use. Different from ours, this method uses a learned low-level policy instead of a motion generator. With this baseline we evaluate the effect of integrating RL and MG instead of learning a low-level policy from scratch.


The action space of ReLMoGen and the baselines have drastically different time horizons. For fair comparison, we set the episode length to be roughly equivalent in wall-clock time of simulation across algorithms: 25 subgoal steps for ReLMoGen and 750 joint motor steps for the baselines. To evaluate performance, we use success rate and SPL~\cite{anderson2018evaluation} for navigation tasks, and task completion (number of drawers/cabinets closed, chairs tucked within $\SI{10}{\degree}/\SI{10}{cm}$ and $\SI{5}{\degree}/\SI{5}{cm}$) for mobile manipulation tasks.

\begin{table*}[t]
  \begin{center}
    
   \resizebox{2\columnwidth}{!}{
  \begin{tabular}{lcccccccc} 
  \toprule
  Task  & \multicolumn{2}{c}{\texttt{PointNav}} & \multicolumn{1}{c}{\texttt{TabletopReachM}} & \multicolumn{2}{c}{\texttt{ArrangeKitchenMM}} & \multicolumn{2}{c}{\texttt{ArrangeChairMM}}  \\ 
  \midrule
  Metric  & SPL & SR & SR & \# Closed $\SI{5}{\degree}/\SI{5}{cm}$ &  \# Closed  $\SI{10}{\degree}/\SI{10}{cm}$ & \# Closed  $\SI{5}{cm}$ & \# Closed $\SI{10}{cm}$ \\
  \midrule
     ReLMoGen-D (ours) &  $0.57/0.02/0.58$ & $0.68/0.01/0.68$ &  $0.95/0.02/0.96$ & $\bf{4.35/1.20/5.72}$ & $\bf{6.10/1.05/7.3}$ & $\bf{0.21/0.03/0.23}$ & $\bf{0.36/0.06/0.43}$ \\
    ReLMoGen-R (ours) & $\bf{0.63/0.09/0.67}$ & $\bf{0.72/0.06/0.77}$ & $\bf{1.0/0.0/1.0}$ & $3.43/0.61/3.94$ & $4.91/0.51/5.25$ & $0.06/0.10/0.17$ & $0.11/0.20/0.34$ \\
   HRL4IN~\cite{li2020hrl4in} & $0.27/0.01/0.28$ & $0.33/0.01/0.35$ & $0.09/0.07/0.19$ & $3.0/0.23/3.3$ & $4.67/0.20/4.95$ & $0.0/0.0/0.0$ & $0.0/0.0/0.0$ \\
   SAC (joint vel.)~\cite{haarnoja2018soft, levine2016end} & $0.60/0.04/0.65 $& $0.60/0.04/0.65$ & $\bf{1.0/0.0/1.0}$ & $3.42/0.19/3.6$ & $4.95/0.29/5.24$ & $0.0/0.0/0.0$& $0.0/0.0/0.0$ \\
   OAC (joint vel.)~\cite{ciosek2019better} & $0.45/0.01/0.46$ & $0.46/0.01/0.47$ & $\bf{1.0/0.0/1.0}$ & $1.99/0.61/2.60$ &  $3.55/ 0.48/4.02$ & $0.0/0.0/0.0$ & $0.0/0.0/0.0$  \\
   \bottomrule
  \end{tabular}
  }
  
   \vspace{2mm}
  
   \resizebox{1.8\columnwidth}{!}{
  \begin{tabular}{lcccccccccc} 
  \toprule
  Task  & \multicolumn{2}{c}{\texttt{PushDoorNav}} & \multicolumn{2}{c}{\texttt{ButtonDoorNav}} & \multicolumn{2}{c}{\texttt{InteractiveObstaclesNav}} \\
  \midrule
  Metric  & SPL & SR& SPL & SR & SPL & SR\\
  \midrule
     ReLMoGen-D (ours) & $0.36/0.36/0.72$ & $0.41/0.40/0.80$ & $0.42/0.17/0.57$ & $0.50/0.19/0.66$ & $0.54/0.011/0.55$ & $0.58/0.02/0.60$  \\
    ReLMoGen-R (ours) & $\bf{0.80/0.02/0.83}$ & $\bf{0.97/0.02/0.99}$ & $\bf{ 0.51/0.15/0.61}$ & $\bf{0.73/0.21/0.87}$ & $\bf{0.76/0.01/0.87}$ & $\bf{0.79/0.11/0.91}$  \\
   HRL4IN~\cite{li2020hrl4in} & $0.0/0.0/0.0$ & $0.0/0.0/0.0$ & $0.0/0.0/0.0$ & $0.0/0.0/0.0$ & $0.0/0.0/0.0$ & $0.0/0.0/0.0$\\
   SAC (joint vel.)~\cite{haarnoja2018soft, levine2016end}  & $0.0/0.0/0.0$ & $0.0/0.0/0.0$ & $0.00/0.01/0.01$ & $0.01/0.01/0.01$ &  $0.50/0.36/0.84$ & $0.51/0.37/0.87$ \\
   OAC (joint vel.)~\cite{ciosek2019better} & $0.0/0.0/0.0$ & $0.0/0.0/0.0$ & $0.00/0.00/0.01$ & $0.01/0.00/0.01$ &  $0.00/0.00/0.01$ & $0.01/0.01/0.01$ \\
   \bottomrule
  \end{tabular}
  }


   \caption{\small{Task completion metrics for two version of ReLMoGen, one using DQN with discrete subgoal parameterization (ReLMoGen-D) and one using SAC with continous subgoal parameterization (ReLMoGen-R).  We compare with two baselines (see Sec.~\ref{sec:baselines}). The entries of this table are in the format of mean/std/max over 3 random seeds and the method with the highest mean value is highlighted in bold.\label{tbl:success}}}
  \end{center}
  \vspace{-7mm}
\end{table*}

\subsection{Analysis}
\label{subsec:res}
\edit{We aim at answering the following research questions with our analysis in this subsection.}

\textbf{Can ReLMoGen solve a wide variety of robotic tasks involving navigation and manipulation?} In Table~\ref{tbl:success}, we show the task completion metrics across all tasks for our methods and baselines. In a nutshell, our method achieves the highest performance across all seven tasks. It also exhibits better sample efficiency than our baselines (see Fig.~\ref{fig:reward}). 

SAC and OAC baseline have comparable performance to our methods for simpler tasks such as \texttt{PointNav} and \texttt{TabletopReachM} but fail completely for harder ones, such as \texttt{PushDoorNav} and \texttt{ChairArragementMM}, due to collisions or their inability to identify objects that are beneficial to interact with. OAC only outperforms SAC with a small margin in one task, which suggests that it remains an open research question on how to conduct deep exploration in robotics domain with high dimensional observation space and continuous action space. To our surprise, HRL4IN baseline perform worse than SAC baseline for several tasks. This is potentially caused by our deviation from the original task setup in \cite{li2020hrl4in} since we do not allow collisions with the robot base during exploration, \edit{while HRL4IN has a collision prone low-level policy.} This is consistent with our insight that using MG instead of a learned low-level policy makes it easier to train the subgoal generation policy, and that RL is best suited to learn the mapping from observations to subgoals. 

One common failure case for the baselines in IN tasks is that the agent harvests all the navigation reward by approaching the goal but gets stuck in front of doors or obstacles, failing to learn meaningful interaction with them. On the other hand, both our ReLMoGen implementations with SGP-R and SGP-D are able to achieve significant success in tasks that involve precise manipulation (e.g. \texttt{ButtonDoorNav}), intermittent reward signal (e.g. \texttt{ArrangeChairMM} and \texttt{ArrangeKitchenMM}) and alternative phases of base and arm motion (all IN and MM tasks). \edit{Empirically, ReLMoGen-D outperforms ReLMoGen-R for tasks that involve more fine-grained manipulation due to its Q-value estimation at every single pixel, but seems to be less sample efficient than it for tasks that only require coarse manipulation.} We argue that the main advantage of ReLMoGen is that it explores efficiently while maintaining high ``subgoal success rates'' thanks to its embedded motion generators, resulting in stable gradients during training. As a bonus, ReLMoGen performs an order of magnitude fewer gradient updates than the baselines, which translates to a much shorter wall-clock time for training \edit{(on average 7x times faster)}. Finally, our ReLMoGen-D model outputs highly interpretable Q-value maps: high Q-value pixels correspond to rewarding interactions, such as buttons, cabinet doors and chair backs. More visualizations can be found on our website.

\begin{figure}[!t]
\begin{center}
\begin{subfigure}{0.15\textwidth}
    \includegraphics[width=\textwidth]{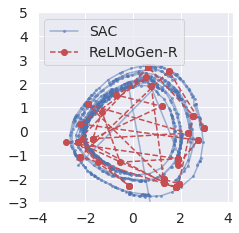}
    \caption{\small{Latent Space}}
\end{subfigure}
\begin{subfigure}{0.162\textwidth}
    \includegraphics[width=\textwidth]{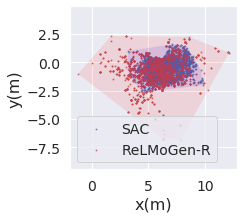}
    \caption{\small{Cartesian Space}}
\end{subfigure}
\begin{subfigure}{0.15\textwidth}
    \includegraphics[width=\textwidth]{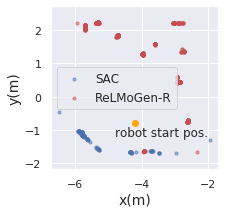}
    \caption{\small{Interaction map}}
\end{subfigure}
\caption{\small{Exploration of ReLMoGen-R and SAC. (a) shows the 2D projection of latent state space: SAC traverses nearby states with low-level actions, while ReLMoGen-R jumps between distant states linked by a motion plan. (b) shows the physical locations visited by ReLMoGen-R and SAC in 100 episodes: ReLMoGen-R covers a much larger area. (c) shows a top-down map of meaningful interactions (duration $\ge$1s) during exploration. ReLMoGen-R is able to interact with the environment more than SAC. }}
\label{fig:exploration}
\end{center}
\vspace{-4mm}
\end{figure}

\begin{table}[t]
    \begin{center}
  \begin{subfigure}{0.22\textwidth}
  \resizebox{\textwidth}{!}{
  \begin{tabular}{lccc} 
  \toprule
    Base MP & Arm MP & Success rate\\ 
  \midrule
    \textbf{RRT-Connect} & \textbf{RRT-Connect} & $\bf{0.99}$ \\
    RRT-Connect & Lazy PRM & $1.0$ $\bf{(+0.01)}$ \\
    Lazy PRM & RRT-Connect & $0.99$ $\bf{(+0.0)}$ \\
    Lazy PRM & Lazy PRM & $1.0$ $\bf{(+0.01)}$ \\
    \bottomrule
  \end{tabular}
  }
 \caption{PushDoorNav Task}
\end{subfigure}
\hfill
 \begin{subfigure}{0.25\textwidth}
 \resizebox{\textwidth}{!}{
  \begin{tabular}{lccc} 
  \toprule
    Base MP & Arm MP & \# Closed ($\SI{10}{\degree}/\SI{10}{cm}$)\\ 
  \midrule
    \textbf{RRT-Connect} & \textbf{RRT-Connect} & $\bf{5.25}$ \\
    RRT-Connect & Lazy PRM & $5.0$ $\bf{(-0.25)}$ \\
    Lazy PRM & RRT-Connect & $5.18$ $\bf{(-0.07)}$\\
    Lazy PRM & Lazy PRM & $5.09$ $\bf{(-0.16)}$\\
    \bottomrule
  \end{tabular}
  }
 \caption{ArrangeKitchenMM Task}
\end{subfigure}
 \caption{\small{Our policy trained with RRT-Connect as the motion planner for base and arm can perform equally well when changing to Lazy PRM at test time (the first row shows the training setup). \label{tbl:mp}}}
\end{center}
\vspace{-8mm}

\end{table}




\textbf{Is ReLMoGen better at exploration?}
Fig.~\ref{fig:exploration} shows the exploration pattern of a random policy for SAC baseline and for ReLMoGen-R. Specifically, we visualize the distribution of the states visited by the policy at the beginning of training. We project the neural network embedding of the visited states onto a 2D plane showing the first two principal components. For SAC and ReLMoGen-R, the trajectories of ten episodes are shown in Fig.~\ref{fig:exploration}(a). We can see that SAC baseline only travels between adjacent states in the feature space because it explores in \textbf{joint space} (considering wheels as joints). On the other hand, ReLMoGen can jump between distant states, as long as they can be connected by the motion generator, because it explores in \textbf{subgoal space}. The visited states by ReLMoGen are indicated in red dots connected with dashed lines. This is also evident when we plot the visited states in physical, Cartesian space in Fig.~\ref{fig:exploration}(b). \edit{From Fig.~}\ref{fig:exploration}\edit{(c), we can see ReLMoGen have more meaningful interactions with the environment during exploration than SAC.}

\textbf{Can ReLMoGen generalize to different types of motion planners?}
During training, we used RRT-Connect as our motion planner. We want to test whether our method can zero-shot generalize to a new motion planner, namely Lazy PRM~\cite{bohlin2000path}, during test time. \edit{We swapped base and/or arm motion planners and tried different parameters (e.g. number of trajectory optimization iterations) for our system, and observed minimal performance drop} (see Table.~\ref{tbl:mp}). Although different motion planners have different sampling schemas and timeout criteria, the subgoals generated by our policy can seamlessly transfer between them. This demonstrates strong practicality and flexibility of our approach. 






\section{Conclusion} 
\label{sec:conclusion}
We introduce ReLMoGen, a hierarchical framework that integrates classical motion generation with reinforcement learning to solve mobile manipulation tasks. ReLMoGen leverages the best from both worlds: learning complex subgoal prediction from high dimensional observations via RL and precise low-level action execution via MG. We demonstrate better task completion and higher training efficiency compared to other learning based approaches. The learned policies with ReLMoGen are also robust and can transfer to different motion planners after training.





\bibliographystyle{IEEEtranN}
\vskip-\parskip
\begingroup
\bibliography{references}

\begin{thebibliography}{62}
\providecommand{\natexlab}[1]{#1}
\providecommand{\url}[1]{#1}
\csname url@samestyle\endcsname
\providecommand{\newblock}{\relax}
\providecommand{\bibinfo}[2]{#2}
\providecommand{\BIBentrySTDinterwordspacing}{\spaceskip=0pt\relax}
\providecommand{\BIBentryALTinterwordstretchfactor}{4}
\providecommand{\BIBentryALTinterwordspacing}{\spaceskip=\fontdimen2\font plus
\BIBentryALTinterwordstretchfactor\fontdimen3\font minus
  \fontdimen4\font\relax}
\providecommand{\BIBforeignlanguage}[2]{{%
\expandafter\ifx\csname l@#1\endcsname\relax
\typeout{** WARNING: IEEEtranN.bst: No hyphenation pattern has been}%
\typeout{** loaded for the language `#1'. Using the pattern for}%
\typeout{** the default language instead.}%
\else
\language=\csname l@#1\endcsname
\fi
#2}}
\providecommand{\BIBdecl}{\relax}
\BIBdecl

\bibitem[LaValle(2006)]{lavalle2006planning}
S.~M. LaValle, \emph{Planning algorithms}.\hskip 1em plus 0.5em minus
  0.4em\relax Cambridge university press, 2006.

\bibitem[Siciliano and Khatib(2007)]{Siciliano:2007:SHR:1209344}
B.~Siciliano and O.~Khatib, \emph{Springer Handbook of Robotics}.\hskip 1em
  plus 0.5em minus 0.4em\relax Berlin, Heidelberg: Springer-Verlag, 2007.

\bibitem[Sutton and Barto(2018)]{sutton2018reinforcement}
R.~S. Sutton and A.~G. Barto, \emph{Reinforcement learning: An
  introduction}.\hskip 1em plus 0.5em minus 0.4em\relax MIT press, 2018.

\bibitem[Arulkumaran et~al.(2017)Arulkumaran, Deisenroth, Brundage, and
  Bharath]{arulkumaran2017deep}
K.~Arulkumaran, M.~P. Deisenroth, M.~Brundage, and A.~A. Bharath, ``Deep
  reinforcement learning: A brief survey,'' \emph{IEEE Signal Processing
  Magazine}, vol.~34, no.~6, pp. 26--38, 2017.

\bibitem[Zhu et~al.(2017)Zhu, Mottaghi, Kolve, Lim, Gupta, Fei-Fei, and
  Farhadi]{zhu2017target}
Y.~Zhu, R.~Mottaghi, E.~Kolve, J.~J. Lim, A.~Gupta, L.~Fei-Fei, and A.~Farhadi,
  ``Target-driven visual navigation in indoor scenes using deep reinforcement
  learning,'' in \emph{2017 IEEE international conference on robotics and
  automation (ICRA)}.\hskip 1em plus 0.5em minus 0.4em\relax IEEE, 2017, pp.
  3357--3364.

\bibitem[Quan et~al.(2020)Quan, Li, and Zhang]{quan2020novel}
H.~Quan, Y.~Li, and Y.~Zhang, ``A novel mobile robot navigation method based on
  deep reinforcement learning,'' \emph{International Journal of Advanced
  Robotic Systems}, vol.~17, no.~3, p. 1729881420921672, 2020.

\bibitem[Zeng et~al.(2020)Zeng, Song, Lee, Rodriguez, and
  Funkhouser]{zeng2020tossingbot}
A.~Zeng, S.~Song, J.~Lee, A.~Rodriguez, and T.~Funkhouser, ``Tossingbot:
  Learning to throw arbitrary objects with residual physics,'' \emph{IEEE
  Transactions on Robotics}, 2020.

\bibitem[Kalashnikov et~al.(2018)Kalashnikov, Irpan, Pastor, Ibarz, Herzog,
  Jang, Quillen, Holly, Kalakrishnan, Vanhoucke, et~al.]{kalashnikov2018qt}
D.~Kalashnikov, A.~Irpan, P.~Pastor, J.~Ibarz, A.~Herzog, E.~Jang, D.~Quillen,
  E.~Holly, M.~Kalakrishnan, V.~Vanhoucke \emph{et~al.}, ``Scalable deep
  reinforcement learning for vision-based robotic manipulation,'' in
  \emph{Conference on Robot Learning}, 2018, pp. 651--673.

\bibitem[Mahler et~al.(2017)Mahler, Liang, Niyaz, Laskey, Doan, Liu, Ojea, and
  Goldberg]{mahler2017dex}
J.~Mahler, J.~Liang, S.~Niyaz, M.~Laskey, R.~Doan, X.~Liu, J.~A. Ojea, and
  K.~Goldberg, ``Dex-net 2.0: Deep learning to plan robust grasps with
  synthetic point clouds and analytic grasp metrics,'' \emph{arXiv preprint
  arXiv:1703.09312}, 2017.

\bibitem[Levine et~al.(2016)Levine, Finn, Darrell, and Abbeel]{levine2016end}
S.~Levine, C.~Finn, T.~Darrell, and P.~Abbeel, ``End-to-end training of deep
  visuomotor policies,'' \emph{The Journal of Machine Learning Research},
  vol.~17, no.~1, pp. 1334--1373, 2016.

\bibitem[Zeng et~al.(2018)Zeng, Song, Welker, Lee, Rodriguez, and
  Funkhouser]{zeng2018learning}
A.~Zeng, S.~Song, S.~Welker, J.~Lee, A.~Rodriguez, and T.~Funkhouser,
  ``Learning synergies between pushing and grasping with self-supervised deep
  reinforcement learning,'' in \emph{2018 IEEE/RSJ International Conference on
  Intelligent Robots and Systems (IROS)}.\hskip 1em plus 0.5em minus
  0.4em\relax IEEE, 2018, pp. 4238--4245.

\bibitem[Osband et~al.(2016)Osband, Blundell, Pritzel, and
  Van~Roy]{osband2016deep}
I.~Osband, C.~Blundell, A.~Pritzel, and B.~Van~Roy, ``Deep exploration via
  bootstrapped dqn,'' in \emph{Advances in neural information processing
  systems}, 2016, pp. 4026--4034.

\bibitem[Osband et~al.(2019)Osband, Van~Roy, Russo, and Wen]{osband2019deep}
I.~Osband, B.~Van~Roy, D.~J. Russo, and Z.~Wen, ``Deep exploration via
  randomized value functions.'' \emph{Journal of Machine Learning Research},
  vol.~20, no. 124, pp. 1--62, 2019.

\bibitem[Levy et~al.(2019)Levy, Konidaris, Platt, and Saenko]{levy2018learning}
A.~Levy, G.~Konidaris, R.~Platt, and K.~Saenko, ``Learning multi-level
  hierarchies with hindsight,'' \emph{International Conference on Learning
  Representations}, 2019.

\bibitem[Nachum et~al.(2018{\natexlab{a}})Nachum, Gu, Lee, and
  Levine]{nachum2018data}
O.~Nachum, S.~S. Gu, H.~Lee, and S.~Levine, ``Data-efficient hierarchical
  reinforcement learning,'' in \emph{Advances in Neural Information Processing
  Systems}, 2018, pp. 3303--3313.

\bibitem[Nachum et~al.(2019)Nachum, Tang, Lu, Gu, Lee, and
  Levine]{nachum2019does}
O.~Nachum, H.~Tang, X.~Lu, S.~Gu, H.~Lee, and S.~Levine, ``Why does hierarchy
  (sometimes) work so well in reinforcement learning?'' \emph{arXiv preprint
  arXiv:1909.10618}, 2019.

\bibitem[Li et~al.(2020)Li, Xia, Mart{\'\i}n-Mart{\'\i}n, and
  Savarese]{li2020hrl4in}
C.~Li, F.~Xia, R.~Mart{\'\i}n-Mart{\'\i}n, and S.~Savarese, ``Hrl4in:
  Hierarchical reinforcement learning for interactive navigation with mobile
  manipulators,'' in \emph{Conference on Robot Learning}, 2020, pp. 603--616.

\bibitem[Mnih et~al.(2013)Mnih, Kavukcuoglu, Silver, Graves, Antonoglou,
  Wierstra, and Riedmiller]{mnih2013playing}
V.~Mnih, K.~Kavukcuoglu, D.~Silver, A.~Graves, I.~Antonoglou, D.~Wierstra, and
  M.~Riedmiller, ``Playing atari with deep reinforcement learning,''
  \emph{arXiv preprint arXiv:1312.5602}, 2013.

\bibitem[Haarnoja et~al.(2018)]{haarnoja2018soft}
T.~Haarnoja \emph{et~al.}, ``Soft actor-critic algorithms and applications,''
  \emph{arXiv preprint arXiv:1812.05905}, 2018.

\bibitem[Kuffner and LaValle(2000)]{kuffner2000rrt}
J.~J. Kuffner and S.~M. LaValle, ``Rrt-connect: An efficient approach to
  single-query path planning,'' in \emph{Proceedings 2000 ICRA. Millennium
  Conference. IEEE International Conference on Robotics and Automation.
  Symposia Proceedings (Cat. No. 00CH37065)}, vol.~2.\hskip 1em plus 0.5em
  minus 0.4em\relax IEEE, 2000, pp. 995--1001.

\bibitem[Bohlin and Kavraki(2000)]{bohlin2000path}
R.~Bohlin and L.~E. Kavraki, ``Path planning using lazy prm,'' in
  \emph{Proceedings 2000 ICRA. Millennium Conference. IEEE International
  Conference on Robotics and Automation. Symposia Proceedings (Cat. No.
  00CH37065)}, vol.~1.\hskip 1em plus 0.5em minus 0.4em\relax IEEE, 2000, pp.
  521--528.

\bibitem[M{\"u}ller et~al.(2018)M{\"u}ller, Dosovitskiy, Ghanem, and
  Koltun]{muller2018driving}
M.~M{\"u}ller, A.~Dosovitskiy, B.~Ghanem, and V.~Koltun, ``Driving policy
  transfer via modularity and abstraction,'' \emph{arXiv preprint
  arXiv:1804.09364}, 2018.

\bibitem[Kaufmann et~al.(2019)Kaufmann, Gehrig, Foehn, Ranftl, Dosovitskiy,
  Koltun, and Scaramuzza]{kaufmann2019beauty}
E.~Kaufmann, M.~Gehrig, P.~Foehn, R.~Ranftl, A.~Dosovitskiy, V.~Koltun, and
  D.~Scaramuzza, ``Beauty and the beast: Optimal methods meet learning for
  drone racing,'' in \emph{2019 International Conference on Robotics and
  Automation (ICRA)}.\hskip 1em plus 0.5em minus 0.4em\relax IEEE, 2019, pp.
  690--696.

\bibitem[Jurgenson and Tamar(2019)]{Tamar-RSS-19}
T.~Jurgenson and A.~Tamar, ``Harnessing reinforcement learning for neural
  motion planning,'' in \emph{Proceedings of Robotics: Science and Systems},
  Freiburg im Breisgau, Germany, June 2019.

\bibitem[Qureshi et~al.(2019)Qureshi, Simeonov, Bency, and
  Yip]{qureshi2019motion}
A.~H. Qureshi, A.~Simeonov, M.~J. Bency, and M.~C. Yip, ``Motion planning
  networks,'' in \emph{2019 International Conference on Robotics and Automation
  (ICRA)}.\hskip 1em plus 0.5em minus 0.4em\relax IEEE, 2019, pp. 2118--2124.

\bibitem[Bansal et~al.(2019)Bansal, Tolani, Gupta, Malik, and
  Tomlin]{bansal2019combining}
S.~Bansal, V.~Tolani, S.~Gupta, J.~Malik, and C.~Tomlin, ``Combining optimal
  control and learning for visual navigation in novel environments,'' in
  \emph{Conference on Robot Learning (CoRL)}, 2019.

\bibitem[Levine and Koltun(2013)]{levine2013guided}
S.~Levine and V.~Koltun, ``Guided policy search,'' in \emph{International
  Conference on Machine Learning}, 2013, pp. 1--9.

\bibitem[Jetchev and Toussaint(2010)]{jetchev2010trajectory}
N.~Jetchev and M.~Toussaint, ``Trajectory prediction in cluttered voxel
  environments,'' in \emph{2010 IEEE International Conference on Robotics and
  Automation}.\hskip 1em plus 0.5em minus 0.4em\relax IEEE, 2010, pp.
  2523--2528.

\bibitem[Rana et~al.(2018)Rana, Mukadam, Ahmadzadeh, Chernova, and
  Boots]{rana2018towards}
M.~Rana, M.~Mukadam, S.~R. Ahmadzadeh, S.~Chernova, and B.~Boots, ``Towards
  robust skill generalization: Unifying learning from demonstration and motion
  planning,'' in \emph{Intelligent robots and systems}, 2018.

\bibitem[Ota et~al.(2020)Ota, Sasaki, Jha, Yoshiyasu, and
  Kanezaki]{ota2020efficient}
K.~Ota, Y.~Sasaki, D.~K. Jha, Y.~Yoshiyasu, and A.~Kanezaki, ``Efficient
  exploration in constrained environments with goal-oriented reference path,''
  \emph{arXiv preprint arXiv:2003.01641}, 2020.

\bibitem[Jiang et~al.(2019)Jiang, Yang, Zhang, and Stone]{jiang2018integrating}
Y.~Jiang, F.~Yang, S.~Zhang, and P.~Stone, ``Integrating task-motion planning
  with reinforcement learning for robust decision making in mobile robots,'' in
  \emph{In Proceedings of the AAMAS}, 2019.

\bibitem[Dragan et~al.(2011)Dragan, Gordon, and Srinivasa]{dragan2011learning}
A.~Dragan, G.~J. Gordon, and S.~Srinivasa, ``Learning from experience in
  manipulation planning: Setting the right goals,'' in \emph{In Proceedings of
  the ISRR}, 2011.

\bibitem[Yamada et~al.(2020)Yamada, Salhotra, Lee, Pflueger, Pertsch, Englert,
  Sukhatme, and Lim]{yamadamotion}
J.~Yamada, G.~Salhotra, Y.~Lee, M.~Pflueger, K.~Pertsch, P.~Englert, G.~S.
  Sukhatme, and J.~J. Lim, ``Motion planner augmented action spaces for
  reinforcement learning,'' \emph{RSS Workshop on Action Representations for
  Learning in Continuous Control}, 2020.

\bibitem[Wu et~al.(2020)Wu, Sun, Zeng, Song, Lee, Rusinkiewicz, and
  Funkhouser]{wu2020spatial}
J.~Wu, X.~Sun, A.~Zeng, S.~Song, J.~Lee, S.~Rusinkiewicz, and T.~Funkhouser,
  ``{Spatial Action Maps for Mobile Manipulation},'' in \emph{Proceedings of
  Robotics: Science and Systems}, Corvalis, Oregon, USA, July 2020.

\bibitem[Heess et~al.(2016)Heess, Wayne, Tassa, Lillicrap, Riedmiller, and
  Silver]{heess2016learning}
N.~Heess, G.~Wayne, Y.~Tassa, T.~Lillicrap, M.~Riedmiller, and D.~Silver,
  ``Learning and transfer of modulated locomotor controllers,'' \emph{arXiv
  preprint arXiv:1610.05182}, 2016.

\bibitem[Kulkarni et~al.(2016)Kulkarni, Narasimhan, Saeedi, and
  Tenenbaum]{kulkarni2016hierarchical}
T.~D. Kulkarni, K.~Narasimhan, A.~Saeedi, and J.~Tenenbaum, ``Hierarchical deep
  reinforcement learning: Integrating temporal abstraction and intrinsic
  motivation,'' in \emph{Advances in neural information processing systems},
  2016, pp. 3675--3683.

\bibitem[Vezhnevets et~al.(2017)Vezhnevets, Osindero, Schaul, Heess, Jaderberg,
  Silver, and Kavukcuoglu]{vezhnevets2017feudal}
A.~S. Vezhnevets, S.~Osindero, T.~Schaul, N.~Heess, M.~Jaderberg, D.~Silver,
  and K.~Kavukcuoglu, ``Feudal networks for hierarchical reinforcement
  learning,'' in \emph{Proceedings of the 34th International Conference on
  Machine Learning-Volume 70}.\hskip 1em plus 0.5em minus 0.4em\relax JMLR.
  org, 2017, pp. 3540--3549.

\bibitem[Nachum et~al.(2018{\natexlab{b}})Nachum, Gu, Lee, and
  Levine]{nachum2018near}
O.~Nachum, S.~Gu, H.~Lee, and S.~Levine, ``Near-optimal representation learning
  for hierarchical reinforcement learning,'' \emph{International Conference on
  Learning Representations}, 2018.

\bibitem[Konidaris et~al.(2011)Konidaris, Kuindersma, Grupen, and
  Barto]{konidaris2011autonomous}
G.~Konidaris, S.~Kuindersma, R.~Grupen, and A.~Barto, ``Autonomous skill
  acquisition on a mobile manipulator,'' in \emph{Twenty-Fifth AAAI Conference
  on Artificial Intelligence}, 2011.

\bibitem[Osband et~al.(2018)Osband, Aslanides, and
  Cassirer]{osband2018randomized}
I.~Osband, J.~Aslanides, and A.~Cassirer, ``Randomized prior functions for deep
  reinforcement learning,'' in \emph{Advances in Neural Information Processing
  Systems}, 2018, pp. 8617--8629.

\bibitem[Ciosek et~al.(2019)Ciosek, Vuong, Loftin, and
  Hofmann]{ciosek2019better}
K.~Ciosek, Q.~Vuong, R.~Loftin, and K.~Hofmann, ``Better exploration with
  optimistic actor critic,'' in \emph{Advances in Neural Information Processing
  Systems}, 2019, pp. 1787--1798.

\bibitem[Anderson et~al.(2018)]{anderson2018evaluation}
P.~Anderson \emph{et~al.}, ``On evaluation of embodied navigation agents,''
  \emph{arXiv preprint arXiv:1807.06757}, 2018.

\bibitem[{Manolis Savva} et~al.(2019){Manolis Savva}, {Abhishek Kadian},
  {Oleksandr Maksymets}, et~al.]{habitat19arxiv}
{Manolis Savva}, {Abhishek Kadian}, {Oleksandr Maksymets} \emph{et~al.},
  ``Habitat: {A} {P}latform for {E}mbodied {AI} {R}esearch,'' in
  \emph{Proceedings of the IEEE/CVF International Conference on Computer Vision
  (ICCV)}, 2019.

\bibitem[Zamora et~al.(2016)Zamora, Lopez, Vilches, and
  Cordero]{zamora2016extending}
I.~Zamora, N.~G. Lopez, V.~M. Vilches, and A.~H. Cordero, ``Extending the
  openai gym for robotics: a toolkit for reinforcement learning using ros and
  gazebo,'' \emph{arXiv preprint arXiv:1608.05742}, 2016.

\bibitem[{Xia} et~al.(2020){Xia}, {Shen}, {Li}, {Kasimbeg}, {Tchapmi},
  {Toshev}, {Martín-Martín}, and {Savarese}]{xia2020interactive}
F.~{Xia}, W.~B. {Shen}, C.~{Li}, P.~{Kasimbeg}, M.~E. {Tchapmi}, A.~{Toshev},
  R.~{Martín-Martín}, and S.~{Savarese}, ``Interactive gibson benchmark: A
  benchmark for interactive navigation in cluttered environments,'' \emph{IEEE
  Robotics and Automation Letters}, vol.~5, no.~2, pp. 713--720, April 2020.

\bibitem[Berenson et~al.(2008)Berenson, Kuffner, and
  Choset]{berenson2008optimization}
D.~Berenson, J.~Kuffner, and H.~Choset, ``An optimization approach to planning
  for mobile manipulation,'' in \emph{2008 IEEE International Conference on
  Robotics and Automation}.\hskip 1em plus 0.5em minus 0.4em\relax IEEE, 2008,
  pp. 1187--1192.

\bibitem[Klingbeil et~al.(2010)Klingbeil, Saxena, and
  Ng]{klingbeil2010learning}
E.~Klingbeil, A.~Saxena, and A.~Y. Ng, ``Learning to open new doors,'' in
  \emph{2010 IEEE/RSJ International Conference on Intelligent Robots and
  Systems}.\hskip 1em plus 0.5em minus 0.4em\relax IEEE, 2010, pp. 2751--2757.

\bibitem[Wang et~al.(2020)Wang, Zhang, Tian, Li, Wang, Lane, Petillot, and
  Wang]{wang2020learning}
C.~Wang, Q.~Zhang, Q.~Tian, S.~Li, X.~Wang, D.~Lane, Y.~Petillot, and S.~Wang,
  ``Learning mobile manipulation through deep reinforcement learning,''
  \emph{Sensors}, vol.~20, no.~3, p. 939, 2020.

\bibitem[Wang et~al.(2019)Wang, Zhou, Shi, Chen, Zhao, and
  Xu]{wang2019shape2motion}
X.~Wang, B.~Zhou, Y.~Shi, X.~Chen, Q.~Zhao, and K.~Xu, ``Shape2motion: Joint
  analysis of motion parts and attributes from 3d shapes,'' in
  \emph{Proceedings of the IEEE Conference on Computer Vision and Pattern
  Recognition}, 2019, pp. 8876--8884.

\bibitem[Chang et~al.(2015)]{chang2015shapenet}
A.~X. Chang \emph{et~al.}, ``Shapenet: An information-rich 3d model
  repository,'' \emph{arXiv preprint arXiv:1512.03012}, 2015.

\bibitem[Hernandez(2017)]{hernandez2017survive}
D.~Hernandez, ``How to survive a robot apocalypse: Just close the door,''
  \emph{The Wall Street Journal}, p.~10, 2017.

\bibitem[{Sergio Guadarrama and others}(2018)]{TFAgents}
\BIBentryALTinterwordspacing
{Sergio Guadarrama and others}, ``{TF-Agents}: A library for reinforcement
  learning in tensorflow,'' \url{https://github.com/tensorflow/agents}, 2018.
  [Online]. Available: \url{https://github.com/tensorflow/agents}
\BIBentrySTDinterwordspacing

\bibitem[Stooke and Abbeel(2019)]{stooke2019rlpyt}
A.~Stooke and P.~Abbeel, ``rlpyt: A research code base for deep reinforcement
  learning in pytorch,'' \emph{arXiv preprint arXiv:1909.01500}, 2019.

\bibitem[{Caelan Reed Garrett}(2018)]{sspybullet}
{Caelan Reed Garrett}, ``{PyBullet Planning.}''
  \url{https://pypi.org/project/pybullet-planning/}, 2018.

\bibitem[Bousmalis et~al.(2018)Bousmalis, Irpan, Wohlhart, Bai, Kelcey,
  Kalakrishnan, Downs, Ibarz, Pastor, Konolige, et~al.]{bousmalis2018using}
K.~Bousmalis, A.~Irpan, P.~Wohlhart, Y.~Bai, M.~Kelcey, M.~Kalakrishnan,
  L.~Downs, J.~Ibarz, P.~Pastor, K.~Konolige \emph{et~al.}, ``Using simulation
  and domain adaptation to improve efficiency of deep robotic grasping,'' in
  \emph{2018 IEEE international conference on robotics and automation
  (ICRA)}.\hskip 1em plus 0.5em minus 0.4em\relax IEEE, 2018, pp. 4243--4250.

\bibitem[Rao et~al.(2020)Rao, Harris, Irpan, Levine, Ibarz, and
  Khansari]{rao2020rl}
K.~Rao, C.~Harris, A.~Irpan, S.~Levine, J.~Ibarz, and M.~Khansari,
  ``Rl-cyclegan: Reinforcement learning aware simulation-to-real,'' in
  \emph{Proceedings of the IEEE/CVF Conference on Computer Vision and Pattern
  Recognition}, 2020, pp. 11\,157--11\,166.

\bibitem[Ramos et~al.(2019)Ramos, Possas, and Fox]{ramos2019bayessim}
F.~Ramos, R.~C. Possas, and D.~Fox, ``Bayessim: adaptive domain randomization
  via probabilistic inference for robotics simulators,'' \emph{arXiv preprint
  arXiv:1906.01728}, 2019.

\bibitem[Chebotar et~al.(2019)Chebotar, Handa, Makoviychuk, Macklin, Issac,
  Ratliff, and Fox]{chebotar2019closing}
Y.~Chebotar, A.~Handa, V.~Makoviychuk, M.~Macklin, J.~Issac, N.~Ratliff, and
  D.~Fox, ``Closing the sim-to-real loop: Adapting simulation randomization
  with real world experience,'' in \emph{2019 International Conference on
  Robotics and Automation (ICRA)}.\hskip 1em plus 0.5em minus 0.4em\relax IEEE,
  2019, pp. 8973--8979.

\bibitem[Kang et~al.(2019)Kang, Belkhale, Kahn, Abbeel, and
  Levine]{kang2019generalization}
K.~Kang, S.~Belkhale, G.~Kahn, P.~Abbeel, and S.~Levine, ``Generalization
  through simulation: Integrating simulated and real data into deep
  reinforcement learning for vision-based autonomous flight,''
  \emph{International Conference on Robotics and Automation (ICRA)}, 2019.

\bibitem[Meng et~al.(2019)Meng, Ratliff, Xiang, and Fox]{meng2019neural}
X.~Meng, N.~Ratliff, Y.~Xiang, and D.~Fox, ``Neural autonomous navigation with
  riemannian motion policy,'' in \emph{2019 International Conference on
  Robotics and Automation (ICRA)}.\hskip 1em plus 0.5em minus 0.4em\relax IEEE,
  2019, pp. 8860--8866.

\bibitem[Xia et~al.(2018)Xia, Zamir, He, Sax, Malik, and
  Savarese]{xia2018gibson}
F.~Xia, A.~R. Zamir, Z.~He, A.~Sax, J.~Malik, and S.~Savarese, ``Gibson env:
  Real-world perception for embodied agents,'' in \emph{Proceedings of the IEEE
  Conference on Computer Vision and Pattern Recognition}, 2018, pp. 9068--9079.

\bibitem[Tan et~al.(2018)Tan, Zhang, Coumans, Iscen, Bai, Hafner, Bohez, and
  Vanhoucke]{tan2018sim}
J.~Tan, T.~Zhang, E.~Coumans, A.~Iscen, Y.~Bai, D.~Hafner, S.~Bohez, and
  V.~Vanhoucke, ``Sim-to-real: Learning agile locomotion for quadruped
  robots,'' \emph{arXiv preprint arXiv:1804.10332}, 2018.

\end{thebibliography}
\endgroup


\renewcommand{\thesection}{\Alph{section}}
\renewcommand{\thesubsection}{A.\arabic{subsection}}

\setcounter{figure}{0} \renewcommand{\thefigure}{A.\arabic{figure}}

\setcounter{table}{0} \renewcommand{\thetable}{A.\arabic{table}}

\clearpage

\section*{Appendix for ReLMoGen: Leveraging Motion Generation in Reinforcement Learning for Mobile Manipulation}





In the appendix, we provide more details about the task specification, training procedure, network structure, and simulation environment, as well as additional experimental results and analysis. We also show that our method can be fine-tuned to transfer to completely unseen scenes and new robot embodiments. Finally, we highlight how the characteristics of our method help bridge the Sim2Real gap.


\subsection{Tasks}
\label{subsec:appendix_task}
\paragraph{Task Specification}
In the following, we include additional details of the seven tasks we evaluate in our experiments and the main challenges they pose to policy learning for visuo-motor control.

    \textbf{\texttt{PointNav}}: In this task, the robot needs to navigate from one point to another without collision. The robot's initial pose (3-DoF) and the goal position (2-DoF) are randomly sampled on the floor plan such that the geodesic distance between them is between $\SI{1}{m}$ and $\SI{10}{m}$. This task evaluates ReLMoGen and the baselines for pure navigation without arm control.

    \textbf{\texttt{TabletopReachM}}: In this task, the robot needs to reach an area on the table in front of it. The goal area is represented by a red visual marker. The task is similar to the \texttt{FetchReach} task in OpenAI Gym~\cite{zamora2016extending}. In our setup, however, the robot is not provided with the ground truth position of the goal, and has to rely on the visual cues from RGB images to detect the goal area and reach it. The goal is randomly sampled on the table surface. 
    
    These first two tasks allow us to benchmark the performance of ReLMoGen in relatively simple navigation and stationary arm control domains, although the benefits of using ReLMoGen are more evident in more complex interactive navigation and mobile manipulation domains.

    \textbf{\texttt{PushDoorNav}}: In this task, the robot needs to push a door open with its arm in order to reach the goal inside the closed room, which is a common scenario in human homes and offices. This is still challenging for most robots~\cite{hernandez2017survive}. To solve this task, the robot needs to place its base in a suited location that allows it to push the door open~\cite{berenson2008optimization, klingbeil2010learning}.

    \textbf{\texttt{ButtonDoorNav}}: In this task, the robot also needs to enter a closed room, but this time the robot can only open the door through pressing a button positioned next to it. The button's position is randomized on the wall next to the door. This task resembles the accessible entrances designed for people with disabilities. To solve this task, the robot needs to exploit the relationship between the button and the door, and controls the arm to press the relatively small button in a precise manner.
    
    \textbf{\texttt{InteractiveObstaclesNav}}: In this task, the robot needs to reach a goal in a region of the environment that is blocked off by two large obstacles. Their size is similar to that of a chair or a small table: $\SI{0.7}{\meter} \times \SI{0.7}{\meter} \times \SI{1.2}{\meter}$. The positions of the obstacles are randomized across episodes but they always block the path towards the goal. The two obstacles have two different colors that link to their weights: the red obstacle weighs $\SI{1.0}{\kilogram}$ and the green obstacle weighs $\SI{1.0e4}{\kilogram}$ (essentially not movable). To solve this task, the robot needs to associate the obstacles' color with their weight using RGB information and decide on which obstacle to interact with. 
    
    For the above three Interactive Navigation tasks~\cite{xia2020interactive}, the robot initial pose and goal position are randomly sampled in two different regions as shown in Fig.~\ref{fig:sim_env}.
    
    \textbf{\texttt{ArrangeKitchenMM}}: In this task, the robot needs to tidy up a messy kitchen, where the cabinet doors and drawers are initially open to different degrees at random. A total of four sets of cabinets and drawers are randomly placed along three walls in the room. The robot needs to close as many cabinet doors and drawers boxes as possible within a time budget. There are several challenges in this task: the agent needs to find the cabinets and drawers using RGB-D information, navigate close to them if they are open, and accurately push them along their axes of unconstrained motion.
    
    \textbf{\texttt{ArrangeChairMM}}: In this task, the robot needs to arrange the chairs by tucking them under the table. The chairs are randomly initialized close to the table. The main challenge in this task is that the agent needs to learn accurate pushing actions that bring the chairs through the narrow passage between the table legs.
    
    An additional challenge in the above two Mobile Manipulation tasks is that there is no goal information provided: the robot has no information about which objects are task-relevant, their pose or their desired final state. The agent needs to learn to detect the task-relevant objects using the visual input, place the base in front of them, and interact with them in the correct manner, a hard perception and exploration problem alleviated by the motion generators of ReLMoGen.

\paragraph{Reward and Evaluation Metrics}
In Table~\ref{tbl:reward} we summarize the reward and evaluation metrics. In our experiments, we used $d_{th}=\SI{0.5}{m}$ and $d_{gth}=\SI{0.1}{m}$.

\begin{table*}[t]
    \begin{center}
    \resizebox{0.95\textwidth}{!}{
  \begin{tabular}{lp{6cm}p{6cm}} 
  \toprule
  Task & Reward & Evaluation Metrics \\
  \midrule
     \texttt{PointNav} & Geodesic distance reduction reward $R_{Nav}$, Success reward $R_{Succ}$& Success: Robot arrive at goal within $d_{th}$, SPL\\
     \texttt{TabletopReachM} & Negative L2 distance reward $R_{Reach}$, Success reward $R_{Succ}$& Success: Robot gripper reach goal within $d_{gth}$\\
     \texttt{PushDoorNav} & Geodesic distance reduction reward $R_{Nav}$, Push door reward $R_{Door}$, Success reward $R_{Succ}$& Success: Robot arrive at goal within $d_{th}$, SPL \\
     \texttt{ButtonDoorNav} & Geodesic distance reduction reward $R_{Nav}$, Push button reward $R_{Button}$, Success reward $R_{Succ}$& Success: Robot arrive at goal within $d_{th}$, SPL \\
     \texttt{InteractiveObstaclesNav} & Geodesic distance reduction reward $R_{Nav}$, Push obstacles reward $R_{Obs}$, Success reward $R_{Succ}$ & Success: Robot arrive at goal within $d_{th}$, SPL \\
     \texttt{ArrangeKitchenMM} & Push drawer reward $R_{Drawer}$ & Drawer boxes and cabinet doors closed within $\SI{5}{\degree}/\SI{5}{cm}$ and $\SI{10}{\degree}/\SI{10}{cm}$\\
     \texttt{ArrangeChairMM} & Push chair reward $R_{Chair}$ & Chairs moved to within $\SI{5}{cm}$ and $\SI{10}{cm}$ of the fully tucked position\\
   \bottomrule
  \end{tabular}}
  \vspace{10pt}
\caption{Reward and metric definition\label{tbl:reward}}
  \end{center}
\end{table*}

\subsection{Training Details}
\label{subsec:appendix_training}
In the following, we provide details on the ReLMoGen algorithm, network architecture, motion generator implementation, training procedure, and hyperparameters for our algorithms and simulation environment.

\paragraph{Algorithm Description}
An detailed description of our ReLMoGen algorithm is included in Algorithm~\ref{relmogen_algo}.

\begin{algorithm*}[ht]
\DontPrintSemicolon
\SetAlgoLined
\SetKwInOut{Input}{Input}\SetKwInOut{Output}{Output}
\SetKwInOut{Parameters}{Parameters}
\Input{\texttt{env}, \texttt{MG}, \texttt{D}}
\Output{$\pi$}
\Parameters{$n_{iter}$, $n_{env\_step}$, $n_{grad\_step}$}
\BlankLine
\For{$iter \gets 1$ \KwTo $n_{iter}$}{
  \For{$step \gets 1$ \KwTo $n_{env\_step}$}{
    $a'_t \gets \pi(o_t)$ \tcp*{sample the next subgoal}
    $\{a_t, a_{t+1}, ..., a_{t + T - 1}\} \gets \texttt{MG}(a'_t)$ \tcp*{motion generator plans for T low-level actions; if the subgoal is infeasible, T = 0}
    $r'_t = 0$ \\
    \For{$i \gets 0$ \KwTo $T - 1$}{
      $o_{t+i+1}, r_{t+i+1} \gets \texttt{env.step}(a_{t+i})$ \\
      $r'_t \gets r'_t + r_{t+i+1}$ \tcp*{accumulate reward within a subgoal execution}
    }
    $\texttt{D} \gets \texttt{D} \cup \{o_t, a'_t, r'_t, o_{t+T}\}$ 
  }
  \For{$step \gets 1$ \KwTo $n_{grad\_step}$}{
    perform gradient updates for $\pi$ with \texttt{D} as defined in \cite{haarnoja2018soft} (policy gradient based) or \cite{mnih2013playing} (Q learning based)
  }
}
\caption{ReLMoGen Algorithm\label{relmogen_algo}}
\end{algorithm*}

\paragraph{Network Structure}
For SGP-R, we use three 2D convolutional layers to process RGB-D images, three 1D convolutional layers to process LiDAR scan, and two fully connected layers with ReLU activation to process additional task information such as goals and waypoints. Each branch is then flattened and processed by one fully connected layer with ReLU activation before concatenation. Finally, the features are passed through two fully connected layers with ReLU activation in the actor network and critic network to output action distribution and estimate Q-values respectively. Our implementation of SGP-R is based upon TF-Agents~\cite{TFAgents}.

For SGP-D, we first pre-process the LiDAR scan into a local occupancy map. For navigation-related tasks, we augment the local occupancy map with additional task information: we also ``draw'' the goal and equidistant waypoints computed from the initial robot's location to the goal on the local map as an additional channel. We use four 2D convolutional layers with stride $2$ to process RGB-D images and local occupancy maps in two different branches. The feature maps from both branches are concatenated. Finally, the feature maps are passed through two 2D deconvolutional layers with stride $2$ to generate Q-value maps for base subgoals ($L$ channels representing $L$ discretized desired base orientations) and Q-value maps for arm subgoal ($K$ channels representing $K$ discretized pushing direction). The spatial dimensions of the Q-value maps are down-sampled $4$ times from the input images. The output action corresponds to the pixel with the maximum Q-value across all $K + L$ action maps. Our implementation of SGP-D is based upon rlpyt~\cite{stooke2019rlpyt}.

\paragraph{Motion Generation and Subgoal Action Spaces}
We built the motion generators used in this paper (RRT-Connect and Lazy PRM) based on~\cite{sspybullet}. The hyperparameters can be found in Table~\ref{tbl:mp_params}. In addition, we provide hyperparameters for our subgoal action spaces. The base subgoal range is $[-\SI{2.5}{m}, -\SI{2.5}{m}]\times [\SI{2.5}{m}, \SI{2.5}{m}]$ around the robot. The arm subgoal space is $[0,\texttt{image\_height}]\times [0,\texttt{image\_width}]$, as the arm subgoal is chosen by picking one point on the depth map. The parameterized pushing action has a maximum pushing distance of $\SI{0.25}{m}$.

\begin{table*}[t]
    \begin{center}
  \resizebox{\textwidth}{!}{
  \begin{tabular}{lcccccc} 
  \toprule
     & \texttt{PushDoorNav} & \texttt{ButtonDoorNav} & \texttt{InteractiveObstaclesNav} & \texttt{ArrangeKitchenMM} & \texttt{ArrangeChairMM}\\ 
     & SR & SR & SR & drawers pushed ($\SI{10}{\degree}/\SI{10}{cm}$) & chairs pushed ($\SI{10}{cm}$) \\
  \midrule
ReLMoGen-R Train & 0.99 & 0.91 & 0.95 & 5.22 & 0.38 \\
ReLMoGen-R Eval & 0.99 \textbf{(+0.0)} & 0.87 \textbf{(-0.04)}& 0.91 \textbf{(-0.04)} & 5.25 \textbf{(+0.03)} &	0.34 \textbf{(-0.04)} \\
  \midrule
ReLMoGen-D Train & 0.85 & 0.62 & 0.53 & 5.45 & 0.3 \\
ReLMoGen-D Eval & 0.8 \textbf{(-0.05)} & 0.66 \textbf{(+0.04)}& 0.6 \textbf{(+0.07)} & 5.72 \textbf{(+0.27)} &	0.43 \textbf{(+0.13)} \\
    \bottomrule
  \end{tabular}
  }
    \vspace{10pt}
 \caption{\small{We observe minimal performance drop due to the domain gap caused by the fact that we disable collision checking in arm motion planning during training. The results are from the best performing checkpoints.\label{tbl:cmp_to_fmp}}}
  \end{center}
\end{table*}

\paragraph{Training Procedures}
To accelerate learning and reduce motion planner failures or timeouts, we disable collision checking in arm motion planning during training. At evaluation time, however, collision checking is enabled for the entire trajectory to ensure feasibility. While this introduces a small domain gap between training and evaluation, we found empirically that this provides substantial benefits for training. We can train faster with fewer collision checking queries and suffer less from the stochastic failures of sampling-based motion planners. The aforementioned domain gap causes little performance drop at evaluation time (see Table~\ref{tbl:cmp_to_fmp}), showing the robustness of our Subgoal Generation Policy.

\paragraph{Hyperparameters}
We summarize the hyperparameters for SGP-R, SGP-D, motion generators, and iGibson simulator in Table~\ref{tbl:sgp-r}, Table~\ref{tbl:sgp-d}, Table~\ref{tbl:mp_params} and Table~\ref{tbl:env_params}.

\begin{table}
\begin{center}
\resizebox{0.5\textwidth}{!}{
\begin{tabular}{lp{3cm}p{3cm}}
\toprule
Hyperparameter & Value \\ \midrule
Num parallel training environments & $16$ \\
Initial collect steps & $200$ \\ 
Collect steps per iteration & $1$ \\ 
Replay buffer size & $\num{1e4}$ \\
Target network update tau & $0.005$ \\
Target network update period & $1$ \\
Train steps per iteration & $1$ \\ 
Batch size & $256$ \\ 
Optimizer & Adam \\
Learning rate & $\num{3e-4}$ \\ 
TD loss type & MSE \\ 
Discount factor & $0.99$ \\ 
Reward scale factor & $1$ \\ \bottomrule
\end{tabular}
}
\vspace{10pt}
\caption{Hyperparameters for SGP-R\label{tbl:sgp-r}}
\end{center}
\end{table}

\begin{table}
\begin{center}
\resizebox{0.5\textwidth}{!}{
\begin{tabular}{lp{3cm}p{3cm}}
\toprule
Hyperparameter & Value \\ \midrule
Num parallel training environments & $16$ \\
Initial collect steps & $1000$ \\ 
Collect steps per iteration & $25$ \\ 
Replay buffer size & $\num{1e4}$ \\
Replay buffer ratio & $8$ \\
Target network update tau & $1$ \\
Target network update period & $1024$ \\
Train steps per iteration & $6$ \\ 
Batch size & $512$ \\ 
Optimizer & Adam \\
Learning rate & $\num{2.5e-4}$ \\ 
TD loss type & Huber \\ 
Discount factor & $0.99$ \\
Double DQN & True \\ 
Initial Epsilon & $0.8$ \\ 
Clip gradient norm & $10$ \\ \bottomrule
\end{tabular}
}
\vspace{10pt}
\caption{Hyperparameters for SGP-D\label{tbl:sgp-d}}
\end{center}
\end{table}

\begin{table}
\begin{center}
\resizebox{0.5\textwidth}{!}{
\begin{tabular}{lp{3cm}p{3cm}}
\toprule
Hyperparameter & Value \\ \midrule
Arm inverse kinematics steps & $100$ \\
Arm inverse kinematics restarts & $50$ \\
Arm inverse kinematics threshold & $\SI{0.05}{m}$ \\
Base motion planning resolution  & $\SI{0.05}{m}$  \\
Arm motion planning resolution  &  $\SI{0.05}{rad}$ \\
RRT-Connect iterations  & $20$ \\
RRT-Connect restarts  & $2$ \\
LazyPRM iterations  & $[500,2000,5000]$ \\
\bottomrule
\end{tabular}
}
\vspace{10pt}
\caption{Hyperparameters for motion generators used in this work.\label{tbl:mp_params}}
\end{center}
\end{table}

\begin{table}
\begin{center}
\resizebox{0.5\textwidth}{!}{
\begin{tabular}{lp{3cm}p{3cm}}
\toprule
Hyperparameter & Value \\ \midrule
Default robot & Fetch \\
Action step (for baselines) & $\SI{0.1}{s}$ \\
Action step (for ReLMoGen) & $\SI{3}{s}$ \\
Physics step & $\SI{0.025}{s}$ \\
RGB-D resolution & $128$ \\ 
RGB-D field of view & $\SI{90}{\degree}$ \\ 
Depth camera range minimum & $\SI{0.35}{m}$ \\
Depth camera range maximum & $\SI{3.0}{m}$ \\
LiDAR num vertical beams & $1$ \\
LiDAR num horizontal rays & $220$ \\
LiDAR num field of view & $\SI{220}{\degree}$ \\
\bottomrule
\end{tabular}
}
\vspace{10pt}
\caption{Hyperparameters for iGibson simulator\label{tbl:env_params}}
\end{center}
\end{table}


\subsection{Fine-tuning Results}
 

\paragraph{Fine-tuning in A New Environment}

Although our policy is trained in a single environment per task, we are able to fine tune it on novel environments and achieve good performance. The fine-tuning procedure is as follows. We first train \texttt{PushDoorNav} task on Scene-A (the scene introduced in the main paper in Fig.~\ref{fig:sim_env}) until convergence. Then we swap half of the training environments with Scene-B (not seen previously). We show that the policy is able to solve \texttt{PushDoorNav} in Scene-B while retaining good performance in Scene-A, using as few as $\num{2e4}$ training episodes (see Table~\ref{tbl:ft_new_scene} for more details). This procedure could be repeated in order to solve \texttt{PushDoorNav} in more scenes.

\begin{table}
\begin{center}
\resizebox{0.5\textwidth}{!}{
\begin{tabular}{lp{1cm}p{1cm}p{1cm}p{1cm}}
\toprule

& \multicolumn{2}{c}{Scene-A}  & \multicolumn{2}{c}{Scene-B (new)} \\
& SR & Reward & SR & Reward \\
\midrule
Before fine-tuning & $0.95$ & $21.8$ & $0.0$ & $2.91$  \\ 
After fine-tuning & $0.97$ & $22.1$  & $0.88$ & $26.60$\\ \bottomrule
\end{tabular}
}
\vspace{10pt}
\caption{\small{Fine-tuning performance for \texttt{PushDoorNav} on a new scene}\label{tbl:ft_new_scene}}
\end{center}
\end{table}

\begin{figure*}[t]
\begin{center}
\begin{subfigure}[b]{0.3\textwidth}
    \includegraphics[width=\linewidth]{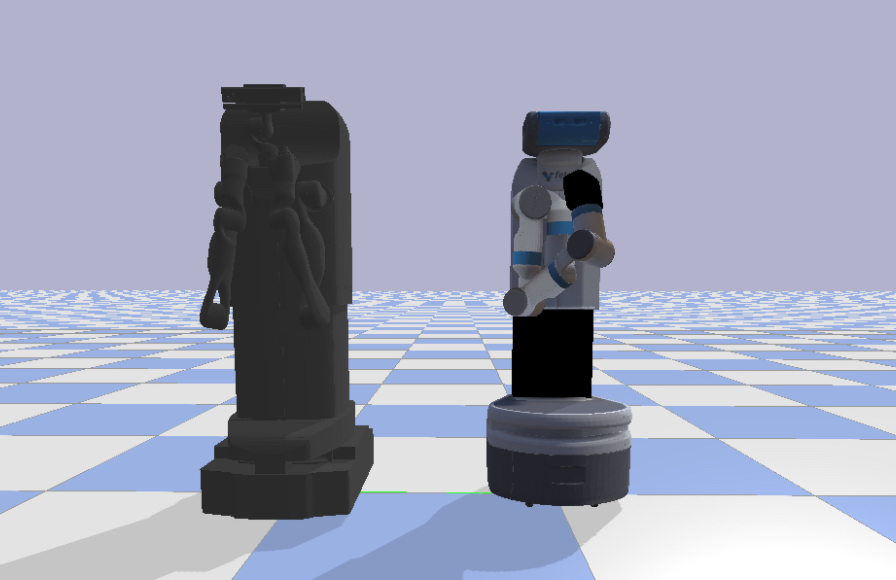}
    \caption{Movo and Fetch}
\end{subfigure}
\begin{subfigure}[b]{0.3\textwidth}
    \includegraphics[width=\linewidth]{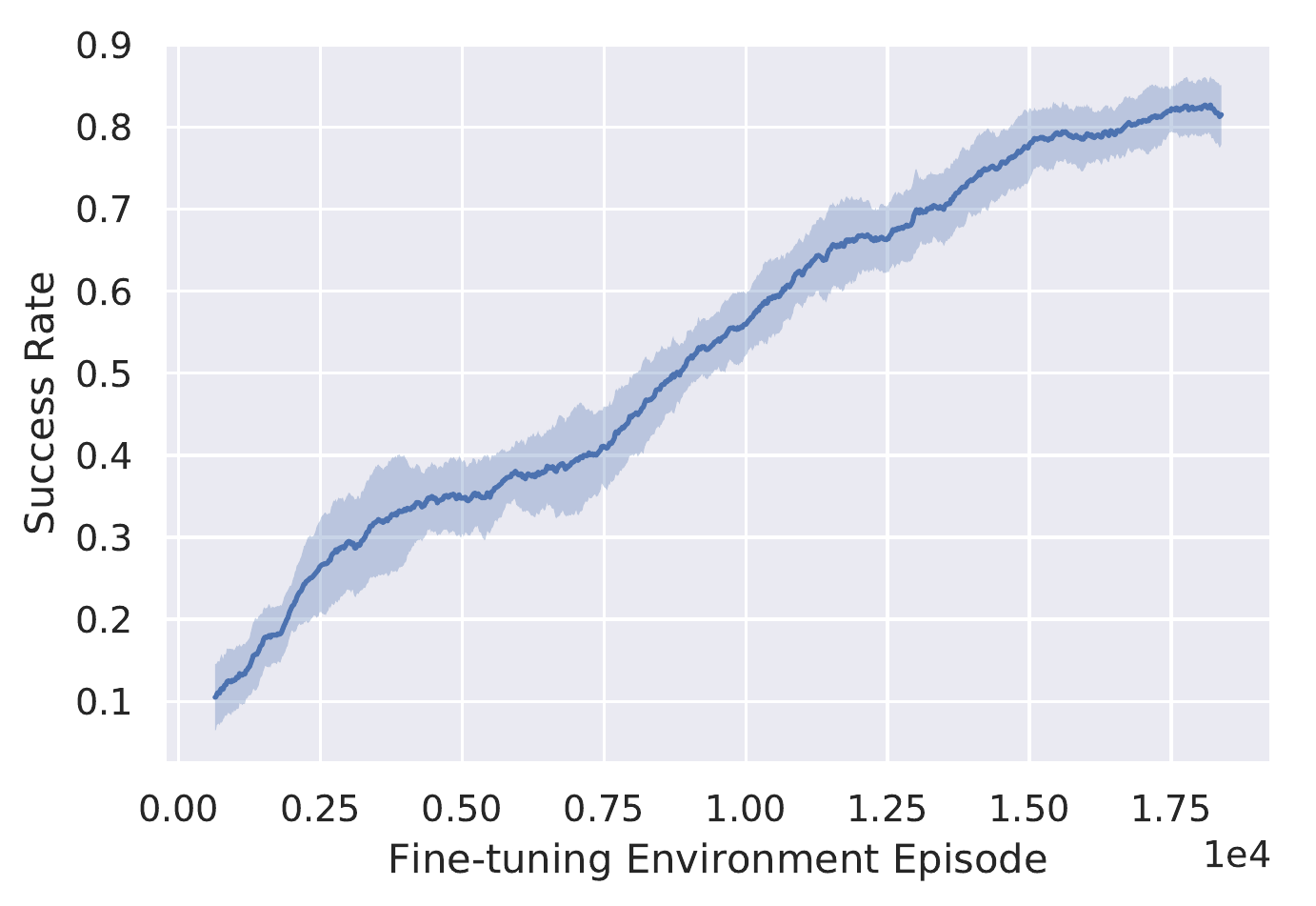}
    \caption{Task Success Rate}
\end{subfigure}
\begin{subfigure}[b]{0.3\textwidth}
    \includegraphics[width=\linewidth]{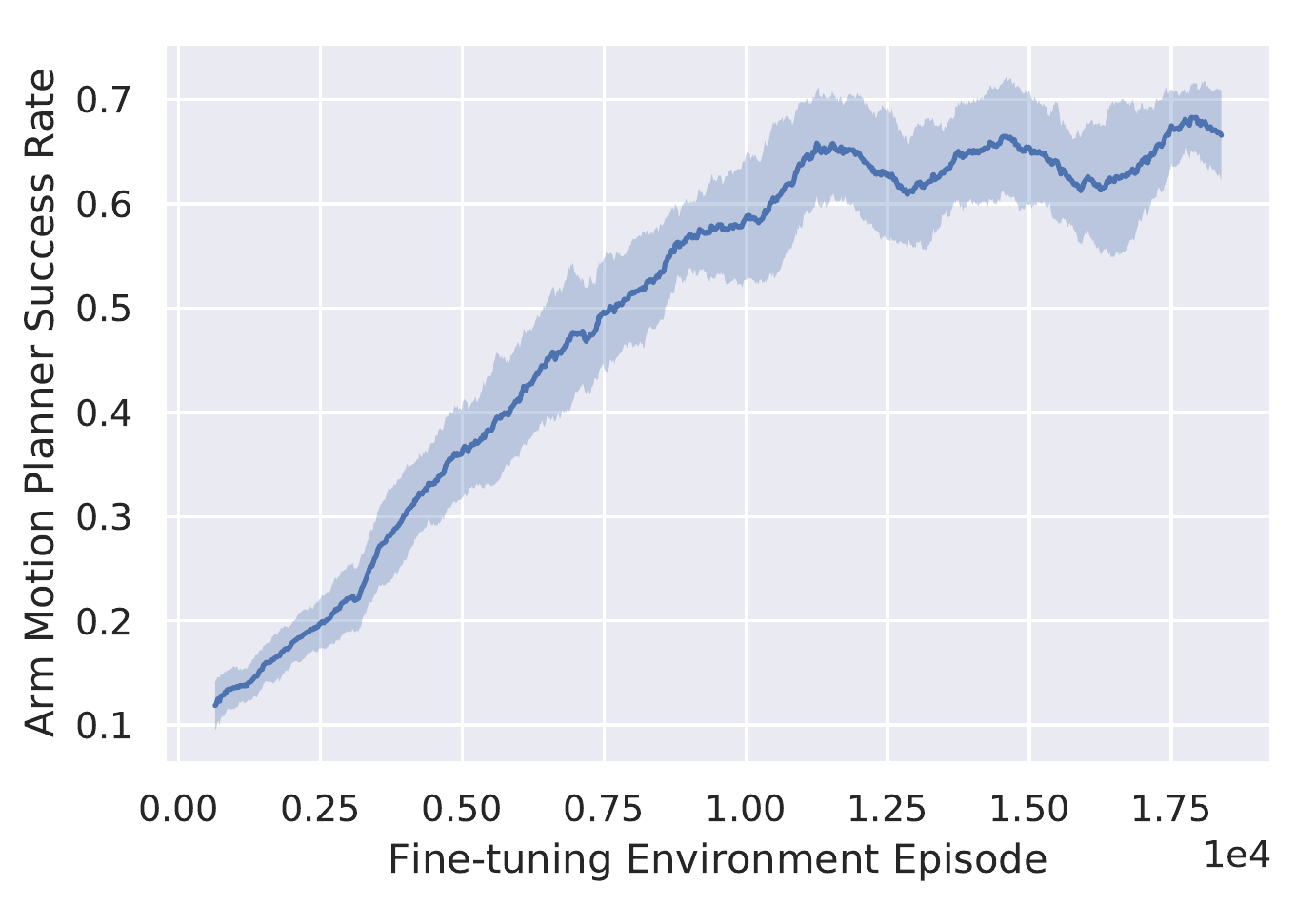}
    \caption{Arm MP Success Rate}
\end{subfigure}
\begin{subfigure}[b]{0.3\textwidth}
    \includegraphics[height=3cm]{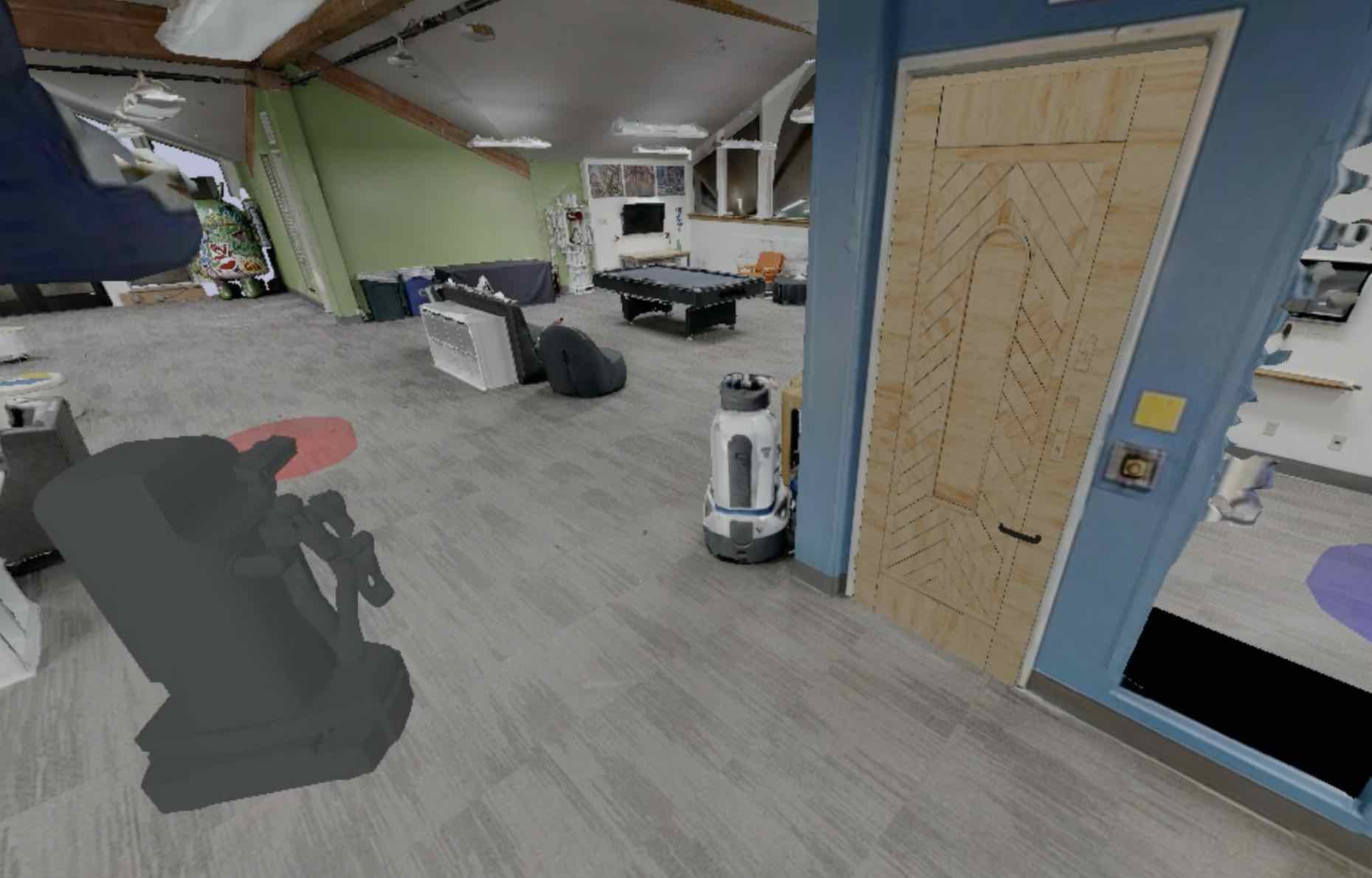}
    \caption{}
\end{subfigure}
\begin{subfigure}[b]{0.2\textwidth}
    \includegraphics[height=3cm]{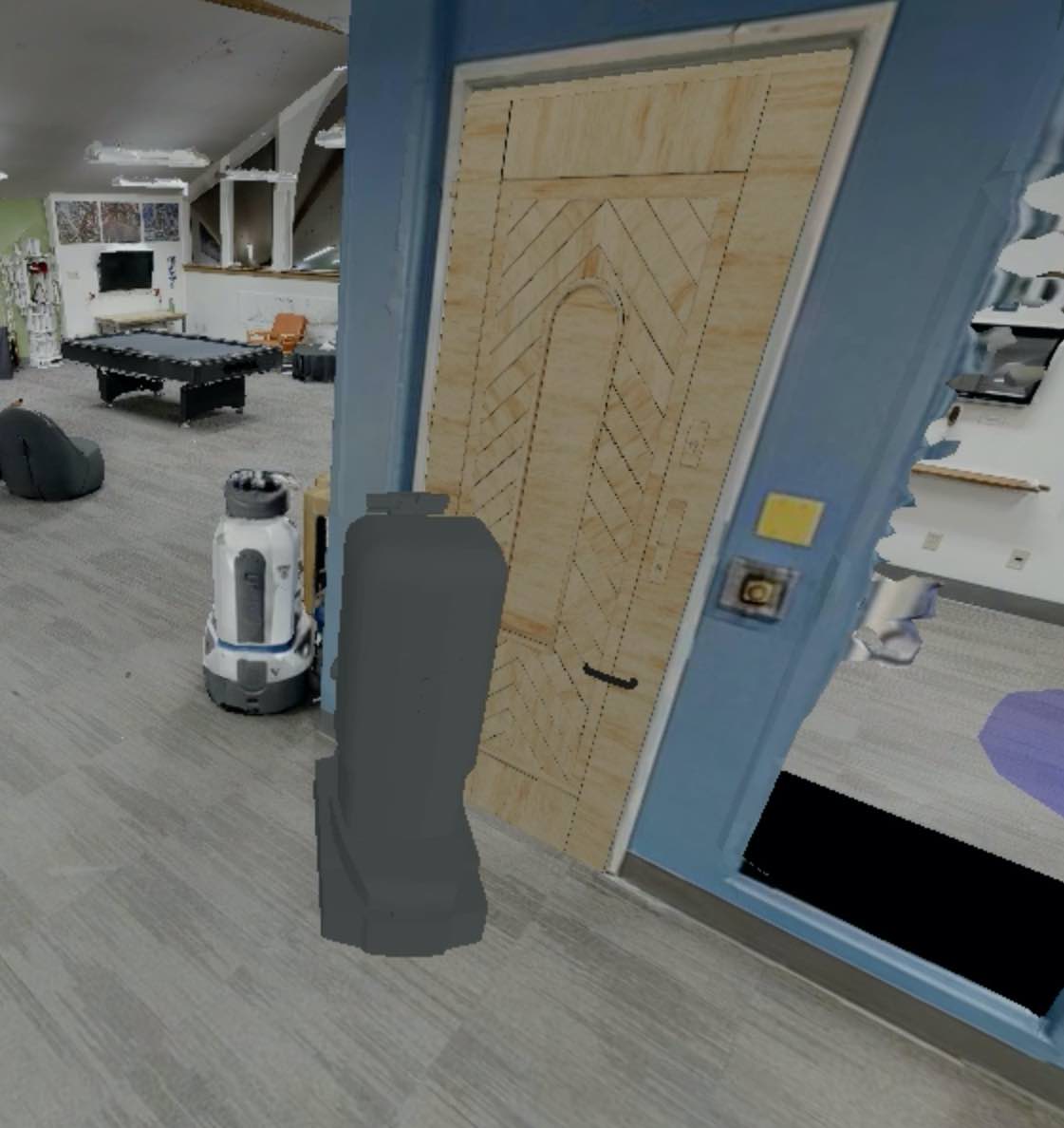}
    \caption{}
\end{subfigure}
\begin{subfigure}[b]{0.2\textwidth}
    \includegraphics[height=3cm]{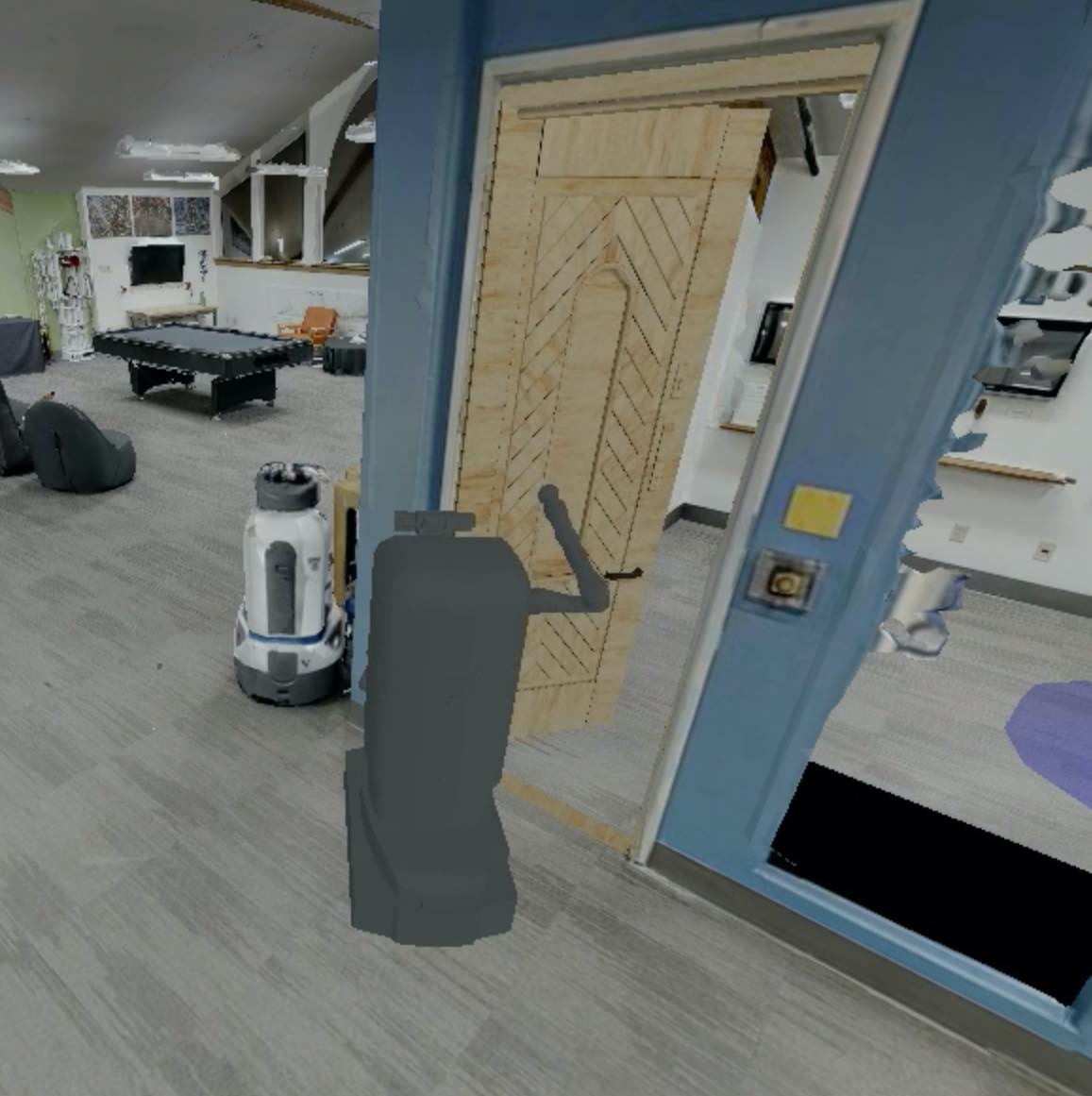}
    \caption{}
\end{subfigure}
\begin{subfigure}[b]{0.2\textwidth}
    \includegraphics[height=3cm]{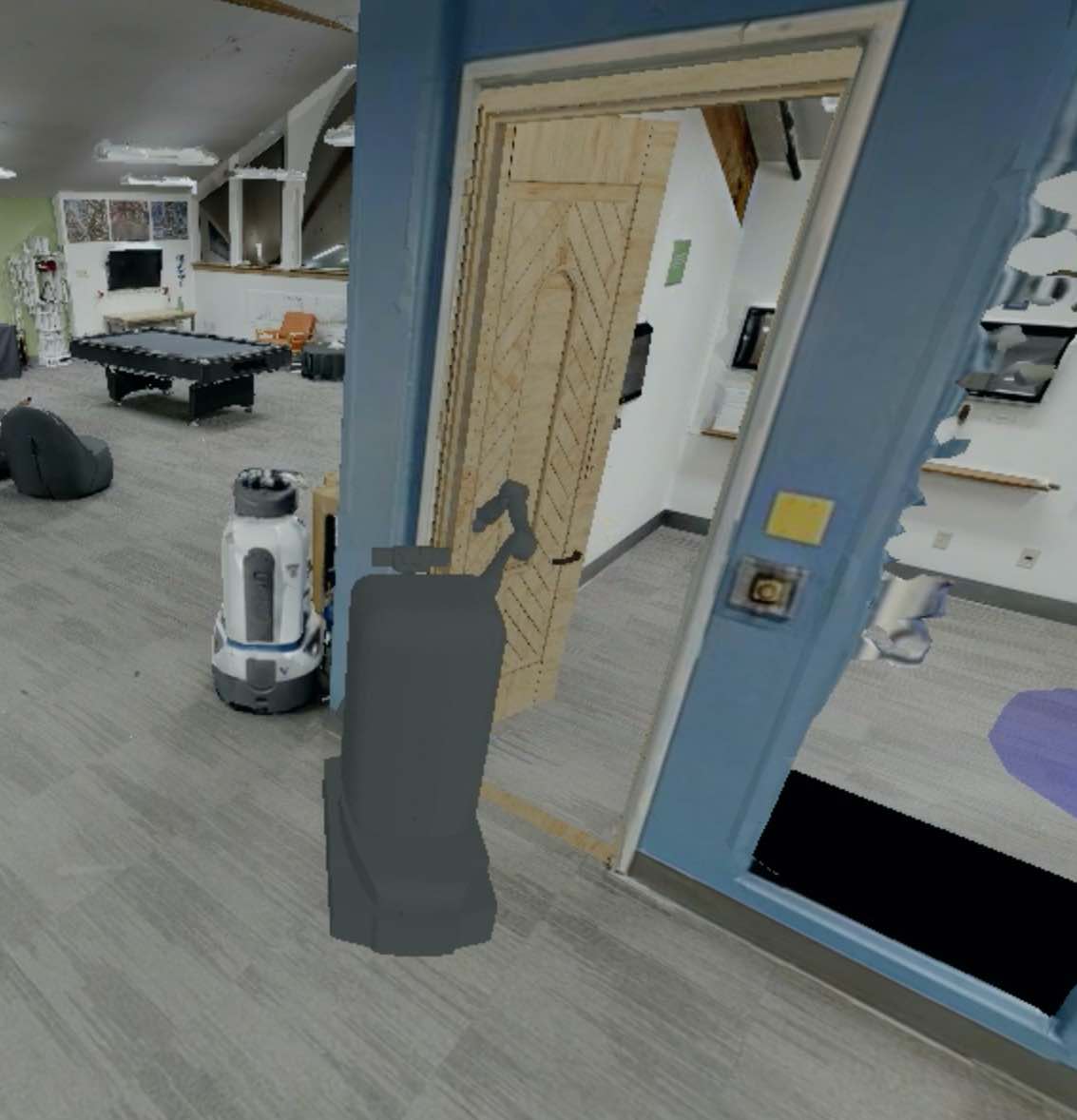}
    \caption{}
\end{subfigure}

\caption{\small{Fine-tuning on the new robot Movo. (a) We choose Movo because it is geometrically similar to Fetch. (b) We show that with only $\num{2e4}$ fine-tuning episodes, we can significantly improve the success rate for the new robot. Our Subgoal Generation Policy learns to adapt the subgoals to better accommodate the new embodiment, e.g. setting the base subgoal slightly further away from the door so that the new, longer arm has enough clearance for planning. (c) shows the arm motion planner success rate through the fine-tuning process, as the subgoal generation gets refined, the arm motion planner success rate increase significantly. (d)-(g) show a successful execution trajectory of Movo Robot on \texttt{PushDoorNav} task.\label{fig:robot_ft}}}
\end{center}
\end{figure*}

\paragraph{Fine-tuning with A New Embodiment}

\begin{table}[t]
    \begin{center}
  
  \begin{subfigure}{0.24\textwidth}
  \resizebox{\textwidth}{!}{
  \begin{tabular}{lcc} 
  \toprule
    Arm MP & Success rate\\ 
  \midrule
    \textbf{RRT-Connect} & $\bf{1.0}$ \\
    Lazy PRM & $1.0$ $\bf{(+0.0)}$ \\
    \bottomrule
  \end{tabular}
  }
 \caption{TabletopReachM}
\end{subfigure}
\hfill
 \begin{subfigure}{0.33\textwidth}
 \resizebox{\textwidth}{!}{
  \begin{tabular}{lccc} 
  \toprule
    Base MP & Arm MP & Success rate\\ 
  \midrule
    \textbf{RRT-Connect} & \textbf{RRT-Connect} & $\bf{0.91}$ \\
    RRT-Connect & Lazy PRM & $0.93$ $\bf{(+0.02)}$ \\
    Lazy PRM & RRT-Connect & $0.91$ $\bf{(+0.0)}$\\
    Lazy PRM & Lazy PRM & $0.87$ $\bf{(-0.04)}$ \\
    \bottomrule
  \end{tabular}
  }
 \caption{InteractiveObstaclesNav}
\end{subfigure}
\hfill
 \begin{subfigure}{0.35\textwidth}
 \resizebox{\textwidth}{!}{
  \begin{tabular}{lccc} 
  \toprule
    Base MP & Arm MP & \# Closed ($\SI{10}{cm}$)\\ 
  \midrule
    \textbf{RRT-Connect} & \textbf{RRT-Connect} & $\bf{0.34}$ \\
    RRT-Connect & Lazy PRM & $0.37$ $\bf{(+0.03)}$ \\
    Lazy PRM & RRT-Connect & $0.35$ $\bf{(+0.01)}$\\
    Lazy PRM & Lazy PRM & $0.38$ $\bf{(+0.04)}$\\
    \bottomrule
  \end{tabular}
  }
 \caption{ArrangeChairMM}
\end{subfigure}
 \caption{\small{This table complements Table~\ref{tbl:mp} and includes more tasks. Our policy trained with RRT-Connect as the motion planner for base and arm can perform equally well when we change to Lazy PRM at test time (the first row shows the setup used at training).\label{tbl:mp_additional}}}
\end{center}
\vspace{-0.5cm}
\end{table}

In this section, we want to stress test our methods to see if they can be transferred onto a new robot. We selected Movo Mobile Manipulator because it has a relatively similar embodiment to that of Fetch. However, there are still some major differences between the two robots such as the size and the shape of the base, the kinematics of the arm, and the on-board camera location. As we expect, zero-shot transfer to Movo doesn't work very well. The typical failure mode is that Movo moves its base too close to the object (because it has a larger base) and doesn't leave enough clearance for the arm motion planner to find a plan for arm subgoals. Following a similar fine-tuning paradigm as before, we first train \texttt{PushDoorNav} task with Fetch until convergence. Then we switch to Movo and continue training. We observe that the performance steadily improves with only $\num{2e4}$ fine-tuning episodes (see Fig.~\ref{fig:robot_ft}). This is a significant improvement over training from scratch. We can achieve this improvement because the rough locations of the subgoals are reasonable, and they just need some small adjustment to better suit the new embodiment. Fig.~\ref{fig:robot_ft} (d)-(g) show  an execution trajectory of Movo Robot on \texttt{PushDoorNav} task, in which we find that compared with Fetch, the robot stops further away in front of the door to facilitate planning for Movo's longer arms.

\subsection{Additional Analysis}

\paragraph{Generalization to New Motion Planners}
In Section~\ref{subsec:res}, we show our methods can zero-shot generalize to Lazy PRM even though they are trained with RRT-Connect. We include additional experimental results in Table~\ref{tbl:mp_additional} to support this point.

\begin{figure*}[t]
\begin{center}
\begin{subfigure}{0.29\textwidth}
    \includegraphics[width=\linewidth]{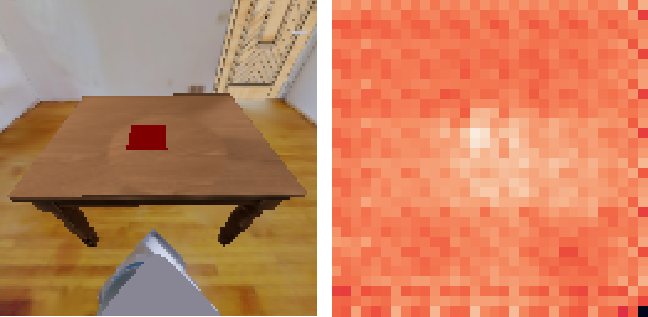}
    \caption{\texttt{TabletopReachM}}
\end{subfigure}
\rulesep
\begin{subfigure}{0.29\textwidth}
    \includegraphics[width=\linewidth]{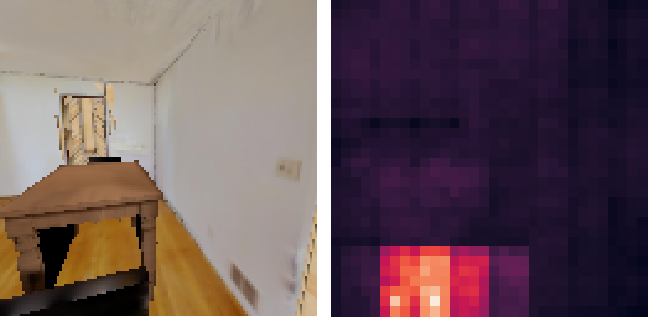}
    \caption{\texttt{ArrangeChairMM}}
\end{subfigure}
\rulesep
\begin{subfigure}{0.33\textwidth}
    \includegraphics[width=\linewidth]{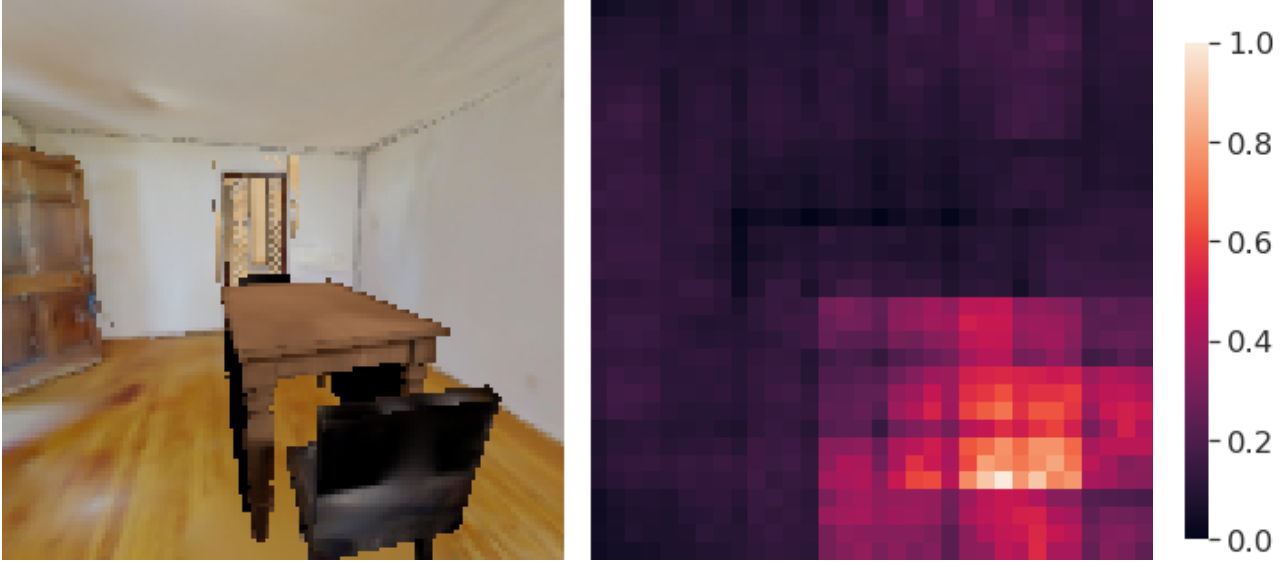}
    \caption{\texttt{ArrangeChairMM}}
\end{subfigure}

\begin{subfigure}{0.29\textwidth}
    \includegraphics[width=\linewidth]{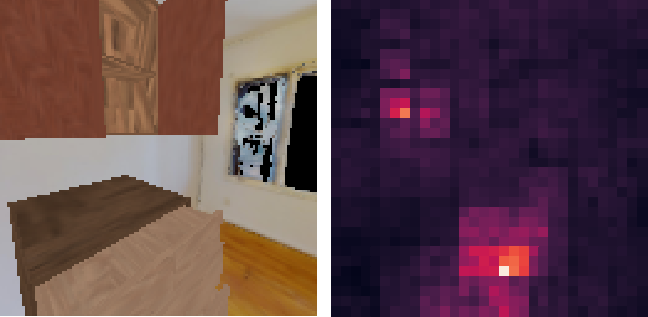}
    \caption{\texttt{ArrangeKitchenMM}}
\end{subfigure}
\rulesep
\begin{subfigure}{0.29\textwidth}
    \includegraphics[width=\linewidth]{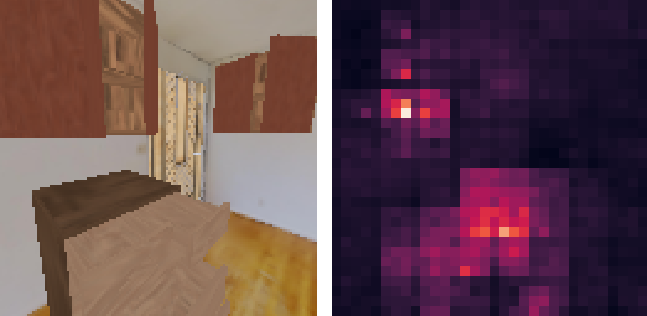}
    \caption{\texttt{ArrangeKitchenMM}}
\end{subfigure}
\rulesep
\begin{subfigure}{0.33\textwidth}
    \includegraphics[width=\linewidth]{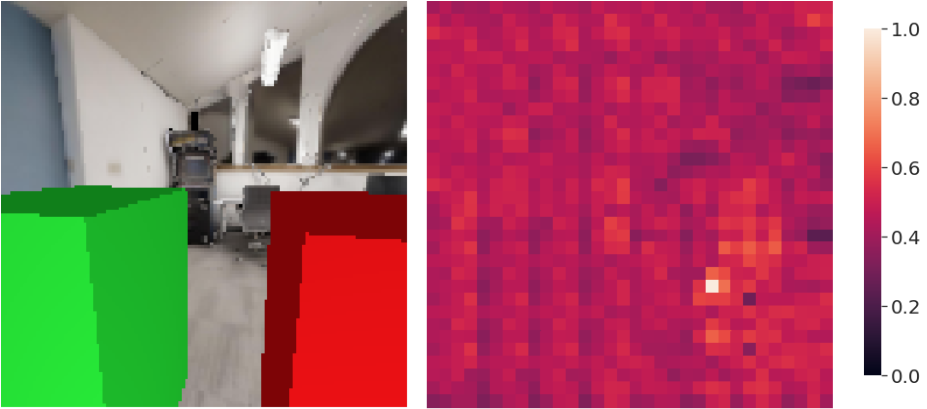}
    \caption{\texttt{InteractiveObstaclesNav}}
\end{subfigure}

\begin{subfigure}{0.29\textwidth}
    \includegraphics[width=\linewidth]{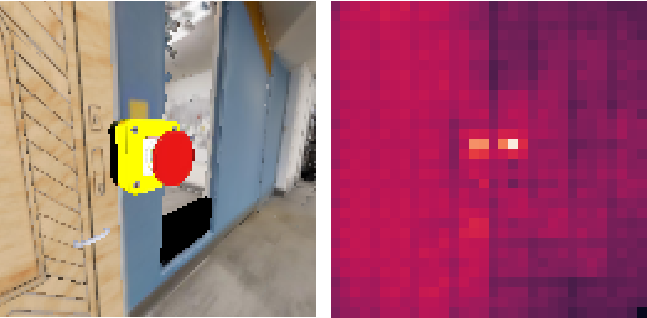}
    \caption{\texttt{ButtonDoorNav}}
\end{subfigure}
\rulesep
\begin{subfigure}{0.29\textwidth}
    \includegraphics[width=\linewidth]{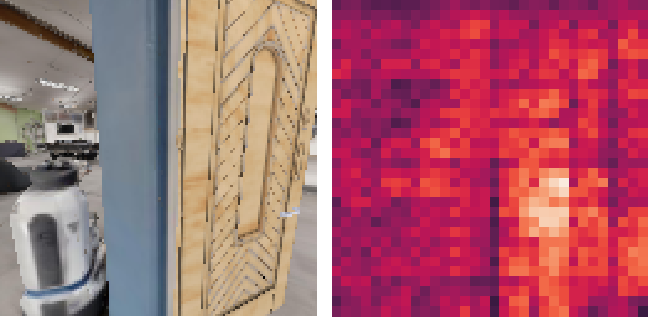}
    \caption{\texttt{PushDoorNav}}
\end{subfigure}
\rulesep
\begin{subfigure}{0.33\textwidth}
    \includegraphics[width=\linewidth]{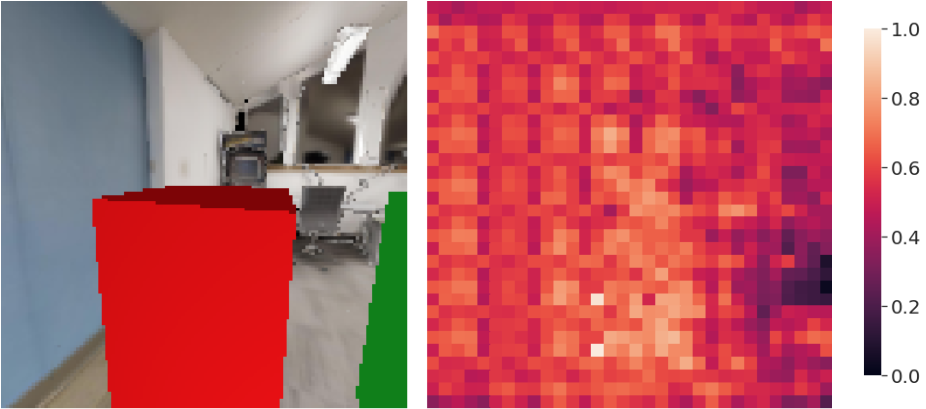}
    \caption{\texttt{InteractiveObstaclesNav}}
\end{subfigure}

\end{center}

\caption{\small{This figure shows visualization of ReLMoGen-D action maps during evaluation. The image pairs contain the input RGB frames on the left and normalized predicted Q-value maps on the right. The predicted Q-value spikes up at image locations that enable useful interactions, e.g. goals, chairs, cabinets, doors, buttons, and obstacles. (a) shows that the agent correctly predicts high Q-value on the goal. (b) and (c) show that the agent learns to push the most suitable part of the chair. (d) shows that the agent prioritizes pushing a drawer that is ``more open" than an almost closed cabinet to harvest more reward. Vice versa for (e). (f) and (i) show that the agent learns only the red obstacle is movable and correctly predicts high Q-value on the red obstacle and low Q-value on the green one.  (g) shows that the agent precisely identify location of the button that activates the door. (h) shows that the agent prioritizes pushing the part of the door that is reachable by the arm.}}
\label{fig:q_value_vis_additional}
\end{figure*}

\paragraph{Subgoal Interpretability}
Fig.~\ref{fig:q_value_vis_additional} shows the Q-value maps generated by ReLMoGen-D across different tasks. We observe that the learned subgoals set by our Subgoal Generation Policy (SGP-D) are highly interpretable. High Q-values usually correspond to beneficial interactions, such as goals, chairs, cabinets, doors, buttons, and obstacle.

\begin{figure*}[h]
\begin{center}
\begin{subfigure}{0.49\textwidth}
\includegraphics[width=\linewidth,trim=1.5cm 0 1cm 0, clip]{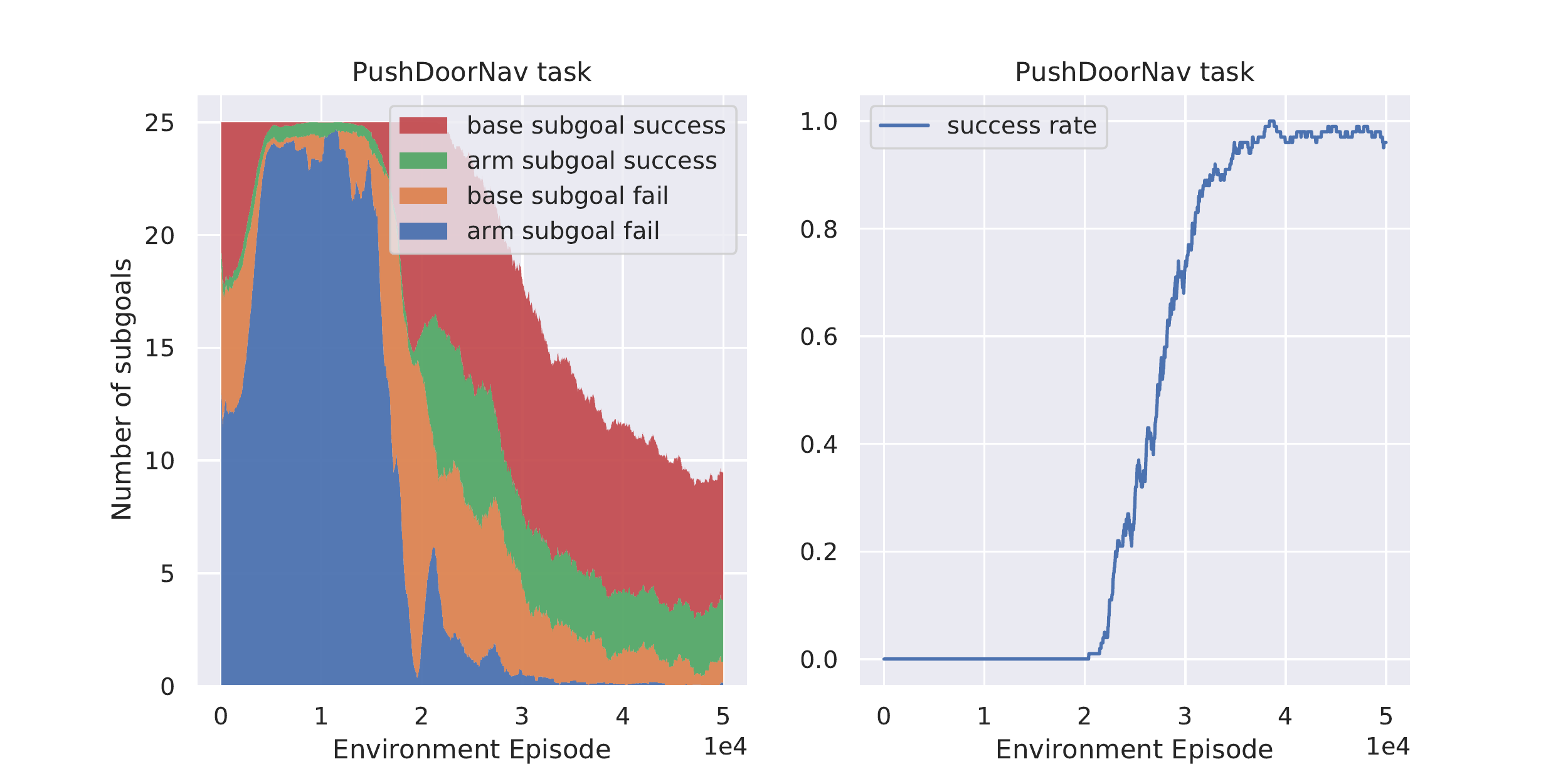}
\caption{ReLMoGen-R on \texttt{PushDoorNav}}
\end{subfigure}
\begin{subfigure}{0.49\textwidth}
\includegraphics[width=\linewidth,trim=1.5cm 0 1cm 0, clip]{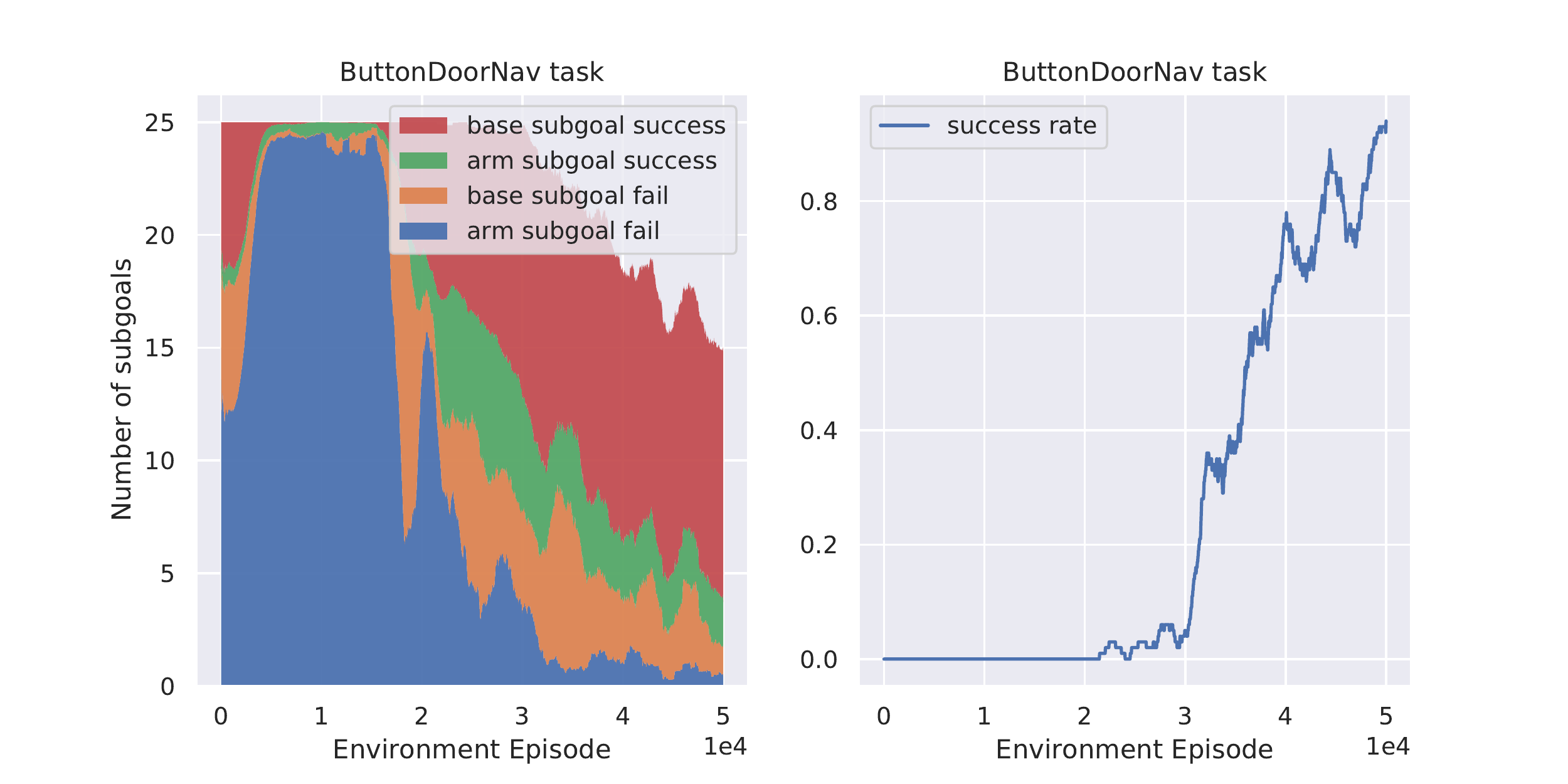}
\caption{ReLMoGen-R on \texttt{ButtonDoorNav}}
\end{subfigure}
\begin{subfigure}{0.49\textwidth}
\includegraphics[width=\textwidth,trim=1.5cm 0 1cm 0, clip]{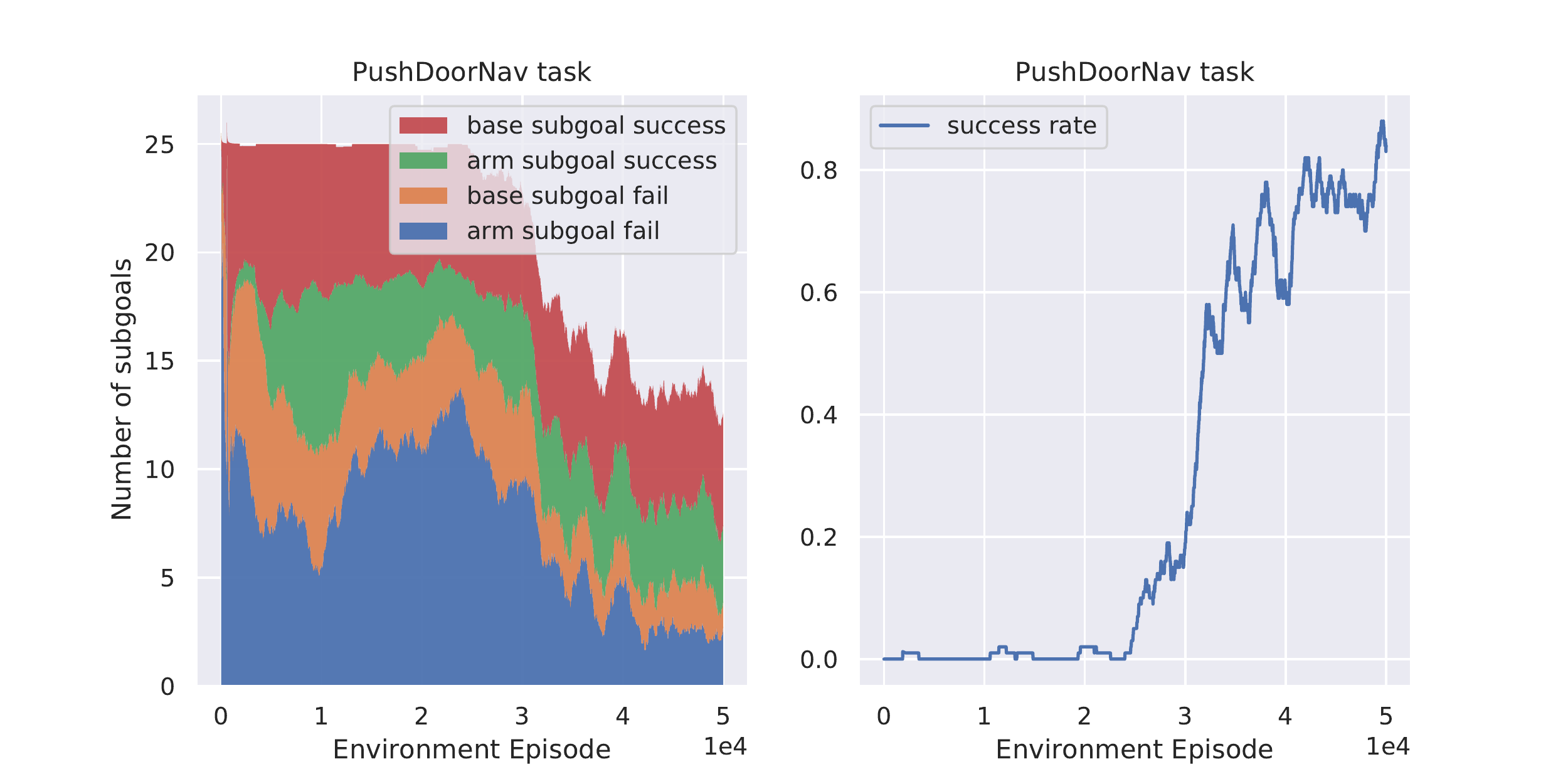}
\caption{ReLMoGen-D on \texttt{PushDoorNav}}
\end{subfigure}
\begin{subfigure}{0.49\textwidth}
\includegraphics[width=\textwidth,trim=1.5cm 0 1cm 0, clip]{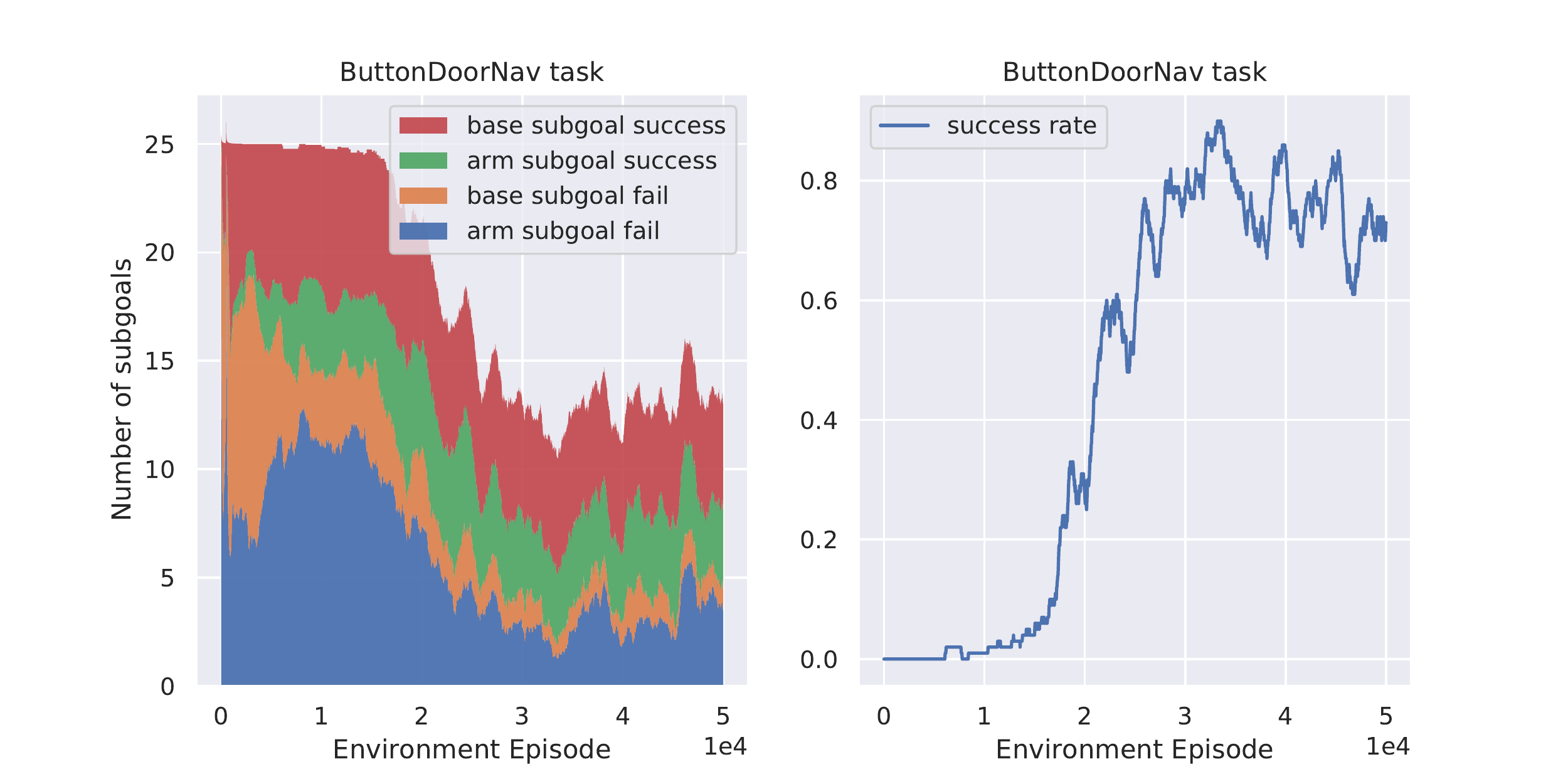}
\caption{ReLMoGen-D on \texttt{ButtonDoorNav}}
\end{subfigure}

\caption{\small{Subgoal distribution during training. The subgoal success rate increases over time, indicating our policy learns to use MG better and set more feasible subgoals as training progresses. The policy is also able to accomplish the task with fewer and fewer subgoals.\label{fig:subgoal_dist}}}
\end{center}
\end{figure*}

\paragraph{Subgoal distribution during training}
We track and visualize the subgoal distribution during training in Fig.~\ref{fig:subgoal_dist}. Base or arm subgoal failures represent the cases in which the base or arm motion planner fails to find feasible plans. We observe that our policy learn to utilize motion generators better and set more feasible subgoals as training progresses.

\begin{figure*}[t!]
\begin{center}
\begin{subfigure}{0.99\textwidth}
\includegraphics[width=\linewidth]{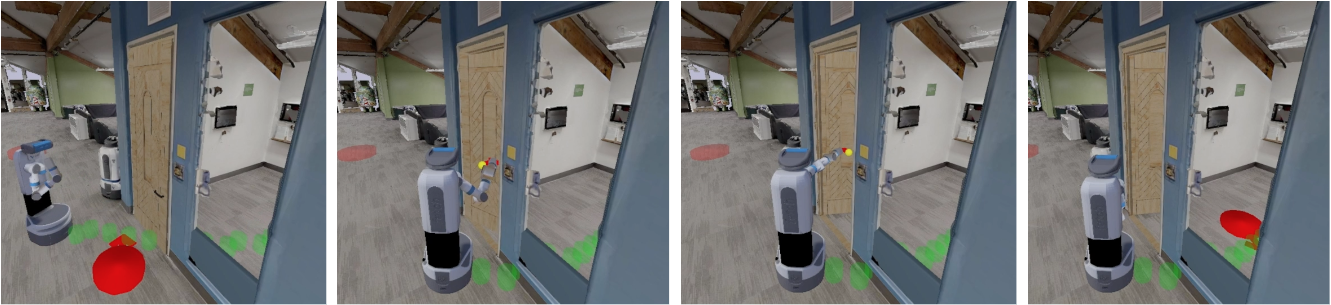}
\caption{\texttt{PushDoorNav}}
\end{subfigure}
\begin{subfigure}{0.99\textwidth}
\includegraphics[width=\linewidth]{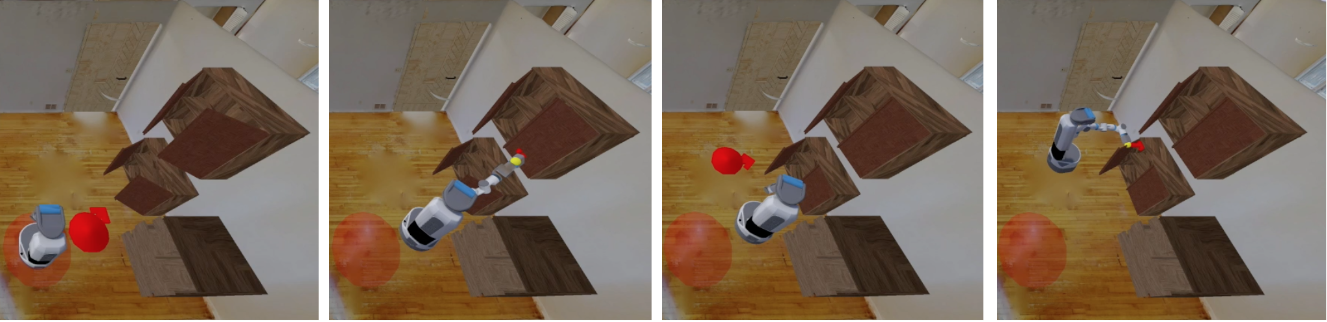}
\caption{\texttt{ArrangeKitchenMM}}
\end{subfigure}
\caption{\small{Policy visualization for ReLMoGen. A base subgoal is depicted as a red circle with an arrow on the floor to indicate the desired base position and yaw angle. An arm subgoal is depicted as a yellow ball that indicates the desired end-effector position, and a red arrow that indicates the desired pushing action from that position. For \texttt{PushDoorNav} task, the robot first navigates to the front of the door, pushes a few times until the door is open, and navigates into the room. In \texttt{ArrangeKitchenMM} task, the robot first navigates to the closest cabinet door, closes it, then navigates to the other side of the cabinet, and closes another door. Please refer to our website for more policy visualization.\label{fig:policy_vis}}}
\end{center}
\end{figure*}

\paragraph{Policy Visualization}
We visualize the robot trajectories and learned subgoals of ReLMoGen for \texttt{PushDoorNav} and \texttt{ArrangeKitchenMM} tasks in Fig.~\ref{fig:policy_vis}. More policy visualization is on our website.

\subsection{Sim2Real Transfer Potential}

We believe the characteristics of our method are well suited to transfer to real robots. In this section we highlight these characteristics together with justifications for the potential of ReLMoGen to transfer from simulation to real (Sim2Real).

First, the solutions presented in our paper for navigation, manipulation and mobile manipulation based on ReLMoGen use only virtual signals from the onboard simulated sensors of the robot; no ground truth information from the environment is used as input to our policy network. For navigation tasks we assume our solution know the initial and goal locations, and the location of the robot in a map of the layout, as it is provided by any 2D localization method using the onboard LiDAR.

Second, we analyze the two main sources of domain gap. Simulation provides an efficient domain to develop and test algorithms. However, due to differences between simulation and the real world, there is a potential risk for the learned policies to not transfer well to a real robot. This risk is built on two main sources, the perception domain gap~\cite{bousmalis2018using, rao2020rl} and the dynamics domain gap~\cite{ramos2019bayessim, chebotar2019closing}. 

\begin{figure*}[ht]
\begin{center}
\begin{subfigure}{0.16\textwidth}
\includegraphics[width=\linewidth]{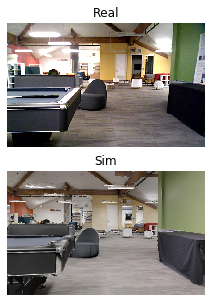}
\caption{\small{RGB}}
\end{subfigure}
\begin{subfigure}{0.185\textwidth}
\includegraphics[width=\linewidth]{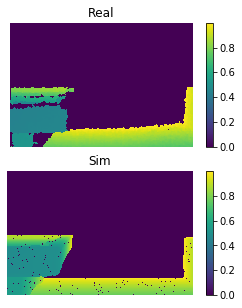}
\caption{\small{Depth}}
\end{subfigure}
\begin{subfigure}{0.102\textwidth}
\includegraphics[width=\linewidth]{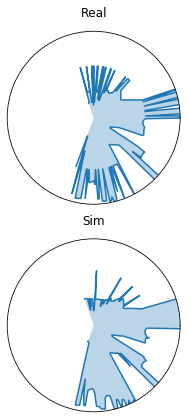}
\caption{\small{LiDAR}}
\end{subfigure}
\begin{subfigure}{0.16\textwidth}
\includegraphics[width=\linewidth]{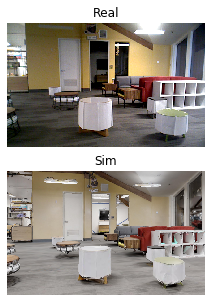}
\caption{\small{RGB}}
\end{subfigure}
\begin{subfigure}{0.185\textwidth}
\includegraphics[width=\linewidth]{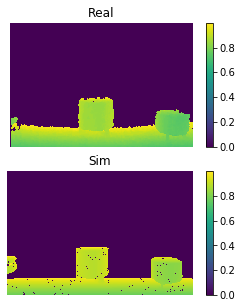}
\caption{\small{Depth}}
\end{subfigure}
\begin{subfigure}{0.102\textwidth}
\includegraphics[width=\linewidth]{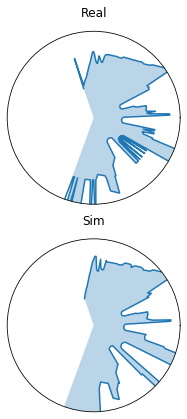}
\caption{\small{LiDAR}}
\end{subfigure}
\caption{\small{Simulation and Real Comparison. (a-c) and (d-f) are two sets of observations at the same location in simulation and in the real world. They are visually highly similar, highlighting the fidelity of our simulator. }\label{fig:sim2real}}
\end{center}
\end{figure*}

\paragraph{Perception Domain Gap}
To reduce the perception domain gap, we used a state-of-the-art robot simulation engine iGibson~\cite{xia2020interactive}, which has been shown previously to facilitate successful sim2real transfer of visual policies~\cite{kang2019generalization,meng2019neural}. Pairs of simulation and real observations at equivalent robot poses are shown in Fig~\ref{fig:sim2real}. The observations are visually similar, which indicates a small perception domain gap. If the perception gap were still to exist, we would include pixel-level domain adaptation methods~\cite{xia2018gibson, rao2020rl} to reduce it.

\paragraph{Dynamics Domain Gap} 
Another major risk for sim2real transfer is the dynamics domain gap~\cite{tan2018sim, chebotar2019closing}: actions in simulation and in the real world do not have the same outcome. In ReLMoGen's proposed structure, the motion generator handles the dynamics domain gap. The motion generator executes with low level joint controllers the trajectories planned by a motion planner. This process can be executed with small deviations to the plan, both in simulation and in the real world. Then the question becomes whether we can transfer between different motion planning methods and implementations, since the real robot may potentially use a different motion generator. We show in the paper (Table~\ref{tbl:mp} and Table~\ref{tbl:mp_additional}) that we can transfer from RRT-Connect to Lazy PRM with minimal performance drop. In other words, our learned Subgoal Generation Policy is able to output base and arm subgoals whose outcome is largely independent of the underlying motion generator, indicating robustness to changes in the motion planner.


\end{document}